\documentclass{article}

\usepackage{graphicx}
\usepackage{booktabs} 

\usepackage{hyperref}

\usepackage[accepted]{icml2024}

\usepackage{amsmath}
\usepackage{amssymb}
\usepackage{mathtools}
\usepackage{amsthm}

\usepackage{hyperref}
\usepackage{xcolor}
\usepackage[utf8]{inputenc} 
\usepackage[T1]{fontenc}    
\usepackage{booktabs}       
\usepackage{amsfonts}       
\usepackage{nicefrac}       

\usepackage{amsthm}
\usepackage{multirow}
\usepackage{mathtools} 

\usepackage{graphicx}
\usepackage{caption}
\usepackage{subcaption}
\usepackage{float}
\usepackage{wrapfig}
\usepackage{lipsum}

\usepackage{amsmath}

\usepackage{pifont}

\usepackage{color}

\usepackage[capitalize,noabbrev]{cleveref}

\theoremstyle{plain}

\theoremstyle{definition}

\theoremstyle{remark}

\usepackage{xcolor}

\usepackage[textsize=tiny]{todonotes}

\icmltitlerunning{Synergistic Integration of Coordinate Network and Tensorial Feature for Improving NeRFs from Sparse Inputs}

\begin{document}

\twocolumn[
\icmltitle{Synergistic Integration of Coordinate Network and Tensorial Feature for Improving Neural Radiance Fields from Sparse Inputs
}



\icmlsetsymbol{equal}{*}

\author{%
  Mingyu Kim, Jun-seong Kim, Se-Young Yun, Jin-Hwa Kim \\
  KAIST AI, POSTECH EE, NAVER AI Lab, SNU AIIS\\
  \texttt{\{callingu, yunseyoung\}@kaist.ac.kr},
  \texttt{gucka28@postech.ac.kr}
  \texttt{j1nhwa.kim@navercorp.com} \\
}

\begin{icmlauthorlist}
\icmlauthor{Mingyu Kim}{yyy}
\icmlauthor{Jun-Seong Kim}{sch}
\icmlauthor{Se-Young  Yun}{yyy}
\icmlauthor{Jin-Hwa Kim}{comp}
\end{icmlauthorlist}

\icmlaffiliation{yyy}{KAIST AI}
\icmlaffiliation{sch}{POSTECH EE}
\icmlaffiliation{comp}{NAVER AI Lab \& SNU AIIS}

\icmlcorrespondingauthor{Jin-Hwa Kim and Se-Young Yun}{j1nhwa.kim@navercorp.com; yunseyoung@kaist.ac.kr}

\icmlkeywords{Machine Learning, ICML}

\vskip 0.3in
]



\printAffiliationsAndNotice{} 

\begin{abstract}
The multi-plane representation has been highlighted for its fast training and inference across static and dynamic neural radiance fields. This approach constructs relevant features via projection onto learnable grids and interpolating adjacent vertices. 
However, it has limitations in capturing low-frequency details and tends to overuse parameters for low-frequency features due to its bias toward fine details, despite its multi-resolution concept. This phenomenon leads to instability and inefficiency when training poses are sparse. 
In this work, we propose a method that synergistically integrates multi-plane representation with a coordinate-based MLP network known for strong bias toward low-frequency signals.
The coordinate-based network is responsible for capturing low-frequency details, while the multi-plane representation focuses on capturing fine-grained details. 
We demonstrate that using residual connections between them seamlessly preserves their own inherent properties.
Additionally, the proposed progressive training scheme accelerates the disentanglement of these two features. 
We demonstrate empirically that our proposed method not only outperforms baseline models for both static and dynamic NeRFs with sparse inputs, but also achieves comparable results with fewer parameters.
\end{abstract}

\section{Introduction}
\label{sec_introduction}

Neural Radiance Fields (NeRFs) have gained recognition for their ability to create realistic images from various viewpoints using the volume rendering technique \citep{mildenhall2021nerf}. 
Early studies have demonstrated that multi-layer perception (MLP) networks, combined with sinusoidal encoding, can effectively synthesize 3-dimensional novel views \citep{mildenhall2021nerf, tancik2020fourier, sitzmann2020implicit, martel2021acorn, lindell2022bacon}. 
These studies have shown that coordinate-based MLP networks exhibit strong low-frequency bias, and incorporating sinusoidal encoding allows for capturing both low and high-frequency signals. 

For broader real-world applicability, extensive efforts have focused on reliably constructing radiance fields in cases of sparse input data \citep{yu2021pixelnerf, wang2021ibrnet, chen2021mvsnerf, jain2021putting}.
One set of solutions tackled this by leveraging a pretrained image encoder to compare rendered scenes against consistent 3D environments \citep{yu2021pixelnerf, wang2021ibrnet, chen2021mvsnerf, jain2021putting}.  
Another approach incorporated additional information, such as depth or color constraints, to maintain 3-dimensional coherence \citep{deng2022depth, yuan2022neural, roessle2022dense, truong2023sparf}. 
Methods progressively adjusting the frequency spectrum of position encoding have proven effective in counteracting overfitting without additional information \citep{yang2023freenerf, song2023harnessing}. However, sinusoidal encoding requires over 5 hours of training time, complicated regularizations, and exhibits a performance gap from explicit representation. 

\begin{figure*}[!t]{
    \vskip 0.2in
    \begin{center}
        \begin{subfigure}{0.42\textwidth}
            \includegraphics[width=\textwidth]{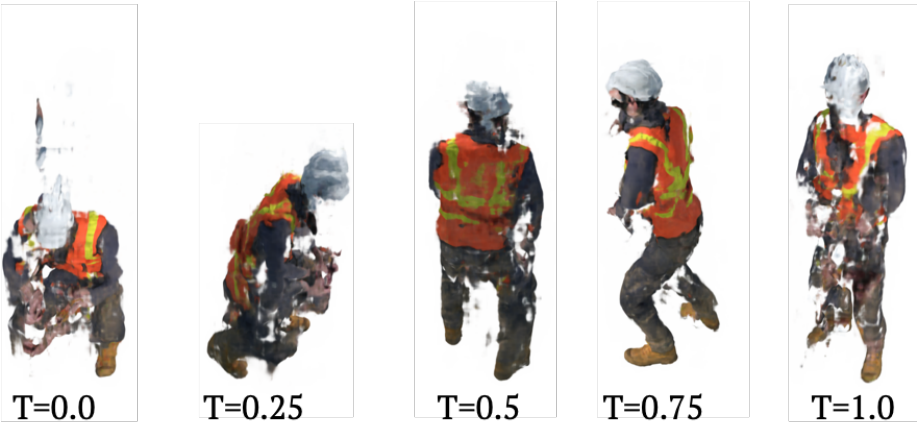}
            \caption{HexPlane}
            \label{fig_introduction_standup_hexplane}
        \end{subfigure}    
        \begin{subfigure}{0.42\textwidth}
            \includegraphics[width=\textwidth]{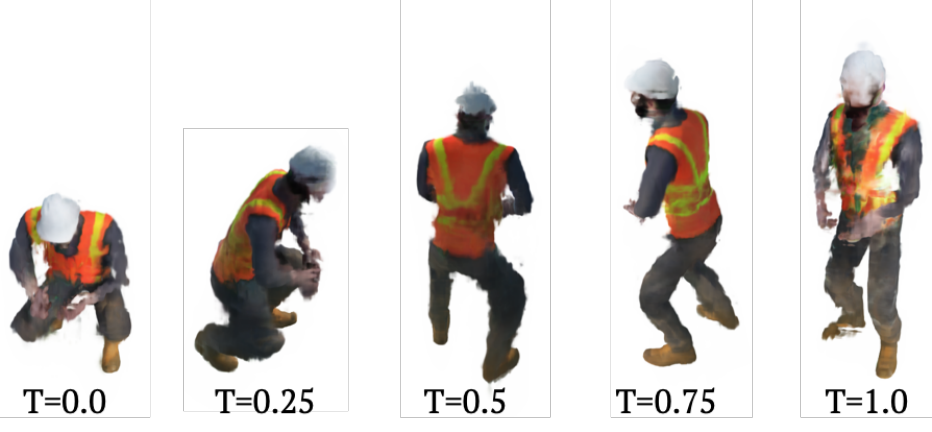}    
            \caption{Ours}
            \label{fig_introduction_standup_ours}
        \end{subfigure}    
        
    \caption{The qualitative results of the \texttt{standup} case in dynamic NeRFs using 25 training poses $($about $17\%$ of the original data$)$. This is challenging due to the limited information available along the time axis. Figure (a) is produced by HexPlane. \citep{cao2023hexplane}. Figure (b) is the rendered image of the proposed method.}
    \label{fig_introduction_images}
    \end{center}
    \vskip -0.2in
}    
\end{figure*}
\vspace{-0.02in}


Approaches explicitly parameterizing spatial attributes through voxel-grid, hashgrid, and multi-plane have been introduced \citep{muller2022instant, chen2022tensorf, chan2022efficient}. 
These methods dramatically reduce training time and produce cleaner and more realistic images, meanwhile demanding excessive memory consumption \citep{lee2023coordinate}. 
The recent works found those representations struggle with low-frequency detail and overfit to high-frequency signals, especially when applying for dynamic scenes, despite using multi-scale representations \citep{fridovich2023k, cao2023hexplane, peng2023representing}. 
While those have marginally had success in the reconstruction of NeRF with the assistance of denoising penalties like total variation \citep{sun2023vgos, fridovich2023k}, 
they still lack adequate representation of low-frequency spectral features like object shapes and dynamic motion, as shown in \autoref{fig_introduction_standup_hexplane}. 

To alleviate this issue, we introduce a simple yet powerful approach to fundamentally improve the performance of static and dynamic NeRFs from sparse inputs. 
In this framework, coordinate-based MLP features handle low-frequency context, while multiple-plane features capture fine-grained details aligning with the spectral bias mentioned earlier. 
This approach yields three main benefits. First, aligning with the distinct spectral biases of heterogeneous features results in a model less sensitive to hyperparameters and performance variations related to the scene. This is achieved by avoiding both underfitting and overfitting. 
Second, it allows for stable training through gradual changes in their spectral biases, as discussed in the works by \citet{lin2021barf, yang2023freenerf}. 
Lastly, it facilitates efficient parameter allocation by replacing the need for a low-resolution grid with coordinate-based features. 

We achieve this by implementing a residual concatenation of coordinates and multi-plane features across the first two hidden layer blocks, enhancing the efficiency in responding to the coordinates. The images generated by the proposed method exhibit enhanced clarity regarding global contexts and fewer artifacts compared to baselines, as illustrated in \autoref{fig_introduction_standup_ours}.
Our extensive experiments show that the proposed method achieves comparable results of multi-plane encoding with high denoising penalties in static NeRFs. Notably, it outperforms baselines in dynamic NeRFs from the sparse inputs. 
To summarize, we make the following contributions:
\begin{itemize}
    \item We prove that explicit parameterization has difficulty capturing low-frequency details, even when using multi-resolution grids or coordinate networks without precise integration.

    \item We validate the proposed method on static and dynamic NeRF tasks, including real-world cases with sparse inputs, while also examining how two features are separated and function independently.
    
    \item When we reduce the number of parameters, the proposed method still shows competitive performance. This results from skipping the allocation of a spatial low-resolution grid and replacing it with coordinate-based features.
\end{itemize}

\section{Related Work}
\label{sec_related_works}

\paragraph{Coordinate-based Network and Sinusoidal Encoding}  
In the initial studies of NeRFs, MLP networks with sinusoidal encoding were used to simultaneously describe low and high-frequency details \citep{mildenhall2021nerf, martin2021nerf, barron2021mip, barron2022mip}. 
However, a classical coordinate network without this encoding was found to be biased toward lower frequencies \citep{rahaman2019spectral, yuce2022structured}. 
The importance of positioning encoding and sinusoidal activation led to the fundamental exploration of the relationship between rendering performance and the frequency values of target signals \citep{tancik2020fourier, sitzmann2020implicit, fathony2021multiplicative, ramasinghe2022frequency}. 
\citet{lindell2022bacon} uncovered that improper high-frequency embedding results in artifacts negatively impacting reconstruction quality. 
\paragraph{Explicit Parameterization} Recent developments in explicit representations, such as voxel-grid, hash encoding, and multi-planes, have gained attention due to their fast training, rendering speed, and superior performance compared to positioning encoding-based networks \citep{liu2020neural, sun2022direct, muller2022instant, chen2022tensorf, cao2023hexplane, fridovich2023k}.
\citet{sun2022direct} introduced the direct voxel field, using minimal MLP layers to speed up training and rendering.
Instant-NGP, based on hash maps, provides multi-resolution spatial features and versatility, extending beyond 3-dimensional spaces to high-resolution 2-dimensional images \citep{muller2022instant}.
The multi-plane approach has been highlighted for its applicability in expanding to 4-dimensional without compromising generality, decomposing targets into multiple planes, with each plane responsible for a specific axis \citep{chen2022tensorf, cao2023hexplane, fridovich2023k}. 
Specifically, while the aforementioned approaches used special on-demand GPU computations for efficiency, this method achieves comparable speed and performance based on general auto-differential frameworks. This widens its applicability to tasks like 3D object generation, video generation, 3D surface reconstruction, and dynamic NeRF \citep{gupta20233dgen, yu2023video, wang2023pet, cao2023hexplane, fridovich2023k}. 
\paragraph{NeRFs in the Sparse Inputs} Early efforts incorporated pre-trained networks trained on large datasets to compensate for the lack of training data \citep{jain2021putting, yu2021pixelnerf, wang2021ibrnet}. 
Another alternative approach incorporated additional information, such as depth or color constraints, to ensure the preservation of 3D coherence \citep{deng2022depth, yuan2022neural, roessle2022dense, truong2023sparf}. 
Without the assistance of off-the-shelf models and additional, this line of works devised new regularization to train NeRFs with fewer than ten views. Reg-NeRF incorporates patch-wise geometry and appearance regularization \citep{niemeyer2022regnerf}. 
This paper verified that their regularization performs well on forward-facing examples like the LLFF dataset. They did not validate object-facing scenes because this assumption demands a high correlation between adjacent views. 
Recently, progressively manipulating the spectrum of positioning encoding from low to high frequency proved effective in mitigating over-fitting without relying on additional information \citep{yang2023freenerf, song2023harnessing}.
Compared to explicit representations, those still suffer from unsatisfactory visual quality, characterized by blurry boundaries. 
Recent studies using total variation regularization on explicit representations get rid of artifacts and construct smoother surfaces \citep{cao2023hexplane, fridovich2023k, sun2023vgos}. 
However, our findings indicate that this regularization can introduce artificial details that seem real but are not in the data. This can also result in the model failing to converge in certain scenes.
We present this problem in the experiments, both qualitatively and quantitatively.
\paragraph{Residual Connection in NeRFs}
The residual connections aim to enhance the efficiency in responding to input signals \citep{he2016identity}. 
In NeRFs, several studies have adopted residual connections to preserve context from earlier stages. 
\citet{shekarforoush2022residual} implemented these connections to accurately maintain a specific spectrum without overpowering high-frequency components.
\citet{mihajlovic2023resfields} utilized residual connections for maintaining temporal coherence from previous frames. 
While aforementioned works primarily leverage MLP layers, updating Motion-Based explicit representation through residual connections in a spatio-temporal domain is also presented \citep{wang2023neural}.

A few works attempted to use explicit parameterization with the sinusoidal encoding of coordinates, but their direction differs from our method since they mainly focus on enriching available features or fewer parameterization, as well as they did not demonstrate the role of tri-planes and coordinate features \citep{wang2023pet, peng2023representing, lee2023coordinate}. 
In this paper, our new approach proposes incorporating two distinct features: coordinate-based and multiple-plane features. 
We emphasize that the disentanglement of these two heterogeneous features is crucial for reliably constructing NeRFs in sparse inputs. The proposed method performs well even with higher-dimensional cases like dynamic NeRFs and extremely limited sparse inputs.


\begin{figure*}[!ht]{
    \centering
    \includegraphics[width=0.85\textwidth]{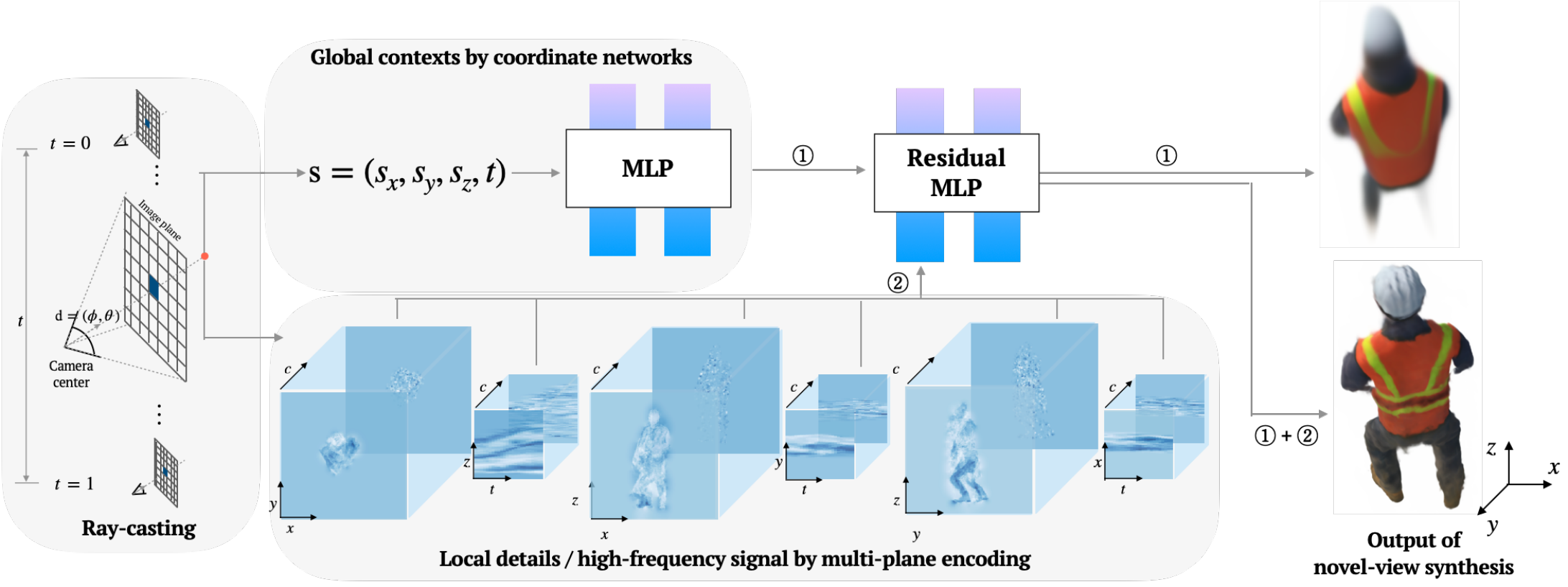}
    \caption{Conceptual illustration of the proposed method utilizing global contexts by coordinate networks and fine-grained details by multi-plane encoding. This method effectively displays two heterogeneous features. The number \texttt{1} indicates the use of coordinate network alone, while the symbol \texttt{1$+$2} means the use of both coordinated-based MLP network and multi-plane representation.}
    \label{fig_Method_overview}
}    
\vspace{-0.07in}
\end{figure*}
\vspace{-0.07in}
\section{Residual Neural Radiance Fields Spanning Diverse Spectrum for Sparse-Inputs}
\label{sec_method}
We propose a novel method that leverages multi-plane spatial features and coordinate-based networks.
In sparse input NeRFs, avoiding overfitting to training data is crucial because NeRFs typically use a scheme where one network is tailored to fit a specific scene.
Particularly, explicit representations based on local updates of grid structures significantly struggle with capturing global contexts. 
The proposed method capitalizes on a combination of distinct coordinate feature encoding techniques and multi-plane representations, which follows multi-plane representation by TensoRF and HexPlane in static and dynamic NeRFs \citep{chen2022tensorf,cao2023hexplane}, as well as ReLU-based coordinate networks. The detailed explanation of these features is included in \autoref{app_background}.

Here, we focus on how these two features are integrated to enhance the performance of NeRFs in handling sparse input data. 
As shown in \autoref{fig_Method_overview}, the proposed method encompasses two distinct contexts; both low and high-frequency information. When the coordinate network is used alone, the output is biased towards low frequency to facilitate global reasoning. However, when all features are engaged, it results in clear and intricate images.
We illustrate the main components in the following subsection. In \autoref{subsec_architecture}, we delve into the proposed residual-based architecture to facilitate the disentanglement of two heterogeneous features. 
Moving on to \autoref{subsec_curriculum_weighting}, we explain a curriculum weighting strategy for multi-plane features.
It ensures that coordinate netework is learned first, followed by channel-wise disentanglement. It aims to provide a more diverse representation without the risk of overfitting where all channels exhibit identical expressions. 
Lastly, \autoref{subsec_loss_function} explains the loss function, which combines photometric loss and denoising multi-plane representations like Laplacian smoothing.

\paragraph{Nomenclature} This framework considers a camera with origin $o$ and a ray direction $d$. 
A ray $\textbf{r}$, composed of $n$ points, is constructed as $s_k = \textbf{o} + \tau_k \cdot \textbf{d}$, where $\tau_k \in \{\tau_1, \cdots, \tau_n\}$. The neural radiance field, parameterized by $\Theta$, predicts the color and density values $c_{\Theta}^k$, $\sigma_{\Theta}^k$ by volume rendering. 
The parameter $\Theta$ consists of MLPs including residual networks $\{\phi_l\}_L$ and multi-plane representations $\{\mathcal{M}$, $\mathcal{V}\}$. The feature corresponding to a ray sample $s_k$ by multi-plane representation is denoted as $f_k$. 
For a more detailed explanation of volume rendering and multi-plane representations, please refer to \autoref{app_background}.

\subsection{Architecture}
\label{subsec_architecture}
\begin{figure*}[t]{
    \centering
    \includegraphics[width=0.92\textwidth]{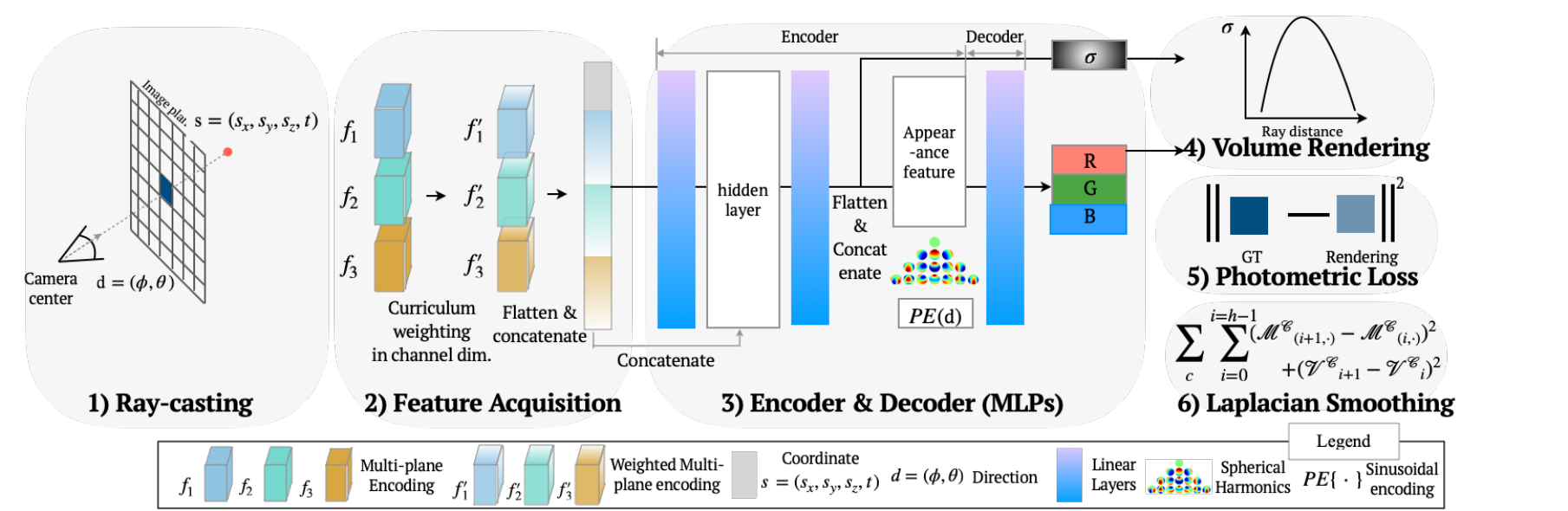}
    \caption{The schematic of the proposed method. The feature acquisition and encoder are discussed in \autoref{subsec_architecture} and \autoref{subsec_curriculum_weighting}. The loss function and regularization are described in \autoref{subsec_loss_function}.}
    \label{fig_Method_schematic}
}    
\vspace{-0.07in}
\end{figure*}
\vspace{-0.07in}
In high-level context, we replace sinusoidal encoding with multi-plane encoding while employing the architecture of the original NeRF \citep{mildenhall2021nerf}. The schematic of our architecture is illustrated in \autoref{fig_Method_schematic}.
A key aspect of our network architecture is the residual concatenation of coordinates value $s_k$ and multi-plane features $f_k$ across the first two blocks.
The residual connection accelerates the efficiency in responding to input values, so the network emphasizes the importance of coordinate networks. We employ ReLU activation $\textit{h}$ to lean toward low-frequency spectral bias \citep{rahaman2019spectral,tancik2020fourier}.
More specifically, the residual connection is defined as follows:
\begin{equation}
    \resizebox{0.905\columnwidth}{!}{$
        \begin{aligned}
        \phi_1(s_k, f_k) &= \textit{h}\big(W_1^2 \cdot \textit{h}(W_1^1 \cdot (s_k \oplus f_k) + b_1^1) + b_1^2\big) \\
        \phi_2(s_k, f_k, \phi_1) &= \textit{h}(W_2^2 \cdot \textit{h}(W_2^1 \cdot (s_k \oplus f_k \oplus \phi_1(s_k, f_k)) + b_2^1) + b_2^2
        \end{aligned}
    $}
    \label{eq_resnet}
\end{equation}

where, $\{W_l, b_l\}_{l=1}^L$ are the weights and biases of the $l$-layer MLP. In the residual connection blocks, when $l \leq 2$, the block includes two pairs of weights and biases. For $l>2$, the subsequent processes contain one pair of weights and biases. $\oplus$ indicates the concatenation of features. 

The output layers use different activations, such as the softplus function for density and the sigmoid function for color.
The proposed residual connection allows the network to robustly maintain low-frequency spectral bias from coordinate networks without interference from multi-plane features. 
Our empirical findings demonstrate that this operation promotes the disentanglement of two features, aligning with a spanning diverse spectrum.
A detailed analysis of this residual connection is provided in 
\autoref{subsec_ablation_study}
along with the performance gain by this architecture. 

\subsection{Curriculum Weighting for Multi-Plane Encoding}
\label{subsec_curriculum_weighting}
\vspace{-0.02in}
The architecture in the proposed method performs well in scenes with mild occlusion and less dynamic motion. 
However, it encounters challenges in severe ill-conditioned situations, such as heavy occlusion and rapid motion, as seen in the \texttt{drums} in the static NeRF and the \texttt{standup} in the dynamic NeRF. 
To alleviate this issue, we propose a curriculum weighting strategy for multi-plane encoding, aiming to manipulate the engagement of multi-plane features per training step. 
This approach trains the coordinate-based network first, followed by the subsequent training of multi-plane features. In this subsection, we denote $t$ as the training iteration. 
Technically, we introduce a weighting factor denoted as $\alpha(t)$ to control the degree of engagement of multi-plane features along multi-plane channel dimensions.
Here, $f_{i,k} \in \mathbb{R}^c$ represents the output of $i$-th plane encoding, and the weighting factor $\gamma(t) = \{\gamma_{1}(t), \cdots, \gamma_{c}(t) \} \in \mathbb{R}^c$ is defined as follows:
\vspace{-0.07in}
\begin{equation}
    \gamma_{j}(t)                                         = 
    \begin{cases}
                0                                       & \text{if } \alpha(t) \leq j \\
                \frac{1-\cos ((\alpha(t) - j) \pi)}{2}  & \text{if } 0 < \alpha(t) - j \leq 1 \\
                1                                       & \text{otherwise, } 
    \end{cases}
    \label{eq_curriculum_weighting}
    \vspace{-0.07in}
\end{equation}

where, $j \in \{1, \cdots, c\}$ is the index of channel dimension and $ \alpha(t) = c \cdot \nicefrac{\small(t-t_s\small)}{\small(t_e-t_s\small)} \in [ t_e, t_s ]$ is proportional to the number of training iterations $t$ in the scheduling interval $[t_s, t_e]$. 
The final features $f_i'$ are obtained by $f_i' = f_i \odot \gamma(t)$.
Hence, this weighting function is applied to each channel of multi-plane features. After reaching the last time-step of curriculum training, all channels of multi-plane features are fully engaged.
It is worth noting that this weighting function is similar to those used in previous works such as \citep{park2021nerfies, lin2021barf, yang2023freenerf, heo2023robust}. 
However, the critical difference is a channel-wise weighting function for multi-plane features. This function can be interpreted as gradually increasing the rank of multi-plane features from the perspective of tensor decomposition \citep{chen2022tensorf}.
Our experiments find that this strategy effectively prevents all channels of multi-plane features from converging to similar patterns. It even facilitates the flat representation of specific channels when they are redundant. This results in a more diverse spectrum and mitigating overfitting issues.

\subsection{Loss Function}
\label{subsec_loss_function}
We introduce a loss function that combines photometric loss and denoising multi-plane representations like Laplacian smoothing.
First, we define the photometric loss $\mathcal{L}_p$ as mean square errors between rendered color $\hat{\textbf{c}}(\textbf{r}$) and ground truth pixel color $\textbf{c}$, $\mathcal{L}_{p}(\Theta, \mathcal{M}, \mathcal{V}) = \sum_{r} \lVert \hat{\textbf{c}}(\textbf{r}; \Theta, \mathcal{M}, \mathcal{V}) - \textbf{c} \rVert^2$.
To tackle the ill-conditioned training problem in NeRFs arising from sparse-input situations, we apply Laplacian smoothing on both feature planes \citep{cao2023hexplane,fridovich2023k}. 
Laplacian smoothing $\mathcal{L}_l$ tends to excessively smooth signals, making them conform to global tendency rather than accurately local finer details \citep{sadhanala2017higher}. More information can be found in \autoref{app_hexplane}. Additionally, we regularize each plane feature using the L1 norm for the sparsity of multi-plane features. We use, $\lVert \mathcal{M} \rVert_1$ and $\lVert \mathcal{V} \rVert_1$ as $\sum_{i=1}^{i=3} \lVert M_i \rVert_1$ and $\sum_{i=1}^{i=3} \lVert V_i \rVert_1$ respectively. The entire loss function is defined as \autoref{eq_loss}. The only difference in the case of static NeRF comes from the dimension of $\mathcal{V}$.
Laplacian loss is not applied to $\mathcal{V}$; the rest of the details are the same as in the 4D case. The hyperparameters and implementation detail can be found in \autoref{app_implementation_details}.
\begin{equation}
    \resizebox{0.85\columnwidth}{!}{$
        \begin{aligned}
        \mathcal{L}(\Theta, &\mathcal{M}, \mathcal{V}) = \mathcal{L}_{p}(\Theta, \mathcal{M}, \mathcal{V})  +\\
        &
         \lambda_1 \sum_{i=1}^{3} \big( \mathcal{L}_{l}(M_i) + \lambda_2 \mathcal{L}_{l}(V_i) \big) + \lambda_3 \big( \lVert \mathcal{M} \rVert_1 + \lVert \mathcal{V} \rVert_1 \big)
        \end{aligned}
    $}
    \label{eq_loss}
    \vspace{-0.07in}
\end{equation}
While increasing the value of $\lambda_1$ allows the removal of floating artifacts by over-smoothing the multi-plane features, it creates undesirable deformation that looks authentic but is not present in the training data.
In addition, too high a value for $\lambda_1$ can increase learning instability due to excessive penalization. Therefore, finding a feasible weight demands extensive trial and error.
However, the proposed method is less sensitive to this issue as the coordinate network itself establishes a bottom line, while the multi-plane encoding compensates for high-frequency details.
We empirically validate this through our experiments.

\section{Experiments}
\label{sec_experiments}
\vspace{-0.02in}
\begin{figure*}[!ht]{
    \centering
    \includegraphics[width=0.85\textwidth]{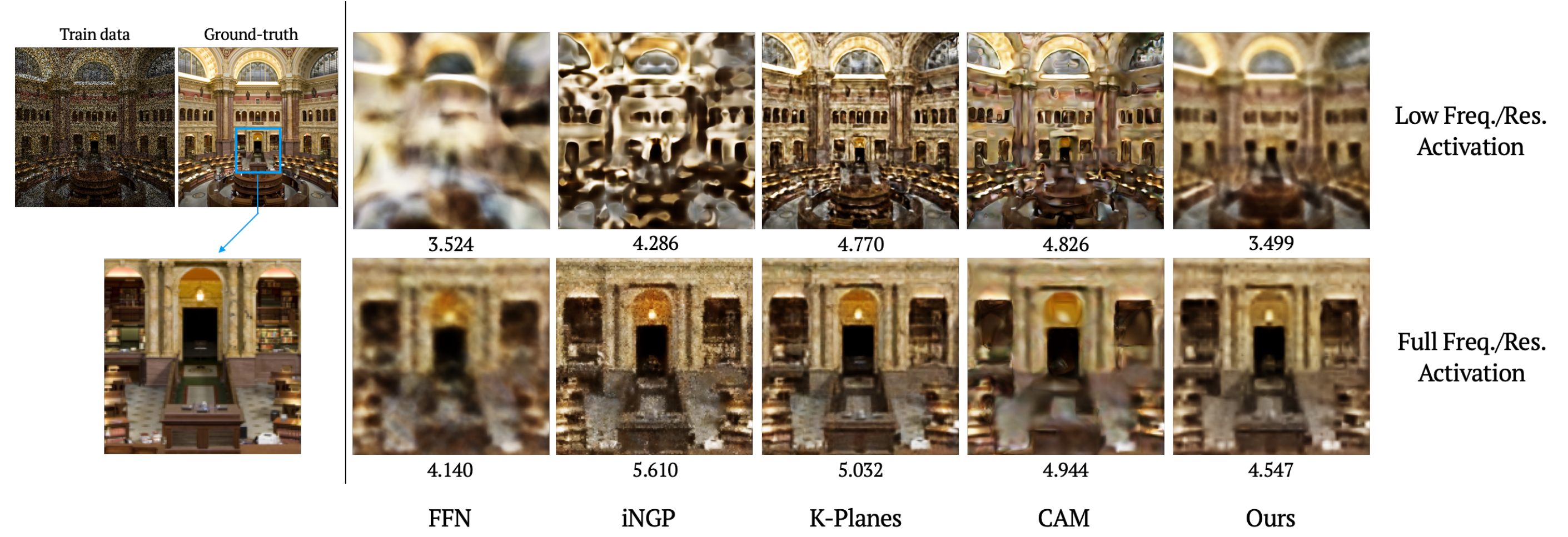}    
    \caption{Qualitative results on the image regression trained with a 50\% random mask. The first row displays rendered images using only low-frequency or resolution features, while the second row shows images engaged with the full range of features. 
    The numeric value indicates the average magnitude spectrum obtained from the Fourier transform.}
    \label{fig_exp_2d}
}    
\vspace{-0.07in}
\end{figure*}
In this section, we present our experiments designed to address three pivotal questions: 
\textit{1) Does existing multi-resolution parameterization and its simple integration with coordinate network adequately function low-frequency representations while producing clear scenes on sparse inputs?}
\textit{2) Does current regularization consistently maintain its effectiveness across various hyper-parameters and scenes, ensuring the capture of 3D coherence on sparse input data?}
\textit{3) Does the chemical integration of heterogeneous features, such as multiple planes and coordinates, substantially improve the performance of both static and dynamic NeRF?}

To answer those questions, we conduct vast experiments over scenarios of two sparse input cases: a few-shot static case and a 4-dimensional dynamic case. To provide a clearer insight into the role of low-frequency representation, we include 2D image regression as an illustrative example.
We also include ablation studies to substantiate the rationale behind the architectural choices. 
We choose the datasets as in-ward-facing object poses, as they are more likely to be occluded by the objects from various viewing locations than forward-facing poses.
For performance evaluation, we employ the PSNR metric to gauge image reconstruction quality. In addition, SSIM and LPIPS scores are reported to assess the perceptual quality of the rendered images.
Further experimental details are described in \autoref{app_experimental_Setting}.

\subsection{Motivation Example: Image Regression}
\vspace{-0.07in}
We start by demonstrating the diversity in spectrum information the proposed method possesses in $\{512\times512\}$ Image Regression tasks.
We compare our method with FFN \citep{tancik2020fourier} as a sinusoidal embedding and iNGP \citep{muller2022instant}, K-Planes \citep{fridovich2023k} as explicit parameterization. We also include CAM \citep{lee2023coordinate}, a similar approach that combines coordinate-based networks and grid-based representation for efficient parameterization.
\autoref{fig_exp_2d} shows how baselines handle low and high freqeuncy or resolution features by presenting the average magnitude spectrum by Fourier feature transform.
Specifically, in the case of low-frequency features or grids, FFN and CAM, which manipulate their spectrum via sinusoidal encoding, use only the lowest 26 frequencies (10\%) of total frequency range. K-Planes, modifed to use four multi-scale tensorial planes, employs only use the lowest resolutional plane with a scale of $32\times32$. iNGP, with 16 levels of spatial hash-grid, uses only the first two lowest resolutional grids. The proposed method only utilizes the coordinate-based MLP networks with four frequeices $\{2^i|i=0,1,2,3\}$. On the other hand, as the finest feature, K-Planes, CAM and Ours all adopt a $128\times128$ plane.

In \autoref{fig_exp_2d}, we observe that all baselines differs the average magnitude spectrum of rendered images between low and full feature engagement, though the extent of manignitude varies. Low feature engagement describes only low-frequency details, resulting in a minimum spectrum magnitude, whereas full feature engagement captures intrigate details, achieving the higest spectrum magnitude. Examining each instance, the sinusoidal method (FFN) faces an underfitting issue as it overly focuses on low frequency details. Explicit representations like iNGP and K-Planes, however, tend to interpret low resolution features focusing on high frequency details despite only low-resolution features are activated. 
Surprisingly, CAM, despite incoporating various spectral sinusodial embeddings, also struggles to capture low frequency details. 
This implies explicit representation such as grid or plane cannot effectively handle low frequency details without careful designs. 

In contrast, our proposed method balances low and high frequency spectral features, resulting in images that capture both types of details. Remarkably, the rendered images encompass a substantial spectral range, varying from 3.499 to 4.547. This range stands out as the most extensive deviation from the baselines, with the exception of iNGP. While iNGP exhibits the widest spectrum among baselines, the image with low resolution features does not adequately capture global reasoning. Subsequent experiments show that the effectiveness in utilizing low-frequency context for global consistency and then transitioning to high-frequency context to capture the finest details in both static and dynamic NeRF under sparse inputs. 

\subsection{Static Radiance Fields}
\label{subsec_3d_static_nerf}
\begin{figure*}[!t]{
    \centering
    \includegraphics[width=0.80\textwidth]{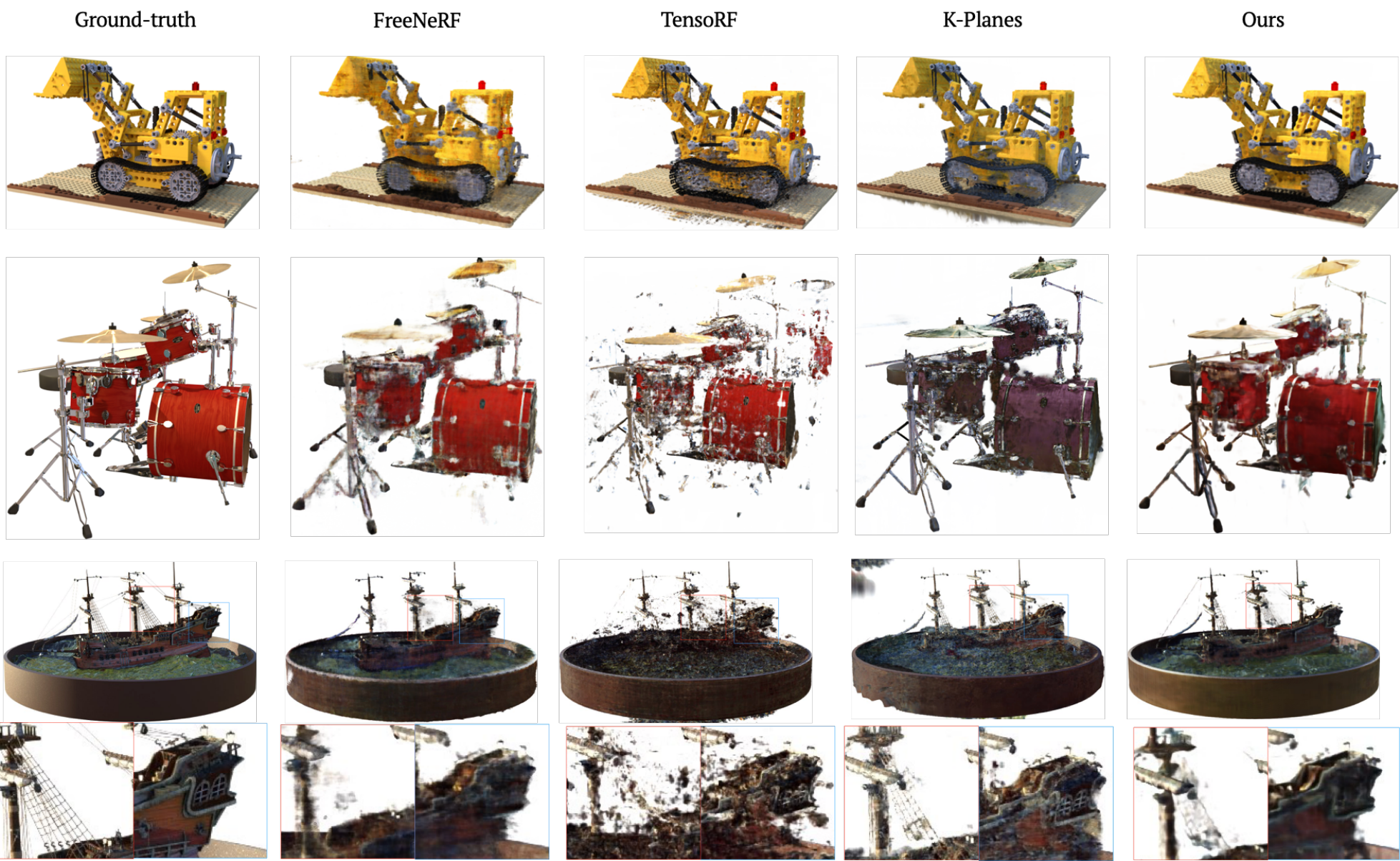}    
    \caption{Rendered images of \texttt{lego}, \texttt{drums} and \texttt{ship} cases in the static NeRF dataset by FreeNeRF, TensoRF, K-Planes and ours. The rendered images are $\{83, 129, 95\}$-th in the test set, respectively.}
    \label{fig_exp_static_nerf}
}    
\vspace{-0.07in}
\end{figure*}
\vspace{-0.07in}
\begin{table*}[!t]
    \caption{Result of evaluation statistics on the static NeRF datasets. We conduct five trials for each scene and report average scores. Average PSNR, SSIM, and LPIPS are calculated across all scenes. We indicates best performance as \textbf{bold} and second best as \underline{underline}}
    \label{table_static_nerf}
    \vspace{-0.07in}
    \centering
    \resizebox{1.00\textwidth}{!}{%
        \begin{tabular}{lccccccccccc}
        \toprule
        \multirow{2}{*}{Models} & \multicolumn{8}{c}{PSNR $\uparrow$}                                                  & \multirow{2}{*}{\begin{tabular}[c]{@{}l@{}}Avg. \\ PSNR\end{tabular} $\uparrow$}    & \multirow{2}{*}{\begin{tabular}[c]{@{}l@{}}Avg. \\ SSIM\end{tabular} $\uparrow$}    & \multirow{2}{*}{\begin{tabular}[c]{@{}l@{}}Avg. \\ LPIPS\end{tabular} $\downarrow$} \\
        \cmidrule(r){2-9}
                                        & chair             & drums             & ficus                & hotdog            & lego              & materials         & mic               & ship              &                               &                    &                       \\
        \midrule            
        Simplified\_NeRF                & 20.35             & 14.19             & \underline{21.63}    & 22.57             & 12.45             & 18.98             & 24.95             & 18.65             & 19.22                         & 0.827              & 0.265                 \\
        DietNeRF                        & 21.32             & 14.16             & 13.08                & 11.64             & 16.12             & 12.20             & 24.70             & 19.34             & 16.57                         & 0.746              & 0.333                 \\
        HALO                            & 24.77             & 18.67             & 21.42                & 10.22             & 22.41             & 21.00             & 24.94             & 21.67             & 20.64                         & 0.844              & 0.200                 \\
        FreeNeRF                        & 26.08             & \underline{19.99} & 18.43                & \underline{28.91} & 24.12             & \underline{21.74} & 24.89             & \underline{23.01} & 23.40                         & 0.877              & 0.121                 \\
        \midrule
        DVGO                            & 22.35             & 16.54             & 19.03                & 24.73             & 20.85             & 18.50             & 24.37             & 18.17             & 20.57                         & 0.829              & 0.145                 \\
        VGOS                            & 22.10             & 18.57             & 19.08                & 24.74             & 20.90             & 18.42             & 24.18             & 18.16             & 20.77                         & 0.838              & 0.143                 \\
        iNGP                            & 24.76             & 14.56             & 20.68                & 24.11             & 22.22             & 15.16             & 26.19             & 17.29             & 20.62                         & 0.828              & 0.184                 \\        
        TensoRF                         & 26.23             & 15.94             & 21.37                & 28.47             & 26.28             & 20.22             & 26.39             & 20.29             & 23.15                         & 0.864              & 0.129                 \\
        K-Planes                        & \underline{27.30} & \textbf{20.43}    & \textbf{23.82}       & 27.58             & \underline{26.52} & 19.66            & \textbf{27.30}    & 21.34             & \underline{24.24}             & \textbf{0.897}     & \textbf{0.085}        \\
        \midrule
        Ours                            & \textbf{28.02}    & 19.55             & 20.30                & \textbf{29.25}    & \textbf{26.73}    & \textbf{21.93}    & \underline{26.42} & \textbf{24.27}    & \textbf{24.56}                & \underline{0.896}  & \underline{0.092}     \\
        \bottomrule
    \end{tabular}
    }%
    \vspace{-0.07in}
\end{table*}
We conducted 3-dimensional static NeRF experiments on the NeRF-synthetic dataset to evaluate whether our model adequately captures both the global context of a scene and fine details without introducing undesirable artifacts under sparse input conditions. Consistent with prior studies such as  \citep{jain2021putting,yang2023freenerf}, we trained all models with 8 views.
We compare our proposed models with sinusoidal encoding methods; Simplified NeRF, DietNeRF \citep{jain2021putting}, HALO \citep{song2023harnessing} and FreeNeRF \citep{yang2023freenerf} and for explicit spatial parameterization methods; DVGO \citep{sun2022direct}, VGOS \citep{sun2023vgos}, iNGP \citep{muller2022instant}, TensoRF \citep{chen2022tensorf} and K-Planes \citep{fridovich2023k}.
For all baselines, we applied regularization techniques congruent with their inherent characteristics and configurations.

The quantitative rendering results are shown in \autoref{table_static_nerf} and \autoref{fig_exp_static_nerf}. More detailed numeric values are contained at \autoref{app_statistics_result}.
First, we observed that the proposed method outperforms the previous state-of-the-art method, FreeNeRF, in terms of both PSNR and perceptual quality. 
Sinusoidal encoding-based networks fail to capture high-frequency details and are prone to underfit in data with high-resolution structures, (\texttt{ficus, lego}).
In contrast, grid-based models show robust results in reconstructing high-frequency structures.
However, for data with a strong non-Lambertian effect (\texttt{drums, ship}), grid-based models tend to miss the global shape and are prone to overfit in high frequency.
Our proposed multi-plane encoding technique can exclusively capture fine-grained details while maintaining global shape learned by coordinate features, leading to more robust novel view synthesis in sparse-input scenarios. This phenomenon consistently occurs in real-world datasets. For more detail, please refer to \autoref{app_tanksandtemples}.
\subsection{Dynamic Radiance Fields}
\begin{table}[!t]
    \caption{Result of evaluation statistics on the D-NeRF datasets. HexPlane employs the weight of denoising regularization as $\lambda_1=0.01$ via grid-search. Average PSNR, SSIM, and LPIPS are calculated across all scenes. We indicate the best performance as \textbf{bold} for each case.}
    \label{table_dynamic_nerf}
    \centering
    \resizebox{0.92\columnwidth}{!}{%
        \begin{tabular}{clccc}
        \toprule
        \begin{tabular}[c]{@{}l@{}}Training \\ views\end{tabular} & Models & \begin{tabular}[c]{@{}l@{}}Avg. \\ PSNR\end{tabular} $\uparrow$ & \begin{tabular}[c]{@{}l@{}}Avg. \\ SSIM\end{tabular} $\uparrow$ & \begin{tabular}[c]{@{}l@{}}Avg. \\ LPIPS\end{tabular} $\downarrow$ \\                                                                                                       
        \midrule                                                                                    
        \multirow{3}{*}{15 poses}                                                   & HexPlane                  & 21.93                     & 0.921                      & 0.092                    \\                                                                                                                                                                                                                                                                                                                                
                                                                                    & K-Planes                  & 21.50                     & 0.922                      & \textbf{0.086}           \\
                                                                                    &    Ours                   & \textbf{22.30}            & \textbf{0.925}             & 0.087                    \\
        \midrule                                                                                                                                                                
        \multirow{3}{*}{20 poses}                                                   & HexPlane                  & 23.18                     & 0.929                      & 0.082                    \\
                                                                                    & K-Planes                  & 22.58                     & 0.931                      & \textbf{0.070}           \\
                                                                                    &     Ours                  & \textbf{23.93}            & \textbf{0.935}             & 0.072                    \\
        \midrule
        \multirow{3}{*}{25 poses}                                                   & HexPlane                  & 24.15                     & 0.935                      & 0.074                    \\
                                                                                    & K-Planes                  & 22.68                     & 0.929                      & 0.107                    \\
                                                                                    & Ours                      & \textbf{25.34}            & \textbf{0.941}             & \textbf{0.063}           \\
        \bottomrule
    \end{tabular}
    }%
    \vspace{-0.07in}
\end{table}
To demonstrate the robustness of the proposed model on more spare input cases, we conduct our experiences on the dynamic NeRF dataset \citep{pumarola2021d}.
This data set comprises monocular cameras of about 50-100 frames in duration and different inward-facing views for each timestep.
To verify a harsh situation, we also experimented with fewer frames $\{15, 20, 25\}$ sparse in both views and time aspects.
Each pose was sampled uniformly along the time axis for each scene. We compare our method with HexPlane \citep{cao2023hexplane} and K-Planes \citep{fridovich2023k}.

The observations made in \autoref{subsec_3d_static_nerf} are even more evident in the dynamic NeRFs.
The proposed method outperforms every setting of HexPlane in all metrics in the D-NeRFs, as shown in \autoref{table_dynamic_nerf}. 
HexPlane discretizes the continuous time axis into finite bins, making it less responsive to the time-variant motion of objects when the available training poses are sparse. 
In contrast, the proposed method can capture the time-variant motion of objects by harnessing the coordinate-based networks first, with multi-plane encoding supplementing the remaining details. 

\subsection{Stability in Sparse-Input NeRFs}
\label{subsec_stability}
In sparse-input NeRFs, stability is defined as the ability to counteract overfitting. We measure stability by evaluating the minimal performance discrepancy between test viewpoints adjacent to and not adjacent to the training views. Specifically, we examine the variance of PSNR across all test viewpoints in the static NeRF dataset. The total variance of PSNR across all images is calculated using 8,000 images from 8 scenes, each with 200 test viewpoints and five trials.
\begin{table}[!t]
    \caption{Variance of PSNR($\downarrow$) on the static NeRF datasets.}
    \label{table_psnr_variance}
    \vspace{-0.07in}
    \centering
    \resizebox{0.92\columnwidth}{!}{%
    \begin{tabular}{ccccc}
    \toprule
    FreeNeRF & iNGP  & TensoRF & K-Planes & Ours  \\
    \midrule
    17.31    & 23.95 & 23.22   & 19.61    & 18.23 \\
    \bottomrule
    \end{tabular}
    }%
    \vspace{-0.07in}
\end{table}
FreeNeRF, which uses MLP and sinusoidal encoding, shows the lowest variance among baselines. Spatially explicit methods like iNGP and TensoRF exhibit significant variances due to their tendency to overfit the training views. While K-Planes reduces instability compared to these methods, its variances still do not match ours. Quantitatively, our method achieves comparable results to FreeNeRF. 
However, as shown in \autoref{table_psnr_variance}, FreeNeRF generally lacks reconstruction performance. Additionally, K-Planes struggles with reconstructing specific scenes, such as ship. In contrast, our method consistently reconstructs all scenes with high quality, avoiding significant degradation. This is also evident in \autoref{fig_exp_static_nerf}. While FreeNeRF exhibits blurry details and K-Planes displays strange color distortion, our method shows the cleanest results without noticeable distortion or artifacts. Considering that our method shows low variances and achieves the highest PSNR, we emphasize the distinction of our approach in terms of both stability and superior capability.

\subsection{Ablation Study}
\label{subsec_ablation_study}
\begin{table}[!b]
    \caption{Performance evaluation by varying residual connection candidates on the static NeRF dataset with 8 views}
    \label{tab_residual_network}
    \centering
    \resizebox{0.70\columnwidth}{!}{%
    \begin{tabular}{lccc}
        \toprule
        Model & \begin{tabular}[c]{@{}l@{}}Avg. \\ PSNR\end{tabular} $\uparrow$ & \begin{tabular}[c]{@{}l@{}}Avg. \\ SSIM\end{tabular} $\uparrow$ & \begin{tabular}[c]{@{}l@{}}Avg. \\ LPIPS\end{tabular} $\downarrow$ \\
        \midrule
        Ours   & 24.74     & 0.898    & 0.089     \\
        Type 1 & 18.77     & 0.844    & 0.179     \\
        Type 2 & 19.23     & 0.848    & 0.171     \\
        Type 3 & 19.07     & 0.843    & 0.175    \\
        \bottomrule
    \end{tabular}
    }%
    \vspace{-0.07in}
\end{table}
To validate the effectiveness of the proposed architecture, we analyze several types of candidates with respect to residual connections. We consider three candidates: Type 1, where skip connection lies at every layer, Type 2, which has no skip connection; and the last one, where only the coordinate value $s_k$ is residual concatenated. The quantitative result is presented in \autoref{tab_residual_network}. We observe that the strainghtforward implementation of residual connection leads to ineffective training for sprase inputs. However, the proposed method gains remarkble performance gap than others, highlighting the necessity of careful desing for handling two heterogeneous features. For more information, please refer to \autoref{app_analysis_encoding}.
\begin{table}[!t]
    \caption{Performance evaluation of the D-NeRF dataset with training steps up to 15,000 and 25 poses. The rendering time is assessed using 20 poses.}  
    \label{tab_training_params_dynamic}
    \centering
    \resizebox{0.98\columnwidth}{!}{%
        \begin{tabular}{lcccc}
        \toprule
        \multicolumn{1}{c}{\begin{tabular}[c]{@{}c@{}}Model \\ \textbf{Name}\end{tabular}} & \begin{tabular}[c]{@{}c@{}}\# Params\\ {[}M{]}\end{tabular} & \begin{tabular}[c]{@{}c@{}}Avg. \\ PSNR\end{tabular} & \begin{tabular}[c]{@{}c@{}}Avg. Train \\ Time [min]\end{tabular} & \begin{tabular}[c]{@{}c@{}}Avg. Render \\ Time [min]\end{tabular} \\
        \midrule
        K-Planes \footnotesize{(3*32)}                                         & 18.6M                                                       & 23.85                                                & 18.93                                                         & 0.83                                                         \\
        K-Planes \footnotesize{(3*4)}                                         & 1.9M                                                        & 23.41                                                & 13.29                                                         & 0.78                                                         \\
        HexPlane \footnotesize{(72)}                                           & 9.7M                                                        & 24.00                                                & 6.78                                                          & 0.60                                                         \\
        HexPlane \footnotesize{(6)}                                            & 0.8M                                                        & 22.08                                                & 6.38                                                          & 0.68                                                              \\
        Ours \footnotesize{(48)}                                               & 3.4M                                                        & 25.17                                                & 12.22                                                         & 2.14                                                        \\
        Ours \footnotesize{(12)}                                               & 1.0M                                                        & 25.10                                                & 8.77                                                          & 1.73                                                       \\
        \bottomrule
        \multicolumn{5}{l}{* Numbers in brackets are the channel dimensions of each multi-plane.}
        \end{tabular}
    }%
    \vspace{-0.07in}
\end{table}
In terms of insensitivity to hyper-parameters, we evaluate explicit parameterization methods on dynamic NeRFs by reducing the channel dimension as shown in \autoref{tab_training_params_dynamic}. 
While baselines exhibit a performance decrease, the proposed method preserves performance even with only 30\% of number of parameter used. Moreover, the reduced model with only 1.0M parameters surpasses the other full parameterized baselines. This achievement is attributed to the disentanglement of two heterogeneous representations, as redundant multi-plane for low-resolution features are replaced with the coordinate network. 
In addition, we explore the sensitivity of regularization in  \autoref{table_denoising_regularization}. It demonstrates that the proposed method maintains near-optimal performance across all hyper parameters. In one case, TensoRF with $\lambda_1=0.001$ fortunately performs the best at 24.98, but it fails to converge when $\lambda_1$ exceeds 0.01. This indicates its sensitivity to regularization values. K-Plane appears to be more stable, but its overall performance lags behind the proposed method. Moreover, excessive regularization can introduce undesirable modification such as color disturbances. The detailed experimental results are included in \autoref{app_varying_lambda1}.
\begin{table}[!t]
    \caption{Average PSNR across all scenes varying denoising regularization $\lambda_1$. The hyphen indicates not converged.}
    \label{table_denoising_regularization}
    \centering
    \resizebox{1.00\columnwidth}{!}{%
    \begin{tabular}{ccccccc}
    \toprule
    \multicolumn{1}{c}{\multirow{2}{*}{$\lambda_1$}}          & \multicolumn{3}{c}{Static NeRF (8 views)} & \multicolumn{3}{c}{D-NeRF (25 views)} \\
    \cmidrule(r){2-7}
                                    & TensoRF        & K-Planes     & ours             & HexPlane       & K-Planes      & ours        \\
    \midrule                    
    0.0001                          & 24.10          & 24.31        & 23.68            & 22.83          & 24.32         & 24.67       \\
    0.001                           & 24.98          & 24.28        & 24.47            & 23.86          & 24.01         & 25.38       \\
    0.01                            & -              & 24.28        & 24.55            & 24.15          & 24.02         & 25.74       \\
    0.1                             & -              & 23.64        & 24.23            & 23.46          & 23.55         & 25.84       \\
    1.0                             & -              & 22.05        & 22.99            & 21.95          & 22.62         & 25.42       \\
    \bottomrule
    \end{tabular}
    }%
\vspace{-0.07in}
\end{table}
\begin{table}[!t]
\caption{Quantitative results between activation and absence of curriculum weighting for multi-plane encoding. Mean and variance of PSNR are presented for each scene. We conduct three trials using random seeds to measure average PSNR and its variance }
\label{table_comparision_between_cl}
\centering
\resizebox{1.00\columnwidth}{!}{%
\begin{tabular}{clccc|cccc}
\toprule
\multirow{2}{*}{}                                                  & \multirow{2}{*}{Metric}                                         & \multicolumn{3}{c|}{Static NeRF} & \multicolumn{4}{c}{D-NeRF}                                                                                                 \\
\cmidrule(r){3-9}
\multicolumn{2}{c}{}                                                          & Drums   & Lego    & Mic    & \begin{tabular}[c]{@{}c@{}}Hell\\ Warrior\end{tabular} & Lego  & \begin{tabular}[c]{@{}c@{}}Stand\\ up\end{tabular} & Trex  \\
\midrule

\multirow{2}{*}{CL}                                                & Mean($\uparrow$)     & 20.20   & 26.84   & 26.61  & 19.46                                                  & 24.05 & 26.22                                              & 26.70 \\
                                                                   & Variance($\downarrow$) & 2.31    & 7.20    & 6.28   & 2.05                                                   & 2.93  & 5.42                                               & 3.22  \\
\midrule
\multirow{2}{*}{\begin{tabular}[c]{@{}c@{}}Non-\\ CL\end{tabular}} & Mean($\uparrow$)     & 19.84   & 26.54   & 26.49  & 19.83                                                  & 23.94 & 25.59                                              & 26.63 \\
                                                                   & Variance($\downarrow$) & 2.55    & 6.26    & 9.16   & 2.80                                                   & 4.15  & 4.99                                               & 3.40 \\
\bottomrule
\end{tabular}
}%
\vspace{-0.07in}
\end{table}
\vspace{-0.07in}

To validate the curriculum weighting, we conduct a comparison between the proposed method and the same architecture that does not utilize progressive training. We choose \{\texttt{Lego}, \texttt{Drums}, \texttt{Mic}\} from the static NeRF, and \{\texttt{Hellwarrior}, \texttt{Lego}, \texttt{Standup}, \texttt{Trex}\} from the dynamic NeRF where this weighting is applied (See \autoref{app_implementation_details}). We denote \textit{CL} as the activation of progressive training and \textit{Non\_CL} as its absence. 
In static NeRFs, we observe that \textit{CL} consistently has a positive impact on performance improvement to Average PSNR, despite the fact that their improvement on reconstruction is minor, ranging from 0.2 to 0.4 in all cases. For variances, \{\texttt{drums}, \texttt{lego}\} show no significant difference, but the \texttt{Mic} result indicates that \textit{CL} mitigates instability, with variances decreasing by 2.8. This reduction means significantly less discrepancy between images adjacent and not adjacent to training views.
The effectiveness of progressive training is more pronounced in dynamic NeRFs. While it does not provide significant improvement in the \texttt{hellwarrior} case, it evidently enhances performance in the \texttt{standup} case, leading to 0.6 increase in average PSNR. In terms of stability, we observe that most cases are less sensitive to overfitting with progressive training. Although variance slightly increases in the \texttt{standup}, it is not significant issue considering average PSNR improvement. In summary, progressive training influences on either performance improvement or mitigation of instability in sparse-input NeRF by gradually engaging multi-plane channels. This allows the initial channels to learn global details, while later channels focus more on finer details. The detailed explanation, including graphics, is provided in \autoref{app_disentangle}.
\section{Conclusion}
\label{sec_conclusion}
\vspace{-0.07in}
In this paper, we introduce refined tensorial radiance fields that seamlessly incorporate coordinate networks. 
The coordinate network enables the capture of global context, such as object shapes in the static NeRF and dynamic motions in the dynamic NeRF dataset. This property allows multi-plane encoding to focus on describing the finest details. 
Through extensive experiments, we demonstrate that the proposed method consistently outperforms the baselines and their regularization in the few-shot regime. Notably, the proposed method exhibits strong stability, showing less discrepancy between images adjacent and non-adjancet to training views. Additionally, it preserves performance even with a reduced number of parameters. 

\section*{Impact Statement}
\label{sec:Ethics}
Novel view synthesis is a task to understand the shape and appearance of objects and scenes from a sparse set of images or video.
Our model, in particular, can reconstruct fine-detailed 3D shapes with an accurate appearance just from given fewer inputs, both in static and dynamic scenes.

Like previous works, our model can obtain fine reconstruction results only if sufficiently distributed views are given.
Recovering high-fidelity 3D shapes and appearances of objects from fewer inputs offers numerous practical applications. 
However, it also introduces potential drawbacks, such as the leading to the creation of potentially misleading media or potentially facilitating design theft, by duplicating physical objects.

\section*{Reproducibility Statement}
\label{sec:reproducibility}
For reproducibility, our code is available at \url{https://github.com/MingyuKim87/SynergyNeRF}. Both training and evaluation codes are included for convenience. Qualitative results can be found on our \href{https://mingyukim87.github.io/SynergyNeRF/}{project page.}

\section*{Acknowledgements}
\label{sec:Ack}
This work was supported by Institute of Information \& communications Technology Planning \& Evaluation (IITP) grant funded by the Korea government(MSIT) [No.2022-0-00641, XVoice: Multi-Modal Voice Meta Learning]. A portion of this work was carried out during an internship at NAVER AI Lab. We also extend our gratitude to ACTNOVA for providing the computational resources required.

\nocite{langley00}

\bibliography{icml2024}

\begin{thebibliography}{47}
\providecommand{\natexlab}[1]{#1}
\providecommand{\url}[1]{\texttt{#1}}
\expandafter\ifx\csname urlstyle\endcsname\relax
  \providecommand{\doi}[1]{doi: #1}\else
  \providecommand{\doi}{doi: \begingroup \urlstyle{rm}\Url}\fi

\bibitem[Barron et~al.(2021)Barron, Mildenhall, Tancik, Hedman, Martin-Brualla,
  and Srinivasan]{barron2021mip}
Barron, J.~T., Mildenhall, B., Tancik, M., Hedman, P., Martin-Brualla, R., and
  Srinivasan, P.~P.
\newblock Mip-nerf: A multiscale representation for anti-aliasing neural
  radiance fields.
\newblock In \emph{Proceedings of the IEEE/CVF International Conference on
  Computer Vision}, pp.\  5855--5864, 2021.

\bibitem[Barron et~al.(2022)Barron, Mildenhall, Verbin, Srinivasan, and
  Hedman]{barron2022mip}
Barron, J.~T., Mildenhall, B., Verbin, D., Srinivasan, P.~P., and Hedman, P.
\newblock Mip-nerf 360: Unbounded anti-aliased neural radiance fields.
\newblock In \emph{Proceedings of the IEEE/CVF Conference on Computer Vision
  and Pattern Recognition}, pp.\  5470--5479, 2022.

\bibitem[Cao \& Johnson(2023)Cao and Johnson]{cao2023hexplane}
Cao, A. and Johnson, J.
\newblock Hexplane: A fast representation for dynamic scenes.
\newblock In \emph{Proceedings of the IEEE/CVF Conference on Computer Vision
  and Pattern Recognition}, pp.\  130--141, 2023.

\bibitem[Chan et~al.(2022)Chan, Lin, Chan, Nagano, Pan, De~Mello, Gallo,
  Guibas, Tremblay, Khamis, et~al.]{chan2022efficient}
Chan, E.~R., Lin, C.~Z., Chan, M.~A., Nagano, K., Pan, B., De~Mello, S., Gallo,
  O., Guibas, L.~J., Tremblay, J., Khamis, S., et~al.
\newblock Efficient geometry-aware 3d generative adversarial networks.
\newblock In \emph{Proceedings of the IEEE/CVF Conference on Computer Vision
  and Pattern Recognition}, pp.\  16123--16133, 2022.

\bibitem[Chen et~al.(2021)Chen, Xu, Zhao, Zhang, Xiang, Yu, and
  Su]{chen2021mvsnerf}
Chen, A., Xu, Z., Zhao, F., Zhang, X., Xiang, F., Yu, J., and Su, H.
\newblock Mvsnerf: Fast generalizable radiance field reconstruction from
  multi-view stereo.
\newblock In \emph{Proceedings of the IEEE/CVF International Conference on
  Computer Vision}, pp.\  14124--14133, 2021.

\bibitem[Chen et~al.(2022)Chen, Xu, Geiger, Yu, and Su]{chen2022tensorf}
Chen, A., Xu, Z., Geiger, A., Yu, J., and Su, H.
\newblock Tensorf: Tensorial radiance fields.
\newblock In \emph{European Conference on Computer Vision}, pp.\  333--350.
  Springer, 2022.

\bibitem[Deng et~al.(2022)Deng, Liu, Zhu, and Ramanan]{deng2022depth}
Deng, K., Liu, A., Zhu, J.-Y., and Ramanan, D.
\newblock Depth-supervised nerf: Fewer views and faster training for free.
\newblock In \emph{Proceedings of the IEEE/CVF Conference on Computer Vision
  and Pattern Recognition}, pp.\  12882--12891, 2022.

\bibitem[Fathony et~al.(2021)Fathony, Sahu, Willmott, and
  Kolter]{fathony2021multiplicative}
Fathony, R., Sahu, A.~K., Willmott, D., and Kolter, J.~Z.
\newblock Multiplicative filter networks.
\newblock In \emph{International Conference on Learning Representations}, 2021.
\newblock URL \url{https://openreview.net/forum?id=OmtmcPkkhT}.

\bibitem[Fridovich-Keil et~al.(2023)Fridovich-Keil, Meanti, Warburg, Recht, and
  Kanazawa]{fridovich2023k}
Fridovich-Keil, S., Meanti, G., Warburg, F.~R., Recht, B., and Kanazawa, A.
\newblock K-planes: Explicit radiance fields in space, time, and appearance.
\newblock In \emph{Proceedings of the IEEE/CVF Conference on Computer Vision
  and Pattern Recognition}, pp.\  12479--12488, 2023.

\bibitem[Gupta et~al.(2023)Gupta, Xiong, Nie, Jones, and
  O{\u{g}}uz]{gupta20233dgen}
Gupta, A., Xiong, W., Nie, Y., Jones, I., and O{\u{g}}uz, B.
\newblock 3dgen: Triplane latent diffusion for textured mesh generation.
\newblock \emph{arXiv preprint arXiv:2303.05371}, 2023.

\bibitem[He et~al.(2016)He, Zhang, Ren, and Sun]{he2016identity}
He, K., Zhang, X., Ren, S., and Sun, J.
\newblock Identity mappings in deep residual networks.
\newblock In \emph{Computer Vision--ECCV 2016: 14th European Conference,
  Amsterdam, The Netherlands, October 11--14, 2016, Proceedings, Part IV 14},
  pp.\  630--645. Springer, 2016.

\bibitem[Heo et~al.(2023)Heo, Kim, Lee, Lee, Kim, Kim, and Kim]{heo2023robust}
Heo, H., Kim, T., Lee, J., Lee, J., Kim, S., Kim, H.~J., and Kim, J.-H.
\newblock Robust camera pose refinement for multi-resolution hash encoding.
\newblock In \emph{International Conference on Machine Learning}. PMLR, 2023.

\bibitem[Jain et~al.(2021)Jain, Tancik, and Abbeel]{jain2021putting}
Jain, A., Tancik, M., and Abbeel, P.
\newblock Putting nerf on a diet: Semantically consistent few-shot view
  synthesis.
\newblock In \emph{Proceedings of the IEEE/CVF International Conference on
  Computer Vision}, pp.\  5885--5894, 2021.

\bibitem[Kingma \& Ba(2015)Kingma and Ba]{KingBa15}
Kingma, D. and Ba, J.
\newblock Adam: A method for stochastic optimization.
\newblock In \emph{International Conference on Learning Representations
  (ICLR)}, San Diega, CA, USA, 2015.

\bibitem[Knapitsch et~al.(2017)Knapitsch, Park, Zhou, and
  Koltun]{Knapitsch2017}
Knapitsch, A., Park, J., Zhou, Q.-Y., and Koltun, V.
\newblock Tanks and temples: Benchmarking large-scale scene reconstruction.
\newblock \emph{ACM Transactions on Graphics}, 36\penalty0 (4), 2017.

\bibitem[Lee et~al.(2023)Lee, Rho, Nam, Ko, and Park]{lee2023coordinate}
Lee, J.~C., Rho, D., Nam, S., Ko, J.~H., and Park, E.
\newblock Coordinate-aware modulation for neural fields.
\newblock \emph{arXiv preprint arXiv:2311.14993}, 2023.

\bibitem[Lin et~al.(2021)Lin, Ma, Torralba, and Lucey]{lin2021barf}
Lin, C.-H., Ma, W.-C., Torralba, A., and Lucey, S.
\newblock Barf: Bundle-adjusting neural radiance fields.
\newblock In \emph{Proceedings of the IEEE/CVF International Conference on
  Computer Vision}, pp.\  5741--5751, 2021.

\bibitem[Lindell et~al.(2022)Lindell, Van~Veen, Park, and
  Wetzstein]{lindell2022bacon}
Lindell, D.~B., Van~Veen, D., Park, J.~J., and Wetzstein, G.
\newblock Bacon: Band-limited coordinate networks for multiscale scene
  representation.
\newblock In \emph{Proceedings of the IEEE/CVF conference on computer vision
  and pattern recognition}, pp.\  16252--16262, 2022.

\bibitem[Liu et~al.(2020)Liu, Gu, Zaw~Lin, Chua, and Theobalt]{liu2020neural}
Liu, L., Gu, J., Zaw~Lin, K., Chua, T.-S., and Theobalt, C.
\newblock Neural sparse voxel fields.
\newblock \emph{Advances in Neural Information Processing Systems},
  33:\penalty0 15651--15663, 2020.

\bibitem[Martel et~al.(2021)Martel, Lindell, Lin, Chan, Monteiro, and
  Wetzstein]{martel2021acorn}
Martel, J.~N., Lindell, D.~B., Lin, C.~Z., Chan, E.~R., Monteiro, M., and
  Wetzstein, G.
\newblock Acorn: adaptive coordinate networks for neural scene representation.
\newblock \emph{ACM Transactions on Graphics (TOG)}, 40\penalty0 (4):\penalty0
  1--13, 2021.

\bibitem[Martin-Brualla et~al.(2021)Martin-Brualla, Radwan, Sajjadi, Barron,
  Dosovitskiy, and Duckworth]{martin2021nerf}
Martin-Brualla, R., Radwan, N., Sajjadi, M.~S., Barron, J.~T., Dosovitskiy, A.,
  and Duckworth, D.
\newblock Nerf in the wild: Neural radiance fields for unconstrained photo
  collections.
\newblock In \emph{Proceedings of the IEEE/CVF Conference on Computer Vision
  and Pattern Recognition}, pp.\  7210--7219, 2021.

\bibitem[Mihajlovic et~al.(2023)Mihajlovic, Prokudin, Pollefeys, and
  Tang]{mihajlovic2023resfields}
Mihajlovic, M., Prokudin, S., Pollefeys, M., and Tang, S.
\newblock Resfields: Residual neural fields for spatiotemporal signals.
\newblock \emph{arXiv preprint arXiv:2309.03160}, 2023.

\bibitem[Mildenhall et~al.(2021)Mildenhall, Srinivasan, Tancik, Barron,
  Ramamoorthi, and Ng]{mildenhall2021nerf}
Mildenhall, B., Srinivasan, P.~P., Tancik, M., Barron, J.~T., Ramamoorthi, R.,
  and Ng, R.
\newblock Nerf: Representing scenes as neural radiance fields for view
  synthesis.
\newblock \emph{Communications of the ACM}, 65\penalty0 (1):\penalty0 99--106,
  2021.

\bibitem[M{\"u}ller et~al.(2022)M{\"u}ller, Evans, Schied, and
  Keller]{muller2022instant}
M{\"u}ller, T., Evans, A., Schied, C., and Keller, A.
\newblock Instant neural graphics primitives with a multiresolution hash
  encoding.
\newblock \emph{ACM Transactions on Graphics (ToG)}, 41\penalty0 (4):\penalty0
  1--15, 2022.

\bibitem[Niemeyer et~al.(2022)Niemeyer, Barron, Mildenhall, Sajjadi, Geiger,
  and Radwan]{niemeyer2022regnerf}
Niemeyer, M., Barron, J.~T., Mildenhall, B., Sajjadi, M.~S., Geiger, A., and
  Radwan, N.
\newblock Regnerf: Regularizing neural radiance fields for view synthesis from
  sparse inputs.
\newblock In \emph{Proceedings of the IEEE/CVF Conference on Computer Vision
  and Pattern Recognition}, pp.\  5480--5490, 2022.

\bibitem[Park et~al.(2021)Park, Sinha, Barron, Bouaziz, Goldman, Seitz, and
  Martin-Brualla]{park2021nerfies}
Park, K., Sinha, U., Barron, J.~T., Bouaziz, S., Goldman, D.~B., Seitz, S.~M.,
  and Martin-Brualla, R.
\newblock Nerfies: Deformable neural radiance fields.
\newblock In \emph{Proceedings of the IEEE/CVF International Conference on
  Computer Vision}, pp.\  5865--5874, 2021.

\bibitem[Peng et~al.(2023)Peng, Yan, Shuai, Bao, and
  Zhou]{peng2023representing}
Peng, S., Yan, Y., Shuai, Q., Bao, H., and Zhou, X.
\newblock Representing volumetric videos as dynamic mlp maps.
\newblock In \emph{Proceedings of the IEEE/CVF Conference on Computer Vision
  and Pattern Recognition}, pp.\  4252--4262, 2023.

\bibitem[Pumarola et~al.(2021)Pumarola, Corona, Pons-Moll, and
  Moreno-Noguer]{pumarola2021d}
Pumarola, A., Corona, E., Pons-Moll, G., and Moreno-Noguer, F.
\newblock D-nerf: Neural radiance fields for dynamic scenes.
\newblock In \emph{Proceedings of the IEEE/CVF Conference on Computer Vision
  and Pattern Recognition}, pp.\  10318--10327, 2021.

\bibitem[Rahaman et~al.(2019)Rahaman, Baratin, Arpit, Draxler, Lin, Hamprecht,
  Bengio, and Courville]{rahaman2019spectral}
Rahaman, N., Baratin, A., Arpit, D., Draxler, F., Lin, M., Hamprecht, F.,
  Bengio, Y., and Courville, A.
\newblock On the spectral bias of neural networks.
\newblock In \emph{International Conference on Machine Learning}, pp.\
  5301--5310. PMLR, 2019.

\bibitem[Ramasinghe et~al.(2022)Ramasinghe, MacDonald, and
  Lucey]{ramasinghe2022frequency}
Ramasinghe, S., MacDonald, L.~E., and Lucey, S.
\newblock On the frequency-bias of coordinate-mlps.
\newblock \emph{Advances in Neural Information Processing Systems},
  35:\penalty0 796--809, 2022.

\bibitem[Roessle et~al.(2022)Roessle, Barron, Mildenhall, Srinivasan, and
  Nie{\ss}ner]{roessle2022dense}
Roessle, B., Barron, J.~T., Mildenhall, B., Srinivasan, P.~P., and Nie{\ss}ner,
  M.
\newblock Dense depth priors for neural radiance fields from sparse input
  views.
\newblock In \emph{Proceedings of the IEEE/CVF Conference on Computer Vision
  and Pattern Recognition}, pp.\  12892--12901, 2022.

\bibitem[Sadhanala et~al.(2017)Sadhanala, Wang, Sharpnack, and
  Tibshirani]{sadhanala2017higher}
Sadhanala, V., Wang, Y.-X., Sharpnack, J.~L., and Tibshirani, R.~J.
\newblock Higher-order total variation classes on grids: Minimax theory and
  trend filtering methods.
\newblock \emph{Advances in Neural Information Processing Systems}, 30, 2017.

\bibitem[Shekarforoush et~al.(2022)Shekarforoush, Lindell, Fleet, and
  Brubaker]{shekarforoush2022residual}
Shekarforoush, S., Lindell, D., Fleet, D.~J., and Brubaker, M.~A.
\newblock Residual multiplicative filter networks for multiscale
  reconstruction.
\newblock \emph{Advances in Neural Information Processing Systems},
  35:\penalty0 8550--8563, 2022.

\bibitem[Sitzmann et~al.(2020)Sitzmann, Martel, Bergman, Lindell, and
  Wetzstein]{sitzmann2020implicit}
Sitzmann, V., Martel, J., Bergman, A., Lindell, D., and Wetzstein, G.
\newblock Implicit neural representations with periodic activation functions.
\newblock \emph{Advances in neural information processing systems},
  33:\penalty0 7462--7473, 2020.

\bibitem[Song et~al.(2023)Song, Li, Gong, Chen, Chen, Xu, and
  Yuan]{song2023harnessing}
Song, L., Li, Z., Gong, X., Chen, L., Chen, Z., Xu, Y., and Yuan, J.
\newblock Harnessing low-frequency neural fields for few-shot view synthesis.
\newblock \emph{arXiv preprint arXiv:2303.08370}, 2023.

\bibitem[Sun et~al.(2022)Sun, Sun, and Chen]{sun2022direct}
Sun, C., Sun, M., and Chen, H.-T.
\newblock Direct voxel grid optimization: Super-fast convergence for radiance
  fields reconstruction.
\newblock In \emph{Proceedings of the IEEE/CVF Conference on Computer Vision
  and Pattern Recognition}, pp.\  5459--5469, 2022.

\bibitem[Sun et~al.(2023)Sun, Zhang, Chen, Li, Ji, Zhao, and Xing]{sun2023vgos}
Sun, J., Zhang, Z., Chen, J., Li, G., Ji, B., Zhao, L., and Xing, W.
\newblock Vgos: Voxel grid optimization for view synthesis from sparse inputs.
\newblock \emph{arXiv preprint arXiv:2304.13386}, 2023.

\bibitem[Tancik et~al.(2020)Tancik, Srinivasan, Mildenhall, Fridovich-Keil,
  Raghavan, Singhal, Ramamoorthi, Barron, and Ng]{tancik2020fourier}
Tancik, M., Srinivasan, P., Mildenhall, B., Fridovich-Keil, S., Raghavan, N.,
  Singhal, U., Ramamoorthi, R., Barron, J., and Ng, R.
\newblock Fourier features let networks learn high frequency functions in low
  dimensional domains.
\newblock \emph{Advances in Neural Information Processing Systems},
  33:\penalty0 7537--7547, 2020.

\bibitem[Truong et~al.(2023)Truong, Rakotosaona, Manhardt, and
  Tombari]{truong2023sparf}
Truong, P., Rakotosaona, M.-J., Manhardt, F., and Tombari, F.
\newblock Sparf: Neural radiance fields from sparse and noisy poses.
\newblock In \emph{Proceedings of the IEEE/CVF Conference on Computer Vision
  and Pattern Recognition}, pp.\  4190--4200, 2023.

\bibitem[Wang et~al.(2023{\natexlab{a}})Wang, Hu, He, Wang, Yu, Tuytelaars, Xu,
  and Wu]{wang2023neural}
Wang, L., Hu, Q., He, Q., Wang, Z., Yu, J., Tuytelaars, T., Xu, L., and Wu, M.
\newblock Neural residual radiance fields for streamably free-viewpoint videos.
\newblock In \emph{Proceedings of the IEEE/CVF Conference on Computer Vision
  and Pattern Recognition}, pp.\  76--87, 2023{\natexlab{a}}.

\bibitem[Wang et~al.(2021)Wang, Wang, Genova, Srinivasan, Zhou, Barron,
  Martin-Brualla, Snavely, and Funkhouser]{wang2021ibrnet}
Wang, Q., Wang, Z., Genova, K., Srinivasan, P.~P., Zhou, H., Barron, J.~T.,
  Martin-Brualla, R., Snavely, N., and Funkhouser, T.
\newblock Ibrnet: Learning multi-view image-based rendering.
\newblock In \emph{Proceedings of the IEEE/CVF Conference on Computer Vision
  and Pattern Recognition}, pp.\  4690--4699, 2021.

\bibitem[Wang et~al.(2023{\natexlab{b}})Wang, Skorokhodov, and
  Wonka]{wang2023pet}
Wang, Y., Skorokhodov, I., and Wonka, P.
\newblock Pet-neus: Positional encoding tri-planes for neural surfaces.
\newblock In \emph{Proceedings of the IEEE/CVF Conference on Computer Vision
  and Pattern Recognition}, pp.\  12598--12607, 2023{\natexlab{b}}.

\bibitem[Yang et~al.(2023)Yang, Pavone, and Wang]{yang2023freenerf}
Yang, J., Pavone, M., and Wang, Y.
\newblock Freenerf: Improving few-shot neural rendering with free frequency
  regularization.
\newblock In \emph{Proceedings of the IEEE/CVF Conference on Computer Vision
  and Pattern Recognition}, pp.\  8254--8263, 2023.

\bibitem[Yu et~al.(2021)Yu, Ye, Tancik, and Kanazawa]{yu2021pixelnerf}
Yu, A., Ye, V., Tancik, M., and Kanazawa, A.
\newblock pixelnerf: Neural radiance fields from one or few images.
\newblock In \emph{Proceedings of the IEEE/CVF Conference on Computer Vision
  and Pattern Recognition}, pp.\  4578--4587, 2021.

\bibitem[Yu et~al.(2023)Yu, Sohn, Kim, and Shin]{yu2023video}
Yu, S., Sohn, K., Kim, S., and Shin, J.
\newblock Video probabilistic diffusion models in projected latent space.
\newblock In \emph{Proceedings of the IEEE/CVF Conference on Computer Vision
  and Pattern Recognition}, pp.\  18456--18466, 2023.

\bibitem[Yuan et~al.(2022)Yuan, Lai, Huang, Kobbelt, and Gao]{yuan2022neural}
Yuan, Y.-J., Lai, Y.-K., Huang, Y.-H., Kobbelt, L., and Gao, L.
\newblock Neural radiance fields from sparse rgb-d images for high-quality view
  synthesis.
\newblock \emph{IEEE Transactions on Pattern Analysis and Machine
  Intelligence}, 2022.

\bibitem[Y{\"u}ce et~al.(2022)Y{\"u}ce, Ortiz-Jim{\'e}nez, Besbinar, and
  Frossard]{yuce2022structured}
Y{\"u}ce, G., Ortiz-Jim{\'e}nez, G., Besbinar, B., and Frossard, P.
\newblock A structured dictionary perspective on implicit neural
  representations.
\newblock In \emph{Proceedings of the IEEE/CVF Conference on Computer Vision
  and Pattern Recognition}, pp.\  19228--19238, 2022.

\end{thebibliography}
\bibliographystyle{icml2024}

\newpage
\appendix
\onecolumn
\setcounter{equation}{0}
\renewcommand{\theequation}{\Alph{section}.\arabic{equation}}
\setcounter{table}{0}
\renewcommand{\thetable}{\Alph{section}.\arabic{table}}
\setcounter{figure}{0}
\renewcommand{\thefigure}{\Alph{section}.\arabic{figure}}

\section{Background}
\label{app_background}

Before delving into the details of the proposed method, we briefly review the fundamentals of the neural radiance fields and multi-plane approach. 
We describe TensoRF \citep{chen2022tensorf} for the static NeRFs and HexPlane \citep{cao2023hexplane} for the dynamic NeRFs. These methods are considered representative works in multi-plane encoding and are serve as main baselines in this paper. 

\begin{figure*}[!h]{
    \centering
        \begin{subfigure}{0.49\textwidth}
            \includegraphics[width=\textwidth]{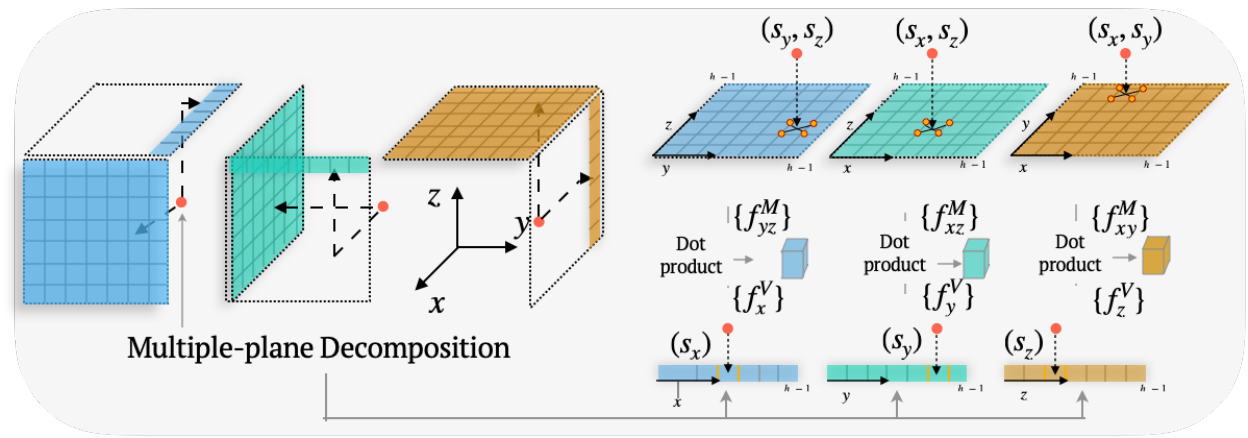}
            \caption{TensoRF}
            \label{fig_background_schematic_tensorf}
        \end{subfigure}    
        \begin{subfigure}{0.49\textwidth}
            \includegraphics[width=\textwidth]{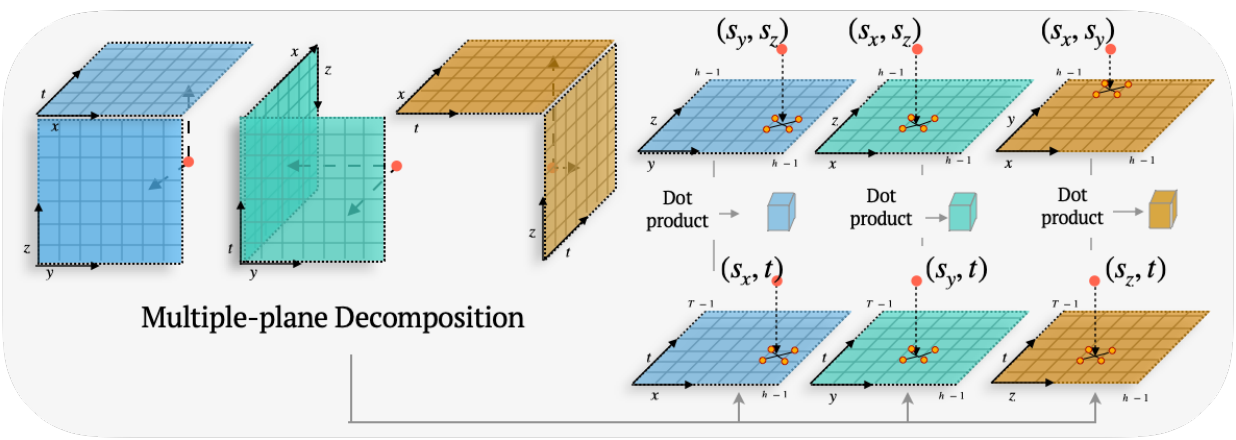}    
            \caption{HexPlane}
            \label{fig_background_schematic_hexplane}
        \end{subfigure}
    \caption{The schematic of baselines that use the multi-plane encoding. (a) TensoRF employs three planes and lines \citep{chen2022tensorf}. (b) HexPlane adopts a total of six multiple planes to include the time axis \citep{cao2023hexplane}.}
    \label{fig_background_schematic}
}    
\end{figure*}
\vspace{-0.02in}
\subsection{Neural radiance fields}
\label{app_nerf}
\citet{mildenhall2021nerf} proposed the original NeRF that uses volume rendering to compute predicted color values for novel view synthesis. In this framework, we consider a camera with origin $o$ and a ray direction $d$. A ray $\textbf{r}$, composed of $n$ points, is constructed as $o + \tau_k \cdot d$, where $\tau_k \in \{\tau_1, \cdots, \tau_n\}$. The neural radiance field, parameterized by $\Theta$, predicts the color and density values $c_{\Theta}^k$, $\sigma_{\Theta}^k$ at each point. Using volume rendering, the predicted color value $\hat{\textbf{c}}(\textbf{r})$ are computed as follows; $\hat{\textbf{c}}(\textbf{r}; \Theta) = \sum_n T_n (1 - \exp(-\sigma_{\Theta}^k(\tau_{k+1} - \tau_{k})))c_{\Theta}^k$. 
Here, the accumulated transmittance is computed by $ T_n = \exp(- \sum_{k < n} \sigma_{\Theta}^k (\tau_{k+1} - \tau_k))$. The network parameters $\Theta$ are trained by minimizing the phometric loss, comparing $\hat{\textbf{c}}(\textbf{r})$ to the ground-truth color $\textbf{c}$. 

However, raw coordinate features alone are insufficient for describing high-frequency details. To resolve this, the paper proposes sinusoidal encoding, which transform coordinates into wide-spectrum frequency components. This encoding enables the description of both low and high-frequency signals, on the other hands,  training can be time-consuming since it relies on implicit learning.

\subsection{TensoRF: Tensorial Radiance Fields}
\label{app_tensorf}
The tensorial radiance fields provide an explicit parameterization using multiple-plane and fewer MLP layers. 
Compared to other explicit parameterization \citep{liu2020neural,sun2022direct,muller2022instant}, multi-plane parameterization efficiently proves to be efficient for 3-dimensional NeRFs, provided that the plane resolution is sufficiently high. For simplicity, we assume that multi-planes share the same dimension in height, width, and depth denoted as $H$. 
This approach employs both plane features denoted as $\mathcal{M} = \{M_{xy}, M_{yz}, M_{zx} \}$ and vector features $\mathcal{V} = \{V_{z}, V_{x}, V_{y} \}$. For convenience, we denote two index variables, $i \in \{ xy, yz, zx\}$ for $\mathcal{M}$ and $j \in \{z, x, y\}$ for $\mathcal{V}$. 
The plane and vector feature is denoted as $M_i \in \mathbb{R}^{c \times H \times H}$, $V_i \in \mathbb{R}^{c \times 1 \times H}$. Both plane and vector features have a channel dimensions $c$ to represent diverse information.
To calculate the feature value at a given point $s:=(s_x, s_y, s_z)$, the point are projected to corresponding planes and lines, and features on the nearest vertices are bilinear interpolated, as illustrated in \autoref{fig_background_schematic_tensorf}.  
After obtaining the feature values from $\mathcal{M}$ and $\mathcal{V}$, denoted as $f^{\mathcal{M}} = \{f^{M}_{xy}, f^{M}_{yz}, f^{M}_{zx}\}$, and $f^{\mathcal{V}} = \{f^{V}_{z}, f^{V}_{x}, f^{V}_{y} \}$ and each feature $f^{\cdot}_i \in \mathbb{R}^c$ , hence $f^{\mathcal{M}}, f^{\mathcal{V}} \in \mathbb{R}^{3c}$. 
We use element-wise multiplication on $f^{\mathcal{M}}$, $f^{\mathcal{V}}$ to get final feature $f = f^{\mathcal{M}} \odot f^{\mathcal{V}} \in \mathbb{R}^{3c}$. For a more detailed explanation of multi-plane encoding, please refer to \autoref{app_sec_multiple-plane encoding}.
TensoRF has independent multi-plane features for density and appearance.
TensoRF predicts occupancy by channel-wise summation of final density features across all planes. Conversely, appearance features are concatenated and then fed into MLP layers or spherical harmonics function.

Multiple-plane encoding is mainly designed to emphasize local representation with the nearest vertices. Therefore, TensoRF proposes gradually increasing the resolutions of the learnable planes and vectors during training to address this locality. This intends the model to learn the global context at the coarser resolution and then enhance finer details at the high resolution.

\subsection{HexPlane}
\label{app_hexplane}
The following work, HexPlane, extends the multi-plane approach by incorporating the time axis, enabling it to work effectively in dynamic NeRFs.  
To achieve this, HexPlane builds upon the line features used in TensoRF, extending them into plane features by adding a time axis. 
This results in six planes, three spatial planes denoted as $\mathcal{M} = \{M_{xy}, M_{yz}, M_{zx} \}$, $M_i \in \mathbb{R}^{c \times H \times H}$ and three temporal planes $\mathcal{V} = \{V_{tz}, V_{tx}, V_{ty} \}$,  $V_i \in \mathbb{R}^{c \times T \times H}$ as shown in \autoref{fig_background_schematic_hexplane}. Likewise the previous subsection, we denote two index variables, $i \in \{ xy, yz, zx\}$ for $\mathcal{M}$ and $j \in \{tz, tx, ty\}$ for $\mathcal{V}$. 
Compared to TensoRF, a key difference is that the sample $s := (s_x, s_y, s_z, t)$ includes the time variable. 
In dynamic NeRFs, dealing with temporal sparsity is a crucial factor for improving performance since the time axis contains relatively sparse information compared to spatial information.
HexPlane addresses this challenge by employing denoising regularization, laplacian smoothing, that constrains similarity among adjacent multi-plane features. 
For an arbitrary plane feature $P$, Laplacian smoothing function $\mathcal{L}_l$ is defined as below, where $h,w$ refer row and column indices:
\begin{equation}
\mathcal{L}_{l}(P)=\sum_c \sum_{hw} \left(\left\|{P}^{c}_{h+1, w}-P^{c}_{h, w}\right\|_2^2+\left\|P^c_{h, w+1}-P^{c}_{h, w}\right\|_2^2\right).
\end{equation}
Specifically, HexPlane applies laplacian smoothing on both plane features but give higher priority to temporal planes.
This emphasize that time information is significant for capturing dynamic motion accurately.
Fundamental operations of HexPlane align with TensoRF, including the direct prediction of density values by multi-plane features and the prediction of color values by concatenating multi-plane features, which are then fed into MLP layers.

\section{Multiple-plane Encoding and Concatenating Coordinate}
\label{app_sec_multiple-plane encoding}

\begin{wrapfigure}{R}{0.60\columnwidth}{
    \centering
    \includegraphics[width=0.6\columnwidth]{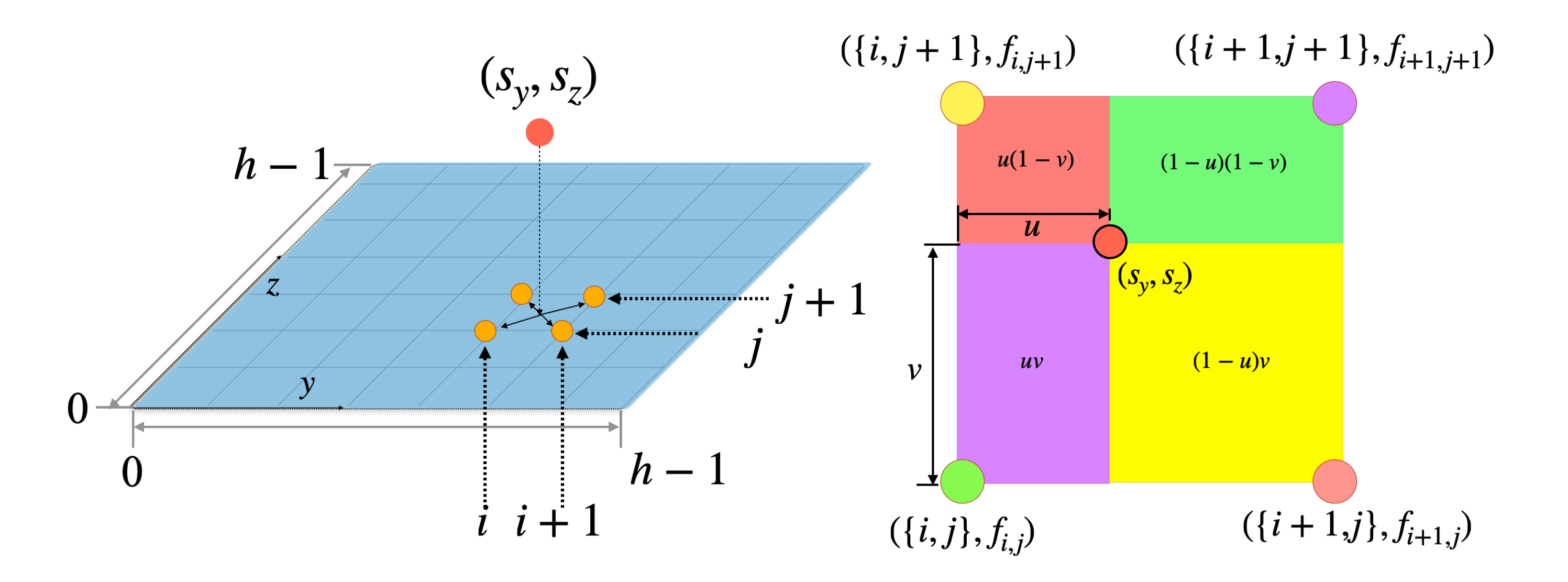}
    \caption{Blinear interpolation}
    \label{app_fig_bilinear_interpolation}
}    
\end{wrapfigure}

In this subsection, we discuss the use of multiple-plane encoding. Instead of directly predicting the density function using low-rank approximation of voxel grid, as done in previous methods \citep{chen2022tensorf,cao2023hexplane}, our focus is on creating spatial features with multiple planes.
For 3-dimensional data, we denote the plane features as $M_i \in \mathbb{R}^{c \times H \times H}$, and vector features $V_i \in \mathbb{R}^{c \times 1 \times H}$. However, in the case of 4-dimensional data, $V$ changes to plane features.
Each plane and vector feature corresponds to an axis in 3-dimensional spaces, such as $\mathcal{M} = \{M_{xy}, M_{yz}, M_{xz}\}$ and $\mathcal{V} = \{V_z, V_x, V_y\}$.
In 4-dimensional spaces, the same notation applies to $\mathcal{M}$, but we introduce a time axis in $\mathcal{V}$ represented as $\mathcal{V} = \{V_{zt}, V_{xt}, V_{yt}\}$. 
The dimensions of $M_{(\cdot)}$ and $V_{(\cdot)}$ are $H \times W$ and $D$, respectively. We assume that all planes and vectors have the same dimension, i.e., $H = W = D$. We use $h$ as the all grid dimension for plane and vector features for simplicity.

To compute multiple-plane features, we use bilinear interpolation. In 3-dimensional data, when 
a data point $s \in \mathbb{R}^3$ is queried, it first drops to the axis for the corresponding dimension, then looks for the nearest vertices.
For example, when obtaining plane features on $M_{x, y}$, $s = (s_x, s_y, s_z)$ drops $s_z$ and then looks for corresponding adjacent vertices in $M_1$. 
When $(i,j) = \lfloor (s_x, s_y) \rfloor$, the adjacent vertices are defined as $\{ (i, j), (i+1, j), (i, j+1), (i+1, j+1) \}$, and their feature values are denoted as $\{f_{(i,j)}, f_{(i+1,j)}, f_{(i,j+1)}, f_{(i+1,j+1)}\}$ at the four nearest grid points. Here, $i, j \in \{0, 1, \cdots, h-1\}$.
The component of multiple-plane encoding ${f(s_x, s_y)}$ is computed by bilinear interpolation as follows:

\begin{equation}
    \label{eq_planes}
    \begin{gathered}
        f_{(s_x, s_y)} = (1 - u)(1 - v)f_{i,j} + u(1 - v)f_{i+1, j} + (1 - u)vf_{i, j+1} + uvf_{i+1, j+1}
    \end{gathered}
\end{equation}

where, $u = \nicefrac{(s_x - i)}{(i+1 - i)}$ is the interpolation factor in the $x$-direction, and $v = \nicefrac{(s_y - j)}{(j+1 - j)}$ is the interpolation factor in the $y$-direction.
The remaining components $(f_{(s_y, s_z)}, f_{(s_z, s_z)})$ are also computed by simply alternating coordinates. For the vector feature, we use linear interpolation, similar to bilinear interpolation but in 1 dimension.
In 3-dimentional data, the features collected are 
$f^M =\{f_{(s_x, s_y)}, f_{(s_y, s_z)}, f_{(s_z, s_x)}\}$ and $f^V = \{f_{s_z}, f_{s_x}, f_{s_y}\}$, 
In 4-dimensional data, we can also use bilinear interpolation for $\mathcal{V}$. 
In this case, the features are $f^M =\{f_{(s_x, s_y)}, f_{(s_y, s_z)}, f_{(s_z, s_x)}\}$ and $f^V = \{f_{(s_z, t)}, f_{(s_x, t)}, f_{(s_y, t)}\}$. 
Then, we combine them by element-wise producting the two vectors $f = f^M \odot  f^V$ to get multiple-plane encoding in $\mathbb{R}^{3c}$.

To reprensete low-frequencies signals apparently, we include the coordinate of a data point $s = \{s_x, s_y, s_z\} \in \mathbb{R}^3$ in 3-dimensional data. In 4-dimensinoal data, these coordinate features become $s = \{s_x, s_y, s_z, t\} \in \mathbb{R}^4$. 
The final result of encoding is the concatenation of two different features: $\textbf{f} = \{f, s\}$. For 3-dimensional data, $\textbf{f}$ is  in $\mathbb{R}^{3+3c}$, and in case of 4-dimensional data, $\textbf{f}$ is in  $\mathbb{R}^{4+3c}$.
\section{Implementation details}
\label{app_implementation_details}

\subsection{Hyper-parameters on the Static NeRF}
The proposed model incorporates multi-plane encoding and MLPs with skip connections. For the multi-plane encoding, we utilize 48-dimensional channels. The resolution of plane features is upscaled to 8,000,000 ($200^3$) by the end of training. 
The hyper-parameters, such as the weight for Laplacian smoothing (\(\lambda_1\)), the curriculum learning schedule, and the initial feature resolution, vary across scenes as part of our hyperparameter tuning to achieve optimal results. However, we use same $\lambda_{2,3}$ across scenes. In this dataset, $\lambda_2=1$ is used, and $\lambda_3$ is initially set to $0.00008$ at the start of training, increasing to $0.00004$. This is the same weighting strategy used in the previous method \cite{chen2022tensorf}. 
Detailed information on the hyperparameters for multi-plane encoding can be found in \autoref{app_implementation_details_static}.
For the decoder, we employ standard fully connected layers with ReLU activations, each containing 256 channels. 
The encoder consists of four fully connected ReLU layers, with a skip connection introduced after the second layer, which concatenates the fused input features. 
Occupancy is directly calculated from the obtained features with \texttt{softplus} function applied to the first channel. 
The RGB decoder, following this, consists of two layers. 
The color values are obtained from the features processed by the RGB decoder through \texttt{sigmoid} activation.

In our experiments, the model was trained over 30,000 iterations with a batch size of 4,096. We utilized the Adam optimizer \citep{KingBa15} with an initial learning rate of 0.02 for multi-plane features and 0.001 for MLPs, following a learning rate schedule inspired by TensorRF \citep{chen2022tensorf}.
\begin{table}[!h]
    \caption{The detailed configuration for the static NeRF experiments. The parameters of curriculum $\{t_e, t_s\}$ are defined in \autoref{eq_curriculum_weighting}. These values are presented as a percentage of the total iteration. The hyphen means that curriculum learning does not apply.} 
    \label{app_implementation_details_static}
    \centering
    \resizebox{0.85\columnwidth}{!}{%
        \begin{tabular}{ccccccccc}
        \toprule
        \multirow{2}{*}{Configs} & \multicolumn{8}{c}{scenes}                                                      \\
        \cmidrule(r){2-9}
                                        & chair                            & drums                            & ficus                            & hotdog                           & lego                             & materials                        & mic                              & ship                                    \\
        \midrule            
        $\lambda_1$                & 0.001 & 0.005 & 0.005 & 0.009 & 0.009 & 0.001 & 0.009 & 0.005 \\
        curriculum learning        & - & \{5, 95\} & - & - & \{10, 50\} & - & \{0, 50\} & - \\
        Initial resolution     & 16 & 3 & 3 & 24 & 48 & 48 & 48 & 3 \\
        \bottomrule
    \end{tabular}
    }
\end{table}
\subsection{Hyper-parameters on the Dynamic NeRF}
The configuration for the dynamic Neural Radiance Fields (NeRF) case adheres to the same settings as the static case. We utilize plane features with 48 channels. The initial voxel resolution is set at 4,096 ($16^3$) and is subsequently upscaled to 8,000,000 ($200^3$). 
For hyper-parameters, $\lambda_2$ is set to 2.5 for all scenes in the dynamic NeRFs to ensure smoother multi-plane features along the time-axis. This approach is proven effective in previous work \cite{cao2023hexplane}. The parameter $\lambda_3$ is set to 0.00001 across all dynamic NeRF scenes. 
For a more detailed description, please refer to \autoref{app_implementation_details_dynamic}.
The structure of the decoder, initial learning rate, and optimizer configuration remain identical to those used in the static NeRF. Any configurations not specified here follow directly from the HexPlane method as described in \citep{cao2023hexplane}.
\begin{table}[!h]
    \caption{The detailed configuration for the static NeRF experiments. The parameters of curriculum $\{t_e, t_s\}$ are defined in \autoref{eq_curriculum_weighting}. These values are presented as a percentage of the total iteration. The hyphen means that curriculum learning does not apply.} 
    \label{app_implementation_details_dynamic}
    \centering
    \resizebox{0.85\columnwidth}{!}{%
        \begin{tabular}{ccccccccc}
        \toprule
        \multirow{2}{*}{Configs} & \multicolumn{8}{c}{scenes}                                                      \\
        \cmidrule(r){2-9}
                                        & \small{boundingballs}                            & \small{hellwarrior}                            & \small{hook}                            & \small{jumpingjacks}                           & \small{lego}                             & \small{mutant}                        & \small{standup}                              & \small{trex}                                    \\
        \midrule            
        $\lambda_1$                & 0.001 & 0.005 & 0.001 & 0.001 & 0.05 & 0.001 & 0.05 & 0.05 \\
        curriculum learning        & - & \{5, 95\} & - & - & \{5, 95\} & - & \{5, 95\} & \{5, 95\} \\
        \bottomrule
    \end{tabular}
    }
\end{table}
\section{Experimental Setup}
\label{app_experimental_Setting}

We conducted the training and evaluation of all models using an NVIDIA A6000 with 48 GB of memory. It's important to note that each experiment was executed once using the seed $0$ as the default. When an experiment explicitly demanded five trials, we utilized five different seeds: \{0, 700, 19870929, 20220401, 20240507\}. For further details regarding the datasets and the baselines, we provide additional explanations in the following subsection.

\subsection{Datasets}
\paragraph{NeRF blender dataset}  
The Blender Dataset \cite{mildenhall2021nerf} is a set of synthetic, bounded, 360\textdegree, in-ward facing multi-view images of static object.
Blender Dataset includes eight different scenes.
Following the previous method\cite{yang2023freenerf,jain2021putting}, for training, we used 8 views with IDs of {26, 86, 2, 55, 75, 93, 16, 73 and 8}  counting from zeros.
While previous works uniformly sampled 25 images from the original test set \cite{yang2023freenerf,jain2021putting}, we evaluate all data using full-resolution images (800 × 800 pixels) for both training and testing.
We downloaded Blender dataset from \url{https://www.matthewtancik.com/nerf}

\paragraph{D-NeRF dataset}
The D-NeRF dataset is a set of synthetic, bounded, 360 degree, monocular videos for dynamic objects \cite{pumarola2021d}.
The D-NeRF dataset includes eight different scenes of varying duration, from 50 frames to 200 frames.
To train the baseline under severe sparsity settings, we sub-sample the number of training views from the original D-NeRF dataset. For instance, in the case of \texttt{bouncingballs} that originally contains 150 views in the training set, we select a total of 25 views, evenly spaced apart, by starting from 0 and increasing by 6 at each step. For other scenes and varying number of views, we apply the same sampling method.
We downloaded D-NeRF dataset from \url{https://github.com/albertpumarola/D-NeRF}

\paragraph{Tank and Temples}
The Tank and Temples dataset includes real-world scenes and corresponding multi-view images of static objects \cite{Knapitsch2017}.
In this study, we select four scenes, \texttt{Family}, \texttt{Barn}, \texttt{Truck} and \texttt{Caterpillar}. Each scene exhibits variations in the number of training poses and images, reflecting different camera distribution. For instance, some poses are placed close to the object, while others are farther away, creating varying levels of difficult. Among scenes, the \texttt{Family} is relatively similar to the original in-ward case.
The dataset is obtained from the following URL: \url{https://dl.fbaipublicfiles.com/nsvf/dataset/TanksAndTemple.zip}

\subsection{Baselines}

In this chapter, we briefly explain the method we compared as a baseline in our experiments.
Regarding TensorRF and Hexplane, we described in detail in \autoref{app_background}.

\paragraph{Diet-NeRF}
Diet-NeRF is a sinusoidal encoding based model \citep{jain2021putting}. The model incorporates auxiliary semantic consistency loss which leverages the pre-trained CLIP network trained on large datasets to compensate for the lack of training data.
Auxiliary semantic consistency loss regularize semantic similarity between rendered view and given input images.
We also compare the simplified NeRF \cite{jain2021putting}.
For implementation we used the codebase in \url{https://github.com/ajayjain/DietNeRF}

\paragraph{Free-NeRF}
Free-NeRF is a model based on sinusoidal encoding \cite{yang2023freenerf}.
\citet{yang2023freenerf} employed progressive activation of positioning embedding within a single model. 
It initially establishes global contextual shape and subsequentially describes fine-grained details.
To reduce floating artifacts, it penalize near-camera density values, following the prior knowledge of object is located in a distance to the camera.
For implementation we used the code from \url{https://github.com/Jiawei-Yang/FreeNeRF/tree/main}

\paragraph{DVGO}
DVGO is a model that uses a three-dimensional dense voxel feature grid \cite{sun2022direct}. 
It utilizes independent voxel features for density and color. 
Shallow MLP follows color encoding. 
In the first stage, coarse geometry explores learning the shape prior of the scene and finding empty voxels. 
Subsequently, in the fine reconstruction stage, they upsample the grid to a higher resolution and apply free-space skipping to optimize the occupied section densely.
We used the code from \url{https://github.com/sunset1995/DirectVoxGO}

\paragraph{Instant-NGP}
Instant NGP model expresses the voxel feature grid using the Hash function \cite{muller2022instant}. 
It allocates features corresponding to each voxel to the hash table, reducing the memory required while allowing collisions. 
Instant NGP utilizes the multi-resolution feature grid and uses features of resolution that log-scale uniformly increase from 16 to 1024-4096. 
It maintains a fast speed by inferring empty spaces through occlusion values such as TensorRF and DVGO and avoiding sampling from void regions.
We used the code from \url{https://github.com/kwea123/ngp_pl}

\paragraph{VGOS}
VGOS is the first example of applying the grid-based method to a few-shot case \cite{sun2023vgos}. The method induces smoothness by adding total variation regularization to the dense grid feature, feature, depth, and color.
In addition, progressive voxel sampling is introduced to prevent floating artifacts under the assumption that there will be a lot in the middle of the occlusion.
We follows the code from
\url{https://github.com/SJoJoK/VGOS}

\paragraph{K-Planes}
K-planes utilizes the Hadamard product of multi-resolution tri-planes to represent voxel features \cite{fridovich2023k}. This approach extends from static three-dimensional scenes to dynamic four-dimensional NeRFs like Hex-plane \cite{cao2023hexplane}. K-Planes incorporates TV Loss and employs various regularization methods including distortion loss to reduce floating point artifacts. Furthermore, it adopts the proposal network method suggested in MipNeRF 360 as a sampling approach \cite{barron2022mip}.
We follow the code from \url{https://github.com/sarafridov/K-Planes}

\paragraph{CAM}
CAM, coordinate-award modulation, is an approach to create parameter efficient neural fields \citep{lee2023coordinate}. It combines explicit representation, such as hash-grid, with a coordinate network. Unlike previous explicit representation that had a large number of parameters due to thier channel dimension. CAM utilizes only two channel explicit representation. Each channel is integrated using an affine transformation with the feature from coodinate network. Additionally, a fourier feature network is used to reduce the number of overall parameters while preserving performance under full poses. However, CAM does not specifically address spectral bias for each feature and sparse input situation. 
We follow the code from \url{https://github.com/maincold2/CAM}
\section{The Evaluation Statistics of Static and Dynamic NeRF Datasets}
\label{app_statistics_result}

\paragraph{Static NeRFs} 
In \autoref{app_table_static_psnr} through \autoref{app_table_static_lpips}, we present the quantitative results for each scene of the synthetic NeRF Dataset. All reported numbers are averages of five experiments, along with their corresponding standard deviations. Our model consistently outperformed all counterpart models across all metrics. We also analyze the performance of the TensorRF model, which incorporates intense additional Laplacian smoothness loss. The optimal $\lambda_1$ value of 0.001 was identified for achieving the best results, as detailed in \autoref{app_varying_lambda1}.
TensorRF with strong Laplacian regularization shows performance comparable to our proposed model. Both two methods exhibit complementary advantages in novel-view rendering results. For qualitative comparison, we showcase novel-view renderings of \texttt{ship} (\autoref{fig_exp_static_nerf}). While TensorRF with $\lambda_1=0.001$ focuses on reconstructing higher-frequency textures, it shows instability in low-frequency information, such as geometry (deck in \texttt{ship}), and displays high-frequency artifacts in color (water regions in \texttt{ship}). Conversely, the proposed method excels in robust optimization, particularly in capturing global information. It enables more accurate 3D geometry and consistent color reconstruction across views. However, it might underfits in scenarios that require intricate details. Despite this, without relying heavily on denoising regularization, the proposed method nearly achieves the best performance, primarily attributed to the coordinate-based networks responsible for capturing the global context.

\begin{table}[!h]
    \caption{The result of average PSNR in the static NeRF. We conduct five trials and use 8 views for training.} 
    \label{app_table_static_psnr}
    \centering
    \resizebox{0.92\columnwidth}{!}{%
        \begin{tabular}{lcccccccc}
        \toprule
        \multirow{2}{*}{Models} & \multicolumn{8}{c}{PSNR $\uparrow$}                                                      \\
        \cmidrule(r){2-9}
                                        & chair                            & drums                            & ficus                            & hotdog                           & lego                             & materials                        & mic                              & ship                                    \\
        \midrule            
        Simplified\_NeRF                & 20.354 \footnotesize{$\pm$0.648} & 14.188 \footnotesize{$\pm$2.596} & 21.629 \footnotesize{$\pm$0.171} & 22.565 \footnotesize{$\pm$1.055} & 12.453 \footnotesize{$\pm$3.103} & 18.976 \footnotesize{$\pm$2.306} & 24.950 \footnotesize{$\pm$0.202}  & 18.648 \footnotesize{$\pm$0.446} \\
        DietNeRF                        & 21.323 \footnotesize{$\pm$2.478} & 14.156 \footnotesize{$\pm$5.143} & 13.082 \footnotesize{$\pm$3.892} & 11.644 \footnotesize{$\pm$6.753} & 16.120 \footnotesize{$\pm$7.121}  & 12.200 \footnotesize{$\pm$7.343}   & 24.701 \footnotesize{$\pm$1.222} & 19.342 \footnotesize{$\pm$4.033} \\
        HALO                            & 24.765 \footnotesize{$\pm$0.280} & 18.674 \footnotesize{$\pm$0.226} & 21.424 \footnotesize{$\pm$0.204} & 10.220 \footnotesize{$\pm$0.388} & 22.407 \footnotesize{$\pm$1.997} & 20.996 \footnotesize{$\pm$0.032} & 24.937 \footnotesize{$\pm$0.078} & 21.665 \footnotesize{$\pm$0.229} \\
        FreeNeRF                        & 26.079 \footnotesize{$\pm$0.545} & 19.992 \footnotesize{$\pm$0.050} & 18.427 \footnotesize{$\pm$2.819} & 28.911 \footnotesize{$\pm$0.232} & 24.121 \footnotesize{$\pm$0.633} & 21.738 \footnotesize{$\pm$0.085} & 24.890 \footnotesize{$\pm$1.733}  & 23.011 \footnotesize{$\pm$0.148} \\
        \midrule            
        DVGO                            & 22.347 \footnotesize{$\pm$0.253} & 16.538 \footnotesize{$\pm$0.081} & 19.032 \footnotesize{$\pm$0.071} & 24.725 \footnotesize{$\pm$0.241} & 20.845 \footnotesize{$\pm$0.129} & 18.497 \footnotesize{$\pm$0.077} & 24.373 \footnotesize{$\pm$0.252} & 18.170 \footnotesize{$\pm$0.148}  \\
        VGOS                            & 22.100 \footnotesize{$\pm$0.036} & 18.568 \footnotesize{$\pm$0.112} & 19.084 \footnotesize{$\pm$0.061} & 24.736 \footnotesize{$\pm$0.073} & 20.895 \footnotesize{$\pm$0.073} & 18.418 \footnotesize{$\pm$0.036} & 24.180 \footnotesize{$\pm$0.148}  & 18.155 \footnotesize{$\pm$0.060}  \\
        iNGP                            & 24.762 \footnotesize{$\pm$0.169} & 14.561 \footnotesize{$\pm$0.082} & 20.678 \footnotesize{$\pm$0.415} & 24.105 \footnotesize{$\pm$0.308} & 22.222 \footnotesize{$\pm$0.076} & 15.159 \footnotesize{$\pm$0.075} & 26.186 \footnotesize{$\pm$0.159} & 17.288 \footnotesize{$\pm$0.135} \\
        TensoRF                         & 26.234 \footnotesize{$\pm$0.062} & 15.940 \footnotesize{$\pm$0.369} & 21.373 \footnotesize{$\pm$0.152} & 28.465 \footnotesize{$\pm$0.387} & 26.279 \footnotesize{$\pm$0.279} & 20.221 \footnotesize{$\pm$0.109} & 26.392 \footnotesize{$\pm$0.320}  & 20.294 \footnotesize{$\pm$0.359} \\
        TensorRF($\lambda_1=0.001$)     & 28.527 \footnotesize{$\pm$0.208} & 19.626 \footnotesize{$\pm$0.134} & 21.963 \footnotesize{$\pm$0.217} & 29.373 \footnotesize{$\pm$0.218} & 29.441 \footnotesize{$\pm$0.270} & 21.911 \footnotesize{$\pm$0.087} & 26.998 \footnotesize{$\pm$0.325} & 22.837 \footnotesize{$\pm$0.717} \\
        K-Planes                        & 27.300 \footnotesize{$\pm$0.192} & 20.427 \footnotesize{$\pm$0.153} & 23.820 \footnotesize{$\pm$0.215} & 27.576 \footnotesize{$\pm$0.254} & 26.520 \footnotesize{$\pm$0.262} & 19.661 \footnotesize{$\pm$0.178} & 27.297 \footnotesize{$\pm$0.144} & 21.337 \footnotesize{$\pm$0.240} \\
        \midrule
        Ours                            & 28.021 \footnotesize{$\pm$0.143} & 19.550 \footnotesize{$\pm$0.587}  & 20.301 \footnotesize{$\pm$0.258} & 29.247 \footnotesize{$\pm$0.656} & 26.725 \footnotesize{$\pm$0.565} & 21.927 \footnotesize{$\pm$0.114} & 26.416 \footnotesize{$\pm$0.199} & 24.266 \footnotesize{$\pm$0.163} \\

        \bottomrule
    \end{tabular}
    }%
\end{table}

\begin{table}[!h]
    \caption{The result of average SSIM in the static NeRF. We conduct five trials and use 8 views for training.} 
    \label{app_table_static_ssim}
    \centering
    \resizebox{0.92\columnwidth}{!}{%
        \begin{tabular}{lcccccccc}
        \toprule
        \multirow{2}{*}{Models} & \multicolumn{8}{c}{SSIM $\uparrow$}                                                      \\
        \cmidrule(r){2-9}
                                        & chair                            & drums                          & ficus                           & hotdog                          & lego                             & materials                        & mic                              & ship                                    \\
        \midrule            
        Simplified\_NeRF                & 0.852 \footnotesize{$\pm$0.003} & 0.773 \footnotesize{$\pm$0.017} & 0.871 \footnotesize{$\pm$0.002} & 0.891 \footnotesize{$\pm$0.004} & 0.738 \footnotesize{$\pm$0.031} & 0.827 \footnotesize{$\pm$0.019} & 0.931 \footnotesize{$\pm$0.001} & 0.736 \footnotesize{$\pm$0.005} \\
        DietNeRF                        & 0.857 \footnotesize{$\pm$0.025} & 0.716 \footnotesize{$\pm$0.133} & 0.653 \footnotesize{$\pm$0.123} & 0.705 \footnotesize{$\pm$0.111} & 0.709 \footnotesize{$\pm$0.148} & 0.662 \footnotesize{$\pm$0.166} & 0.933 \footnotesize{$\pm$0.011} & 0.731 \footnotesize{$\pm$0.043} \\
        HALO                            & 0.883 \footnotesize{$\pm$0.001} & 0.822 \footnotesize{$\pm$0.003} & 0.877 \footnotesize{$\pm$0.002} & 0.806 \footnotesize{$\pm$0.064} & 0.827 \footnotesize{$\pm$0.032} & 0.847 \footnotesize{$\pm$0.003} & 0.931 \footnotesize{$\pm$0.000}   & 0.763 \footnotesize{$\pm$0.001} \\
        FreeNeRF                        & 0.908 \footnotesize{$\pm$0.003} & 0.852 \footnotesize{$\pm$0.001} & 0.866 \footnotesize{$\pm$0.008} & 0.942 \footnotesize{$\pm$0.002} & 0.871 \footnotesize{$\pm$0.003} & 0.862 \footnotesize{$\pm$0.001} & 0.935 \footnotesize{$\pm$0.010}  & 0.778 \footnotesize{$\pm$0.003} \\
        \midrule            
        DVGO                            & 0.860 \footnotesize{$\pm$0.003} & 0.761 \footnotesize{$\pm$0.002} & 0.857 \footnotesize{$\pm$0.001} & 0.904 \footnotesize{$\pm$0.002} & 0.820 \footnotesize{$\pm$0.001} & 0.804 \footnotesize{$\pm$0.002} & 0.933 \footnotesize{$\pm$0.001} & 0.689 \footnotesize{$\pm$0.003} \\
        VGOS                            & 0.857 \footnotesize{$\pm$0.001} & 0.834 \footnotesize{$\pm$0.001} & 0.859 \footnotesize{$\pm$0.000} & 0.905 \footnotesize{$\pm$0.000} & 0.824 \footnotesize{$\pm$0.000} & 0.804 \footnotesize{$\pm$0.001} & 0.932 \footnotesize{$\pm$0.001} & 0.686 \footnotesize{$\pm$0.001} \\
        iNGP                            & 0.899 \footnotesize{$\pm$0.002} & 0.730 \footnotesize{$\pm$0.002} & 0.886 \footnotesize{$\pm$0.004} & 0.904 \footnotesize{$\pm$0.001} & 0.841 \footnotesize{$\pm$0.001} & 0.748 \footnotesize{$\pm$0.002} & 0.946 \footnotesize{$\pm$0.001} & 0.672 \footnotesize{$\pm$0.002} \\
        TensoRF                         & 0.919 \footnotesize{$\pm$0.001} & 0.753 \footnotesize{$\pm$0.007} & 0.882 \footnotesize{$\pm$0.002} & 0.938 \footnotesize{$\pm$0.002} & 0.909 \footnotesize{$\pm$0.003} & 0.843 \footnotesize{$\pm$0.003} & 0.947 \footnotesize{$\pm$0.002} & 0.719 \footnotesize{$\pm$0.006} \\
        TensorRF($\lambda_1=0.001$)     & 0.943 \footnotesize{$\pm$0.001} & 0.856 \footnotesize{$\pm$0.004} & 0.901 \footnotesize{$\pm$0.001} & 0.945 \footnotesize{$\pm$0.001} & 0.941 \footnotesize{$\pm$0.002} & 0.873 \footnotesize{$\pm$0.001} & 0.955 \footnotesize{$\pm$0.002} & 0.772 \footnotesize{$\pm$0.006} \\
        K-Planes                        & 0.935 \footnotesize{$\pm$0.001} & 0.869 \footnotesize{$\pm$0.002} & 0.925 \footnotesize{$\pm$0.001} & 0.949 \footnotesize{$\pm$0.001} & 0.921 \footnotesize{$\pm$0.002} & 0.850 \footnotesize{$\pm$0.001} & 0.958 \footnotesize{$\pm$0.001} & 0.767 \footnotesize{$\pm$0.003} \\
        \midrule
        Ours                            & 0.931 \footnotesize{$\pm$0.001} & 0.860 \footnotesize{$\pm$0.011}  & 0.881 \footnotesize{$\pm$0.002} & 0.948 \footnotesize{$\pm$0.003} & 0.914 \footnotesize{$\pm$0.005} & 0.879 \footnotesize{$\pm$0.001} & 0.949 \footnotesize{$\pm$0.001} & 0.802 \footnotesize{$\pm$0.002} \\
        \bottomrule
    \end{tabular}
    }%
\end{table}
\begin{table}[!h]
    \caption{The result of average LPIPS in the static NeRF. We conduct five trials and use 8 views for training.} 
    \label{app_table_static_lpips}
    \centering
    \resizebox{0.92\columnwidth}{!}{%
        \begin{tabular}{lcccccccc}
        \toprule
        \multirow{2}{*}{Models} & \multicolumn{8}{c}{LPIPS $\downarrow$}                                                      \\
        \cmidrule(r){2-9}
                                        & chair                            & drums                            & ficus                            & hotdog                           & lego                             & materials                        & mic                              & ship                                    \\
        \midrule            
        Simplified\_NeRF                & 0.247 \footnotesize{$\pm$0.010}  & 0.388 \footnotesize{$\pm$0.083} & 0.153 \footnotesize{$\pm$0.007} & 0.239 \footnotesize{$\pm$0.009} & 0.408 \footnotesize{$\pm$0.091} & 0.205 \footnotesize{$\pm$0.042} & 0.100 \footnotesize{$\pm$0.001}   & 0.375 \footnotesize{$\pm$0.005} \\
        DietNeRF                        & 0.177 \footnotesize{$\pm$0.051} & 0.382 \footnotesize{$\pm$0.253} & 0.447 \footnotesize{$\pm$0.201} & 0.539 \footnotesize{$\pm$0.225} & 0.339 \footnotesize{$\pm$0.254} & 0.426 \footnotesize{$\pm$0.282} & 0.079 \footnotesize{$\pm$0.021} & 0.278 \footnotesize{$\pm$0.069} \\
        HALO                            & 0.134 \footnotesize{$\pm$0.003} & 0.234 \footnotesize{$\pm$0.012} & 0.109 \footnotesize{$\pm$0.012} & 0.417 \footnotesize{$\pm$0.113} & 0.149 \footnotesize{$\pm$0.066} & 0.167 \footnotesize{$\pm$0.012} & 0.098 \footnotesize{$\pm$0.004} & 0.290 \footnotesize{$\pm$0.007}  \\
        FreeNeRF                        & 0.101 \footnotesize{$\pm$0.005} & 0.142 \footnotesize{$\pm$0.003} & 0.138 \footnotesize{$\pm$0.068} & 0.069 \footnotesize{$\pm$0.001} & 0.092 \footnotesize{$\pm$0.003} & 0.107 \footnotesize{$\pm$0.002} & 0.094 \footnotesize{$\pm$0.029} & 0.228 \footnotesize{$\pm$0.003} \\
        \midrule            
        DVGO                            & 0.120 \footnotesize{$\pm$0.004} & 0.218 \footnotesize{$\pm$0.003} & 0.102 \footnotesize{$\pm$0.001} & 0.106 \footnotesize{$\pm$0.003} & 0.125 \footnotesize{$\pm$0.001} & 0.149 \footnotesize{$\pm$0.001} & 0.062 \footnotesize{$\pm$0.001} & 0.276 \footnotesize{$\pm$0.004} \\
        VGOS                            & 0.124 \footnotesize{$\pm$0.001} & 0.201 \footnotesize{$\pm$0.002} & 0.100 \footnotesize{$\pm$0.001}   & 0.104 \footnotesize{$\pm$0.001} & 0.123 \footnotesize{$\pm$0.000}   & 0.148 \footnotesize{$\pm$0.001} & 0.063 \footnotesize{$\pm$0.001} & 0.278 \footnotesize{$\pm$0.001} \\
        iNGP                            & 0.098 \footnotesize{$\pm$0.004} & 0.345 \footnotesize{$\pm$0.005} & 0.099 \footnotesize{$\pm$0.006} & 0.144 \footnotesize{$\pm$0.003} & 0.127 \footnotesize{$\pm$0.002} & 0.292 \footnotesize{$\pm$0.003} & 0.058 \footnotesize{$\pm$0.002} & 0.312 \footnotesize{$\pm$0.003} \\
        TensoRF                         & 0.074 \footnotesize{$\pm$0.002} & 0.312 \footnotesize{$\pm$0.011} & 0.105 \footnotesize{$\pm$0.003} & 0.072 \footnotesize{$\pm$0.005} & 0.059 \footnotesize{$\pm$0.002} & 0.129 \footnotesize{$\pm$0.004} & 0.047 \footnotesize{$\pm$0.002} & 0.237 \footnotesize{$\pm$0.010}  \\
        TensorRF($\lambda_1=0.001$)     & 0.047 \footnotesize{$\pm$0.001} & 0.132 \footnotesize{$\pm$0.009} & 0.066 \footnotesize{$\pm$0.001} & 0.050 \footnotesize{$\pm$0.001}  & 0.037 \footnotesize{$\pm$0.002} & 0.069 \footnotesize{$\pm$0.001} & 0.037 \footnotesize{$\pm$0.001} & 0.186 \footnotesize{$\pm$0.007} \\
        K-Planes                        & 0.052 \footnotesize{$\pm$0.002} & 0.107 \footnotesize{$\pm$0.005} & 0.061 \footnotesize{$\pm$0.002} & 0.054 \footnotesize{$\pm$0.001}  & 0.051 \footnotesize{$\pm$0.002} & 0.116 \footnotesize{$\pm$0.003} & 0.036 \footnotesize{$\pm$0.001} & 0.199 \footnotesize{$\pm$0.005} \\
        \midrule
        Ours                            & 0.078 \footnotesize{$\pm$0.001} & 0.139 \footnotesize{$\pm$0.022} & 0.082 \footnotesize{$\pm$0.003} & 0.064 \footnotesize{$\pm$0.005} & 0.057 \footnotesize{$\pm$0.005} & 0.067 \footnotesize{$\pm$0.002} & 0.059 \footnotesize{$\pm$0.001} & 0.191 \footnotesize{$\pm$0.004} \\
        \bottomrule
    \end{tabular}
    }%
\end{table}

\begin{figure}[!ht]{
    \centering
    \includegraphics[width=0.92\columnwidth]{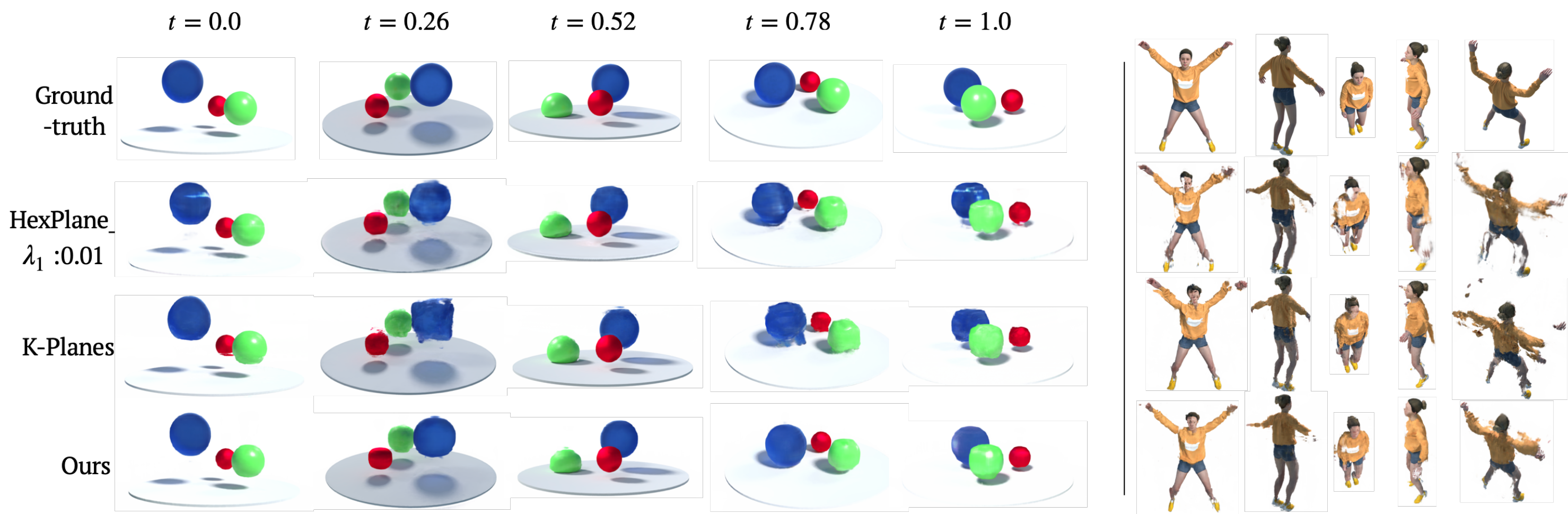}    
    \caption{Rendered images of the \texttt{bouncingballs} and \texttt{jumpingjacks} in the dynamic NeRF dataset by HexPlane with $\lambda_1=0.01$, K-Planes and ours. All models are trained using 25 views}
    \label{app_fig_exp_dynamic_nerf}
}    
\end{figure}

\paragraph{Dynamic NeRFs} 
In the evaluation of Dynamic Neural Radiance Fields (D-NeRF), the experimental results demonstrate a significant performance improvement for the proposed method over baseline approaches. While baselines perform comparably when full poses are available, the proposed method particularly excels as the number of available poses diminishes. This is evident when testing all methods with 25 poses, where a notable performance gap is observed. This gap narrows with a decrease in pose availability, highlighting the challenges of capturing object movement and synthesizing novel views in dynamic scenes with limited data, especially with only \{15, 20\} frames.
Specifically, in scenes with significant movement, such as \texttt{bouncingballs} and \texttt{standup}, the proposed method significantly outperforms others. For example, as depicted in \autoref{app_fig_exp_dynamic_nerf}, while variants of HexPlane and K-Planes struggle to accurately render the shape of the blue ball over time, the proposed method successfully captures this detail, including the reflection on the green ball. In the \texttt{jumpingjack} sequence, the proposed method also shows fewer artifacts and maintains scene boundaries more effectively compared to HexPlane.
Overall, as indicated in \autoref{app_table_dynamic_nerf}, the dynamic NeRF dataset demands a model capable of handling time in a continuous manner. Traditional grid-type explicit representations fall short as they rely on discretizing each feature, including time. In contrast, the proposed method leverages a coordinate network that consists of continuous maps, enhanced by multi-plane representations, enabling superior performance on the D-NeRF dataset compared to other baselines.
\begin{table*}[!t]
    \caption{Result of evaluation statistics on the D-NeRF datasets. HexPlane employs the weight of denoising regularization as $\lambda_1=0.01$ via grid-search. Average PSNR, SSIM, and LPIPS are calculated across all scenes. We indicates best performance as \textbf{bold} for each cases}
    \label{app_table_dynamic_nerf}
    \centering
    \resizebox{0.92\columnwidth}{!}{%
        \begin{tabular}{clccccccccccc}
        \toprule
        \multirow{2}{*}{\begin{tabular}[c]{@{}l@{}}Training \\ views\end{tabular}} & \multirow{2}{*}{Models} & \multicolumn{8}{c}{PSNR $\uparrow$ }                                                                 & \multirow{2}{*}{\begin{tabular}[c]{@{}l@{}}Avg. \\ PSNR\end{tabular} $\uparrow$ } & \multirow{2}{*}{\begin{tabular}[c]{@{}l@{}}Avg. \\ SSIM\end{tabular} $\uparrow$ } & \multirow{2}{*}{\begin{tabular}[c]{@{}l@{}}Avg. \\ LPIPS\end{tabular} $\downarrow$ } \\
        \cmidrule(r){3-10}
                                                                                    &                           & \small{bouncingballs} & \small{hellwarrior} & \small{hook}   & \small{jumpingjacks} & \small{lego}   & \small{mutant} & \small{standup} & \small{trex}   &                            &                            &                             \\
        \midrule                                                                                    
        \multirow{3}{*}{15 views}                                                   & HexPlane                  & 26.56                 & 15.91               & \textbf{21.03}  & 20.35                 & \textbf{23.64}    & \textbf{23.40}    & 21.48             & 23.05             & 21.93                     & 0.921                      & 0.092                    \\                                                                                                                                                                                                                                                                                                                                
                                                                                    & K-Planes                  & 24.10                 & 15.88               & 19.59           & 20.97                 & 23.55             & 22.21             & 20.63             & \textbf{25.08}    & 21.50                     & 0.922                      & \textbf{0.086}           \\
                                                                                    &    Ours                   & \textbf{28.09}        & \textbf{16.48}      & 20.90           & \textbf{21.51}        & 23.54             & 23.38             & \textbf{21.87}    & 24.88             & \textbf{22.30}            & \textbf{0.925}             & 0.087                    \\
        \midrule                                                                                                                                                                
        \multirow{3}{*}{20 views}                                                   & HexPlane                  & 28.45                 & 16.85               & 22.30           & 20.87                 & \textbf{23.73}    & 25.02             & \textbf{23.73}    & 24.45             & 23.18                     & 0.929                      & 0.082                    \\
                                                                                    & K-Planes                  & 25.43                 & 17.25               & 21.07           & 21.40                 & 23.12             & 25.01             & 21.01             & 25.84             & 22.58                     & 0.931                      & \textbf{0.070}           \\
                                                                                    &     Ours                  & \textbf{31.15}        & \textbf{17.99}      & \textbf{22.67}  & \textbf{22.58}        & 23.49             & \textbf{25.86}    & 23.55             & \textbf{26.04}    & \textbf{23.93}            & \textbf{0.935}             & 0.072                    \\
        \midrule
        \multirow{3}{*}{25 views}                                                   & HexPlane                  & 30.49                 & 17.61               & 23.10           & 22.85                 & 24.29             & 25.81             & 23.74             & 25.30             & 24.15                     & 0.935                      & 0.074                    \\
                                                                                    & K-Planes                  & 28.29                 & 9.18$^*$            & 22.01           & 22.49                 & \textbf{24.33}    & 26.02             & 22.77             & \textbf{26.37}    & 22.68                     & 0.929                      & 0.107                    \\
                                                                                    & Ours                      & \textbf{34.61}        & \textbf{19.21}      & \textbf{23.82}  & \textbf{24.46}        & 23.78             & \textbf{26.75}    & \textbf{26.07}    & 26.29             & \textbf{25.34}            & \textbf{0.941}             & \textbf{0.063}           \\
        \midrule
        \multirow{3}{*}{Full views}                                                 & HexPlane                  & 39.21                 & 23.92               & 27.97           & 30.53                 & 24.74             & 32.19             & \textbf{33.09}    & 30.02             & 30.15                     & 0.964                      & 0.039                    \\
                                                                                    & K-Planes                  & 39.76                 & 24.57               & 28.10           & 31.07                 & \textbf{25.13}    & \textbf{32.42}    & 32.99             & \textbf{30.25}    & \textbf{30.54}            & \textbf{0.967}             & \textbf{0.033}           \\
                                                                                    & Ours                      & \textbf{40.25}        & \textbf{24.63}      & \textbf{28.50}  & \textbf{31.70}        & 25.09             & 31.19             & 31.45             & 29.76             & 30.20                     & 0.960                      & 0.049                    \\
        \bottomrule
        \multicolumn{8}{l}{$^*$ indicates the model does not converge}
    
    \end{tabular}
    }%
\end{table*}

\paragraph{Variance of PSNR on the static NeRF datasets.} 
We elaborate on the variance of PSNR for each instance in the static NeRF dataset in \autoref{app_table_psnr_variance}. Specifically, we examine the variance of PSNR across all test viewpoints in the static NeRF dataset. The total variance of PSNR across all images is calculated using 8,000 images from 8 scenes, each with 200 test viewpoints and five trials. 

Examining \autoref{app_table_psnr_variance}, while FreeNeRF records the lowest variance across nearly all scenes except for Ficus and Mic, it consistently lags behind the top score across all scenes. This observation highlights FreeNeRF tends to underfit. Notably, in scenes like Ficus and Mic where caputring intricate and delicate signals such as thin leaves or mesh structures is crucial, FreeNeRF struggles. This suggest that FreeNeRF is stable, but it faces  difficulties in effectively rendering complex details effectively. 
On the other hand, methods like iNGP and TensoRF exhibit poor average PSNR scores such as Drums and Ship, and achieve low variances. This pattern indicates a failure to learn effectively, resulting in meaningless output and negligible difference between training and test views due to overfitting. Although these methods sometimes perform well in scenes with simple geometry like Hotdog and Lego, they often lack stability and underperform in scenarios requiring sparse input handling for NeRF applications.

However, as stated in \autoref{subsec_stability}, our method consistently shows low variances in most scenes—except for hotdog and materials, where we achieve the highest PSNR—we emphasize the distinction of our approach in terms of both stability and superior capability compared to K-Planes.
\begin{table*}[!ht]
\caption{Variance of PSNR($\downarrow$) on the static NeRF dataset.}
    \label{app_table_psnr_variance}
    \centering
\begin{tabular}{ccccccccc|c}
\toprule
         & Chair & Drums & Ficus & Hotdog & Lego  & Materials & Mic   & Ship  & Total \\
\midrule
FreeNeRF & 5.07  & 1.72  & 8.72  & 11.16  & 6.42  & 2.13      & 17.28 & 6.48  & 17.31 \\
iNGP     & 8.43  & 3.13  & 1.37  & 12.39  & 7.78  & 2.96      & 9.02  & 6.03  & 23.95 \\
TensoRF  & 10.88 & 2.86  & 2.17  & 13.11  & 10.27 & 2.82      & 8.06  & 5.71  & 23.22 \\
K-Planes & 10.74 & 2.55  & 2.83  & 27.19  & 10.76 & 2.99      & 9.23  & 11.48 & 19.61 \\
Ours     & 3.82  & 2.82  & 2.00  & 16.38  & 8.72  & 4.09      & 9.68  & 6.01  & 18.23 \\
\bottomrule
\end{tabular}
\end{table*}

\newpage

\section{Experiments on Varying Denoising Hyper-parameter $\lambda_1$}
\label{app_varying_lambda1}

We evaluated the role of Total Variation or Laplacian smoothing regularization for TensoRF, HexPlane, and the proposed method by incrementally increasing the regularization parameter $\lambda_1$ from $0.0001$ to $1.0$, each step multiplying by a factor of $10$. The results, displayed in \autoref{app_table_denoising_static}, show that our proposed method outperforms all scenarios in both static and dynamic NeRF datasets, except at  $\lambda_1=0.001$ in the static dataset. At this value, TensoRF achieved the highest PSNR of 24.98, but it struggled to converge at higher $\lambda_1$ values, highlighting its sensitivity and difficulties in training robustly with varying regularization strengths.
For dynamic NeRF datasets, where time sparsity presents additional challenges, HexPlane's performance varied between PSNR scores of 21.95 to 24.15, whereas our method ranged from 24.67 to 25.74, indicating a lesser dependency on denoising regularization. This suggests that the coordinate networks used in our method provide robust regularization for multi-plane encoding, reducing the necessity for intensive hyperparameter tuning across different scenes.
Furthermore, excessive regularization led to undesirable modifications such as color disturbances in TensoRF's rendering of the \texttt{ship} scene with $\lambda_1=0.001$, as evidenced by our results. Our method, on the other hand, maintained near-optimal performance across a broad range of $\lambda_1$ values without necessitating excessive denoising regularization, thanks to its ability to capture global contexts through coordinate-based networks. 

\begin{figure}[!h]{
    \centering
    \includegraphics[width=1.00\columnwidth]{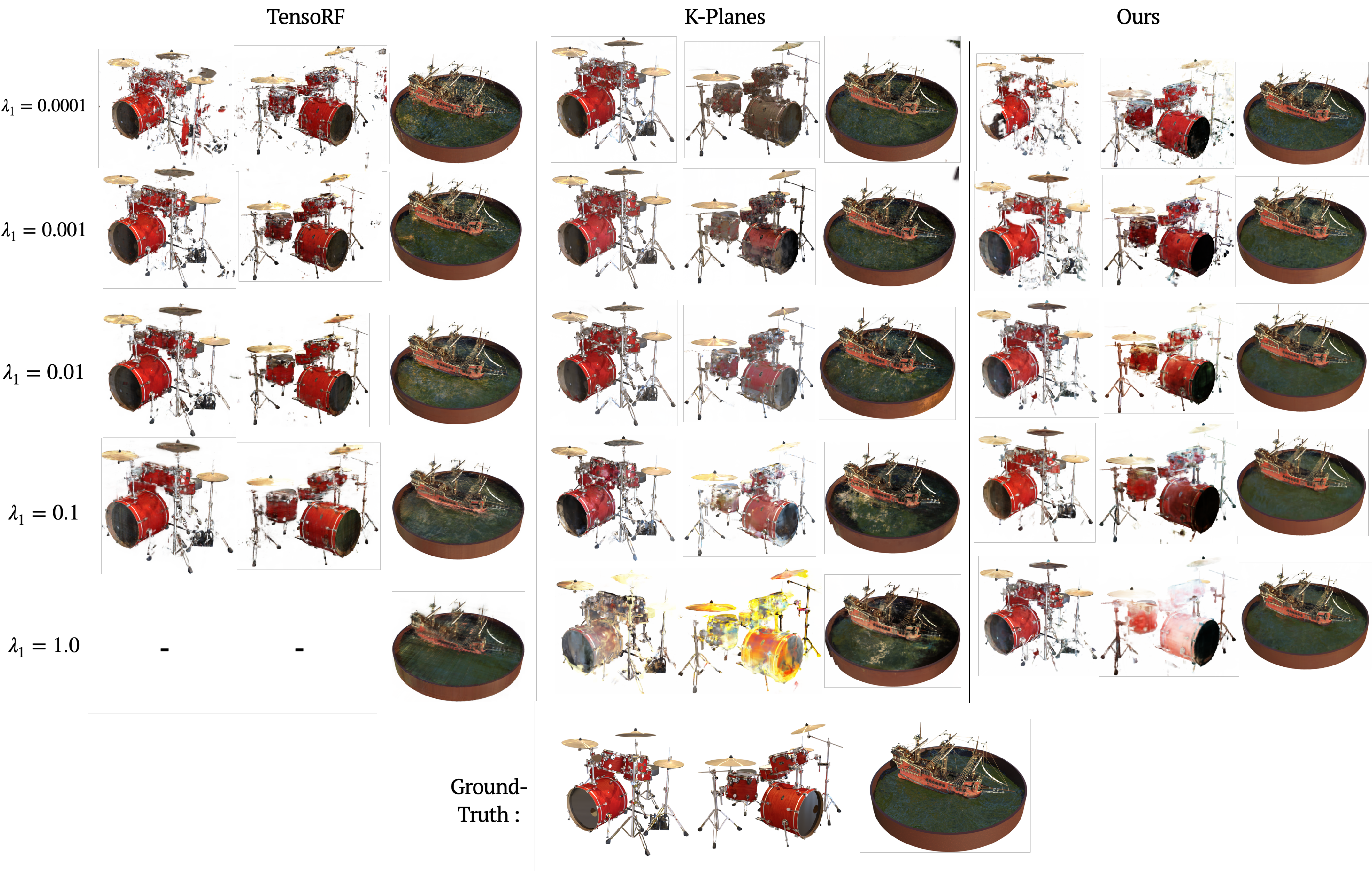}    
    \caption{Rendered images of \texttt{drums} and \texttt{ship} cases in the static NeRF dataset by TensoRF, K-Planes and ours with varying $\lambda_1$. We select the 61st and 36th images for the \texttt{drums} scene and the 156th images for the \texttt{ship}.}
    \label{app_fig_exp_static_nerf_varying_lambda}
}    
\end{figure}

\begin{figure}[!ht]{
    \centering
    \includegraphics[width=0.95\columnwidth]{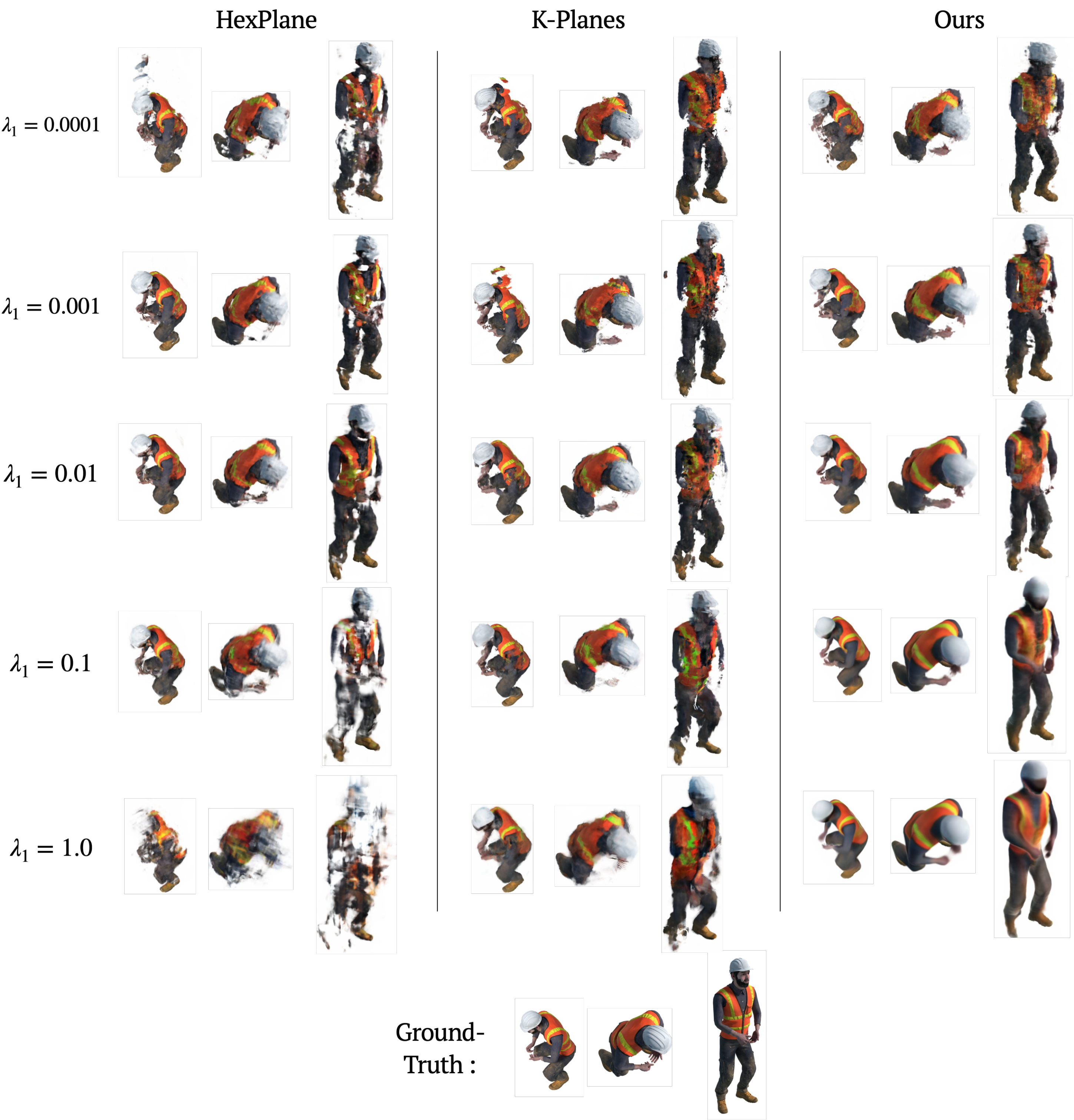}    
    \caption{Rendered images of \texttt{standup} cases in the dynamic NeRF dataset by HexPlane and ours with varying $\lambda_1$. We evaluate \{0, 10, 19\}th views in the test dataset.}
    \label{app_fig_exp_dynamic_nerf_varying_lambda}
}    
\end{figure}

To provide deeper analysis, \autoref{app_fig_exp_static_nerf_varying_lambda} qualitatively illustrates how the dependency of TensoRF, K-Planes and the proposed model on the denoising weight affects performance. While TensoRF could reduce floating artifacts with appropriate denoising levels, excessive regularization led to unwanted color distortions, necessitating a delicate balance in regularization weight tuning.
In contrast, our model displayed consistent performance across various $\lambda_1$ settings without introducing undesigned artifacts, even as denoising intensity increased. This capability stems from the model's reliance on coordinate features that anchor low-frequency information, providing a stable base for robust reconstructions.
In dynamic scenes, as input sparsity increases, \autoref{app_table_denoising_dynamics} and \autoref{app_fig_exp_dynamic_nerf_varying_lambda} affirm that denoising regularization alone is insufficient in both HexPlane and K-Planes. For example, they exhibit degraded performance compared to our model and the optimal $\lambda_1$ values for each model fluctuate across scenes. This underscores a heavy reliance on regularization.
Conversely, our proposed model demonstrated high adaptability and robustness regardless of the regularization intensity, underscoring that our feature-fusion strategy is inherently robust against sparse inputs. Consequently, regularization synergizes with our model design, aiding in more realistic rendering without producing undesired artifacts and effectively handling sparse input cases.
\begin{table}[!h]
    \caption{The comparison of Ours, K-Planes and TensoRF in the static NeRF dataset. We conduct experiments varying the value of $\lambda_1$. All models are trained using 8 views. The hyphen means that the model is not converged.}
    \label{app_table_denoising_static}
    \centering
    \resizebox{0.98\columnwidth}{!}{%
        \begin{tabular}{lccccccccccc}
        \toprule
        \multirow{2}{*}{Models} & \multicolumn{8}{c}{PSNR $\uparrow$}                                                     & \multirow{2}{*}{\begin{tabular}[c]{@{}l@{}}Avg. \\ PSNR\end{tabular} $\uparrow$}    & \multirow{2}{*}{\begin{tabular}[c]{@{}l@{}}Avg. \\ SSIM\end{tabular} $\uparrow$}    & \multirow{2}{*}{\begin{tabular}[c]{@{}l@{}}Avg. \\ LPIPS\end{tabular} $\downarrow$} \\
        \cmidrule(r){2-9}
                                        & chair  & drums  & ficus  & hotdog & lego   & materials & mic    & ship                                    &                               &                               &            \\
        \midrule            
        TensoRF ($\lambda_1=0.0001)$    & 27.15  & 16.85  & 21.84  & 29.35  & 28.03  & 21.41     & 26.99  & 21.17                                   & 24.10                        & 0.880                         & 0.103      \\
        TensoRF ($\lambda_1= 0.001)$    & 28.24  & 19.94  & 21.94  & 29.46  & 29.04  & 22.03     & 26.62  & 22.58                                   & 24.98                        & 0.898                         & 0.078      \\
        TensoRF ($\lambda_1=  0.01)$    & 27.97  & 20.04  & -      & 29.22  & 28.93  & 21.98     & -      & 23.24                                   & -                            & -                             & -          \\
        TensoRF ($\lambda_1=   0.1)$    & -      & 19.80  & -      & 28.12  & 27.11  & 21.37     & -      & 21.93                                   & -                            & -                             & -          \\
        TensoRF ($\lambda_1=   1.0)$    & -      & -      & -      & 25.97  & 24.55  & 19.36     & -      & 22.24                                   & -                            & -                             & -          \\
        \midrule
        K-Planes ($\lambda_1=0.0001)$   & 27.16  & 20.50  & 23.82  & 27.75  & 26.29  & 19.87     & 27.46  & 21.68                                   & 24.31                        & 0.897                         & 0.083      \\
        K-Planes ($\lambda_1= 0.001)$   & 27.08  & 20.20  & 23.26  & 27.94  & 27.06  & 20.02     & 26.76  & 21.94                                   & 24.28                        & 0.900                         & 0.081      \\
        K-Planes ($\lambda_1=  0.01)$   & 27.10  & 20.27  & 22.62  & 27.64  & 26.48  & 20.59     & 27.08  & 22.46                                   & 24.28                        & 0.899                         & 0.082      \\
        K-Planes ($\lambda_1=   1.0)$   & 23.54  & 17.53  & 22.31  & 27.08  & 26.01  & 19.74     & 26.46  & 21.93                                   & 23.64                        & 0.893                         & 0.090      \\
        K-Planes ($\lambda_1=   0.1)$   & 25.98  & 19.60  & 20.72  & 26.11  & 24.15  & 19.09     & 24.56  & 20.73                                   & 22.05                        & 0.876                         & 0.112      \\
        \midrule   
        Ours ($\lambda_1=0.0001)$       & 27.79  & 17.67  & 19.30  & 28.62  & 24.81  & 21.49     & 26.16  & 23.57                                   & 23.68                        & 0.884                         & 0.111      \\
        Ours ($\lambda_1= 0.001)$       & 27.94  & 19.04  & 20.07  & 29.13  & 27.26  & 21.85     & 26.93  & 23.55                                   & 24.47                        & 0.893                         & 0.091      \\
        Ours ($\lambda_1=  0.01)$       & 27.61  & 19.21  & 20.17  & 29.51  & 27.31  & 21.55     & 26.74  & 24.27                                   & 24.55                        & 0.895                         & 0.098      \\
        Ours ($\lambda_1=   0.1)$       & 27.07  & 19.60  & 20.55  & 29.09  & 25.43  & 22.50     & 26.13  & 23.56                                   & 24.23                        & 0.889                         & 0.108      \\
        Ours ($\lambda_1=   1.0)$       & 25.12  & 17.99  & 19.89  & 27.64  & 22.74  & 21.98     & 25.55  & 23.05                                   & 22.99                        & 0.876                         & 0.136      \\
        \bottomrule
    \end{tabular}
    }%
\end{table}

\begin{table}[!h]
    \caption{The comparison of Ours and HexPlane in the dynamic NeRF dataset. We conduct experiments varying the value of $\lambda_1$. All models are trained using 25 views. We use seed 0 for reproducibility.}
    \label{app_table_denoising_dynamics}
    \centering
    \resizebox{0.95\columnwidth}{!}{%
        \begin{tabular}{lccccccccccc}
        \toprule
        \multirow{2}{*}{Models} & \multicolumn{8}{c}{PSNR $\uparrow$}                                                     & \multirow{2}{*}{\begin{tabular}[c]{@{}l@{}}Avg. \\ PSNR\end{tabular} $\uparrow$}    & \multirow{2}{*}{\begin{tabular}[c]{@{}l@{}}Avg. \\ SSIM\end{tabular} $\uparrow$}    & \multirow{2}{*}{\begin{tabular}[c]{@{}l@{}}Avg. \\ LPIPS\end{tabular} $\downarrow$} \\
        \cmidrule(r){2-9}
                                        & \small{bouncingballs} & \small{hellwarrior}   & \small{hook}      & \small{jumpingjacks}  & \small{lego}      & \small{mutant}    & \small{standup}   & \small{trex}                           &                              &                               &            \\
        \midrule            
        HexPlane ($\lambda_1=0.0001)$   & 28.80                 & 16.32                 & 21.44             & 21.98                 & 23.81             & 24.67             & 21.30             & 24.34                                  & 22.83                        & 0.926                         & 0.082      \\
        HexPlane ($\lambda_1= 0.001)$   & 30.25                 & 16.86                 & 22.61             & 22.70                 & 24.21             & 26.03             & 23.07             & 25.19                                  & 23.86                        & 0.934                         & 0.070      \\
        HexPlane ($\lambda_1=  0.01)$   & 30.49                 & 17.61                 & 23.10             & 22.86                 & 24.29             & 25.81             & 23.74             & 25.30                                  & 24.15                        & 0.935                         & 0.074        \\
        HexPlane ($\lambda_1=   0.1)$   & 29.64                 & 18.24                 & 22.13             & 21.75                 & 23.72             & 24.63             & 23.08             & 24.53                                  & 23.46                        & 0.928                         & 0.090      \\
        HexPlane ($\lambda_1=   1.0)$   & 26.60                 & 17.79                 & 21.05             & 19.73                 & 23.53             & 22.75             & 19.88             & 24.30                                  & 21.95                        & 0.917                         & 0.117      \\
        \midrule
        K-Planes ($\lambda_1=0.0001)$   & 29.39                 & 16.72                 & 22.69             & 23.98                 & 24.03             & 26.42             & 24.47             & 26.88                                  & 24.32                        & 0.937                         & 0.074      \\
        K-Planes ($\lambda_1= 0.001)$   & 29.22                 & 17.92                 & 22.29             & 22.73                 & 24.12             & 26.20             & 23.22             & 26.35                                  & 24.01                        & 0.939                         & 0.061      \\
        K-Planes ($\lambda_1=  0.01)$   & 29.38                 & 18.29                 & 22.33             & 22.78                 & 23.82             & 26.18             & 23.02             & 26.33                                  & 24.02                        & 0.938                         & 0.062        \\
        K-Planes ($\lambda_1=   0.1)$   & 28.85                 & 17.53                 & 21.52             & 22.52                 & 24.02             & 26.00             & 22.74             & 25.25                                  & 23.55                        & 0.931                         & 0.074      \\
        K-Planes ($\lambda_1=   1.0)$   & 25.29                 & 17.90                 & 20.99             & 21.63                 & 23.61             & 25.06             & 21.73             & 24.73                                  & 22.62                        & 0.928                         & 0.087      \\
        \midrule            
        Ours ($\lambda_1=0.0001)$       & 32.80                 & 18.34                 & 23.39             & 23.18                 & 23.79             & 26.33             & 23.77             & 25.77                                  & 24.67                        & 0.936                         & 0.071      \\
        Ours ($\lambda_1= 0.001)$       & 34.13                 & 19.01                 & 23.90             & 24.72                 & 23.92             & 26.86             & 24.26             & 26.22                                  & 25.38                        & 0.942                         & 0.062      \\
        Ours ($\lambda_1=  0.01)$       & 33.71                 & 19.69                 & 23.83             & 24.77                 & 24.20             & 26.89             & 25.96             & 26.86                                  & 25.74                        & 0.943                         & 0.064        \\
        Ours ($\lambda_1=   0.1)$       & 32.91                 & 19.80                 & 24.08             & 24.63                 & 24.36             & 26.85             & 27.69             & 26.40                                  & 25.84                        & 0.941                         & 0.074      \\
        Ours ($\lambda_1=   1.0)$       & 32.21                 & 19.52                 & 24.33             & 24.36                 & 23.51             & 26.23             & 27.18             & 26.05                                  & 25.42                        & 0.937                         & 0.088      \\
        \bottomrule
    \end{tabular}
    }%
\end{table}

\newpage

\clearpage
\section{Ablation Study on Residual Neural Radiance Fields}
\label{app_analysis_encoding}

\begin{figure}[!ht]{
    \centering
        \begin{subfigure}{0.42\columnwidth}
            \includegraphics[width=\columnwidth]{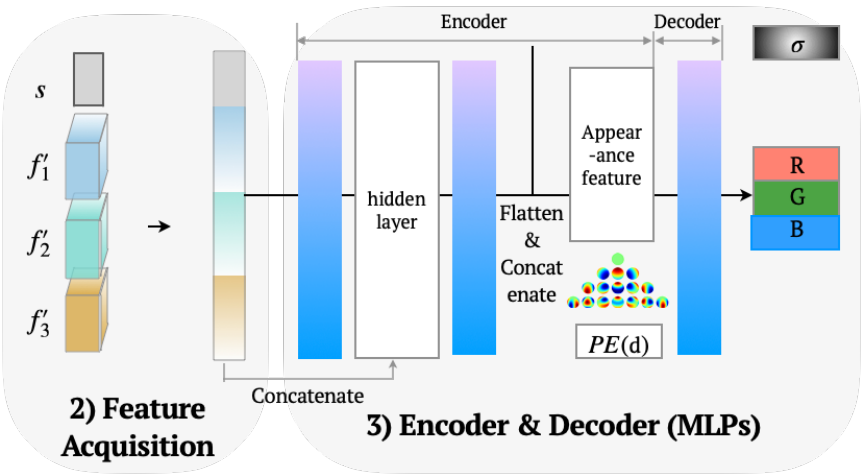}
            \caption{Ours}
            \label{app_fig_encoder_sturcture_ours}
        \end{subfigure}    
        \begin{subfigure}{0.42\columnwidth}
            \includegraphics[width=\columnwidth]{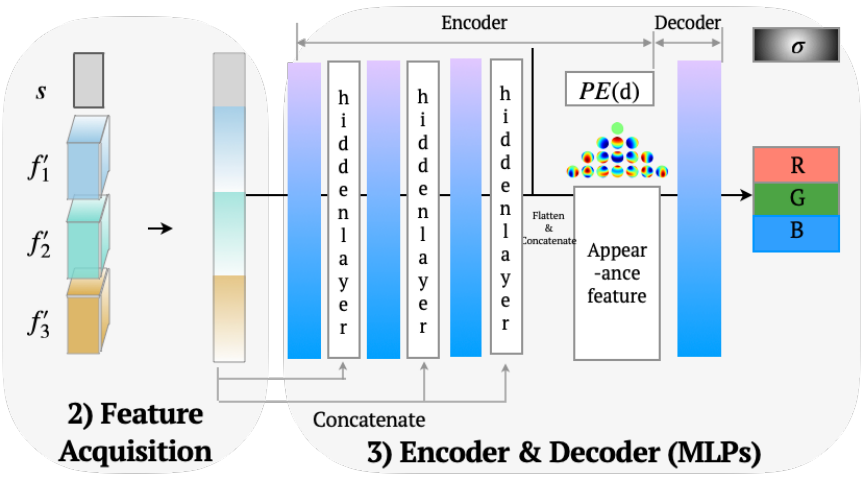}    
            \caption{Type 1}
        \label{app_fig_encoder_sturcture_type_1}
        
        \end{subfigure}    
        \begin{subfigure}{0.42\columnwidth}
            \includegraphics[width=\columnwidth]{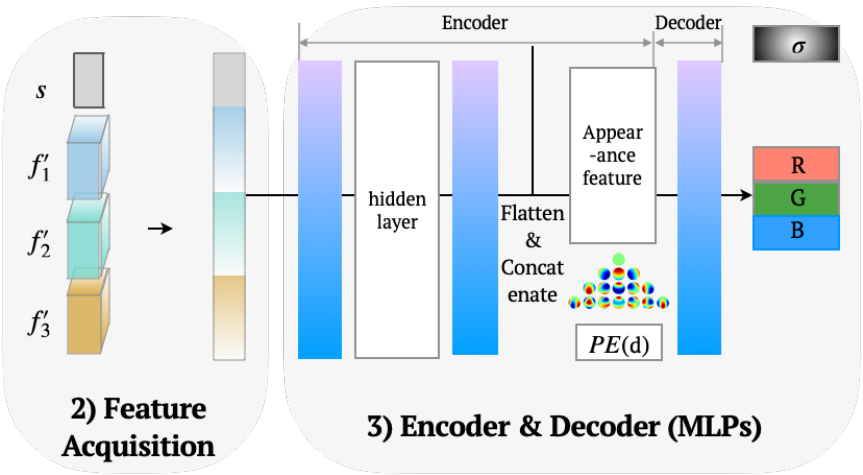}    
            \caption{Type 2}
            \label{app_fig_encoder_structure_type_2}
        \end{subfigure}    
        \begin{subfigure}{0.42\columnwidth}
            \includegraphics[width=\columnwidth]{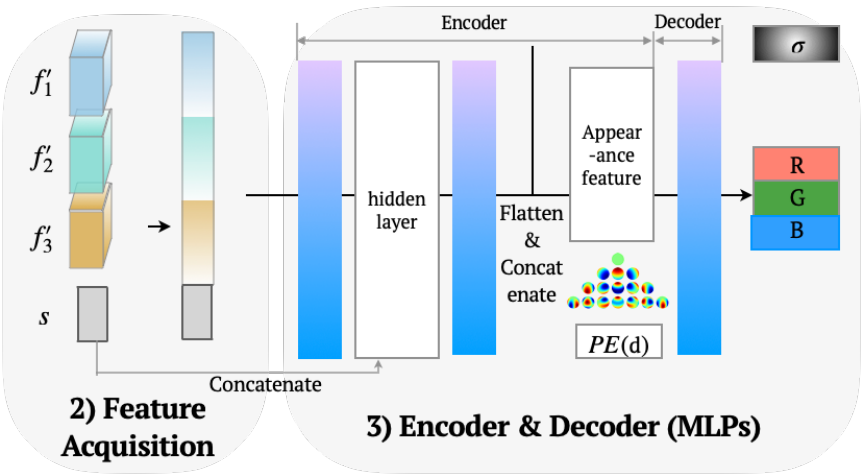}    
            \caption{Type 3}
            \label{app_fig_encoder_structure_type_3}
        \end{subfigure}    
        
    \caption{The graphical representation for encoder structures used in \autoref{app_table_residual_static} and \autoref{app_table_residual_dynamic}.}
    \label{app_encoder_structure}
}    
\end{figure}

The encoder of our model utilizes a skip-connection of fused features. To justify the design choice of our model, we compare the results of various encoder structures in static and dynamic cases. All possible candidates for encoder structures are listed, and their graphical representations are also included in \autoref{app_encoder_structure}. 

\begin{itemize}
    \item Type 1 : Skip connection lies on every layer
    \item Type 2 : No skip connection, and employs fully connected MLPs
    \item Type 3 : Skip connection, but only coordinate $s$ is concatenated.
\end{itemize}

\begin{figure}[!ht]{
    \centering
        \begin{subfigure}{0.45\columnwidth}
            \includegraphics[width=\columnwidth]{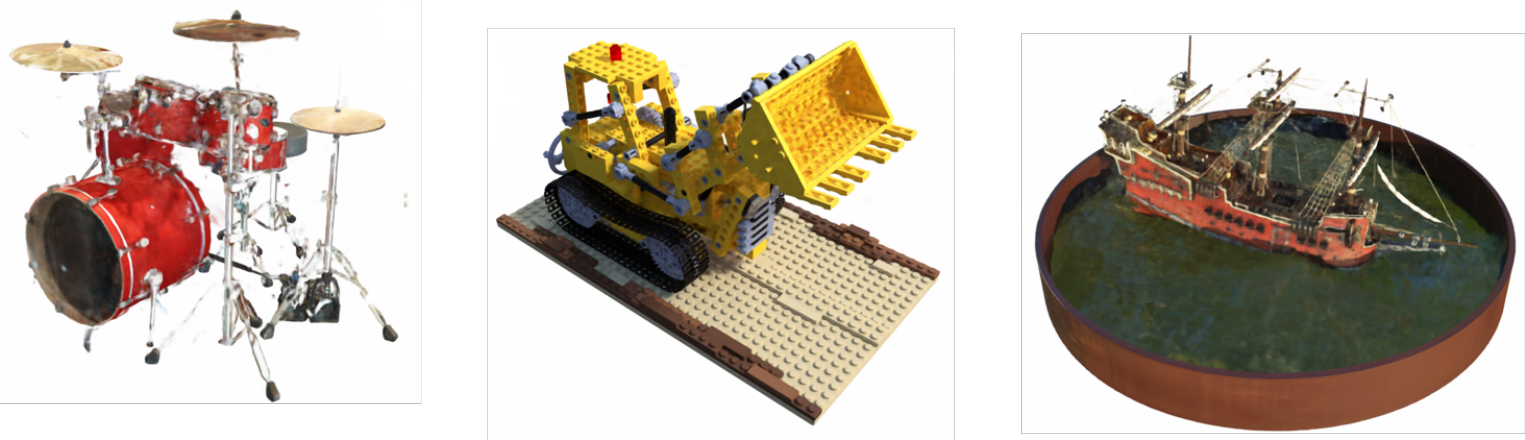}
            \caption{Ours}
            \label{app_encoder_ours_static}
        \end{subfigure}    
        \begin{subfigure}{0.45\columnwidth}
            \includegraphics[width=\columnwidth]{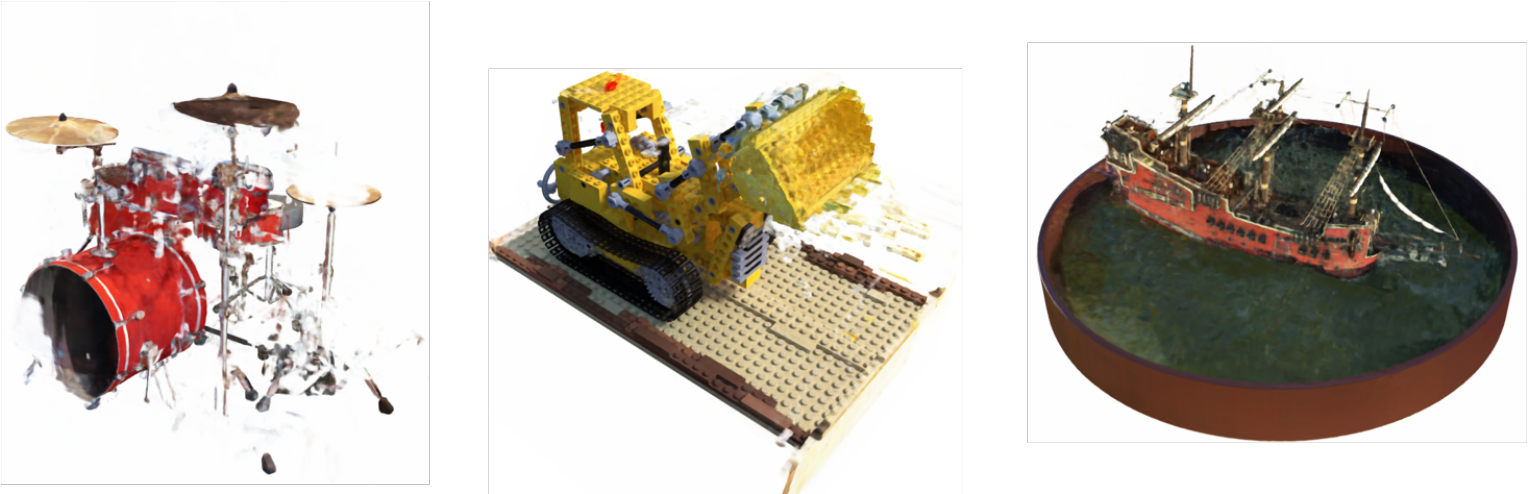}    
            \caption{Type 1}
        \label{app_encoder_type_1_static}
        
        \end{subfigure}    
        \begin{subfigure}{0.45\columnwidth}
            \includegraphics[width=\columnwidth]{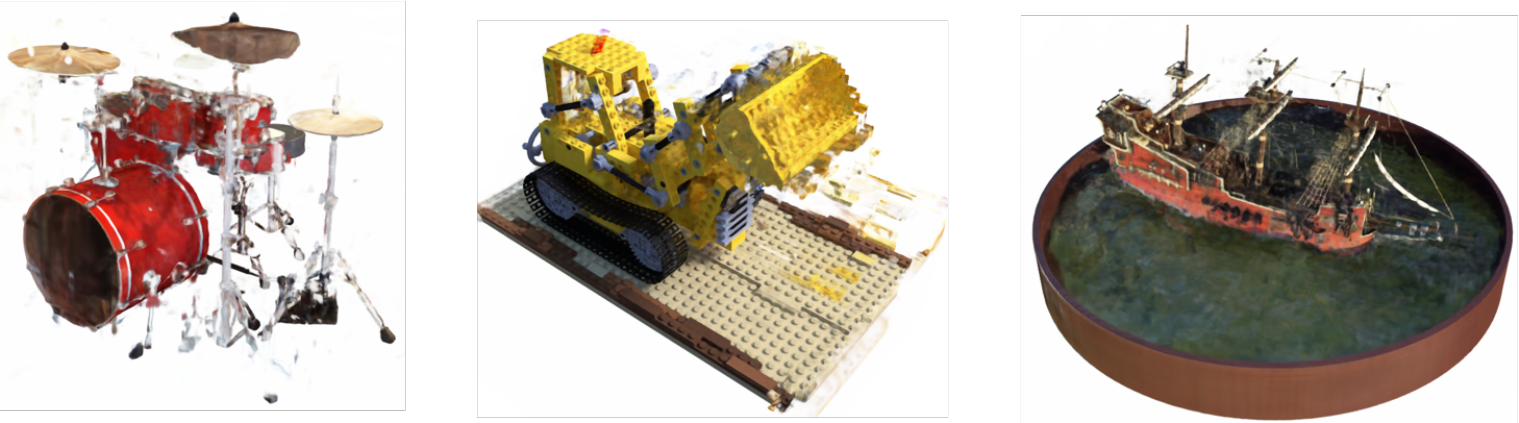}    
            \caption{Type 2}
            \label{app_encoder_type_2_static}
        \end{subfigure}    
        \begin{subfigure}{0.45\columnwidth}
            \includegraphics[width=\columnwidth]{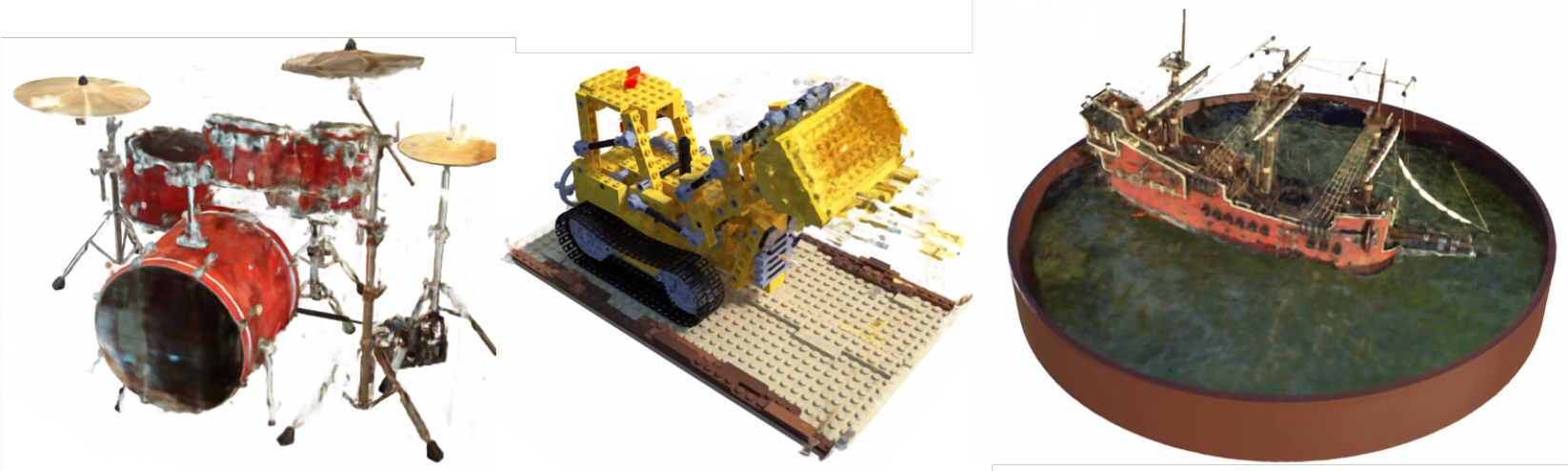}    
            \caption{Type 3}
            \label{app_encoder_type_3_static}
        \end{subfigure}    
        
    \caption{Rendered images are generated by alternating encoder structures. We selected the \texttt{drums}, \texttt{lego}, and \texttt{ship} scenes to follow the settings used in previous experiments.}
    \label{app_encoder_structure_rendering}
}    
\end{figure}

Through \autoref{app_table_residual_static} and \autoref{app_table_residual_dynamic}, we determine that the proposed model represents the optimal architecture. Their visual outcomes are illustrated in \autoref{app_encoder_structure_rendering}. Partiularly, in the case of dynamic NeRFs, inducing smoothness in the temporal axis is crucial. In Type 3, where only the coordinate feature is used for skip-connection, it shows slightly better performance than our model. However, our model design demonstrates robustness across both static and dynamic cases, confirming the suitability of our model design choices.

\begin{table}[!ht]
    \caption{The comparison of encoding structures. We evaluate four types of encoding structures including ours. All hyperparameters are consistent with those described in the original setting included \autoref{app_implementation_details}. All models are trained using 8 views in the static NeRF dataset. We use seed 0 for reproducibility.}
    \label{app_table_residual_static}
    \centering
    \resizebox{0.95\columnwidth}{!}{%
        \begin{tabular}{lccccccccccc}
        \toprule
        \multirow{2}{*}{Models} & \multicolumn{8}{c}{PSNR $\uparrow$}                                                     & \multirow{2}{*}{\begin{tabular}[c]{@{}l@{}}Avg. \\ PSNR\end{tabular} $\uparrow$}    & \multirow{2}{*}{\begin{tabular}[c]{@{}l@{}}Avg. \\ SSIM\end{tabular} $\uparrow$}    & \multirow{2}{*}{\begin{tabular}[c]{@{}l@{}}Avg. \\ LPIPS\end{tabular} $\downarrow$} \\
        \cmidrule(r){2-9}
                                        & chair  & drums  & ficus  & hotdog & lego   & materials & mic    & ship                                    &                               &                               &            \\
        \midrule            
        Ours                            & 28.15  & 20.09  & 20.04  & 29.43  & 27.58  & 22.06     & 26.41  & 24.18                                  & 24.74                        & 0.898                         & 0.089      \\
        Type 1                          & 23.83  & 17.85  & 19.14  & 18.45  & 20.54  & 12.97     & 14.61  & 22.78                                  & 18.77                        & 0.844                         & 0.179      \\
        Type 2                          & 26.15  & 18.02  & 19.53  & 17.78  & 19.73  & 11.72     & 18.06  & 22.87                                  & 19.23                        & 0.848                         & 0.171        \\
        Type 3                          & 25.16  & 19.40  & 19.33  & 17.94  & 20.88  & 11.85     & 14.62  & 23.35                                  & 19.07                        & 0.843                         & 0.175      \\
        \bottomrule
    \end{tabular}
    }%
\end{table}

\begin{table}[!ht]
    \caption{The comparison of encoding structures. We evaluate four types of encoding structures including ours. All models are trained using 25 views in the dynamic NeRF dataset. All hyperparameters are consistent with those described in the original setting included \autoref{app_implementation_details}. We use seed 0 for reproducibility.}
    \label{app_table_residual_dynamic}
    \centering
    \resizebox{0.95\columnwidth}{!}{%
        \begin{tabular}{lccccccccccc}
        \toprule
        \multirow{2}{*}{\textbf{Models}} & \multicolumn{8}{c}{PSNR $\uparrow$}                                                     & \multirow{2}{*}{\begin{tabular}[c]{@{}l@{}}Avg. \\ PSNR\end{tabular} $\uparrow$}    & \multirow{2}{*}{\begin{tabular}[c]{@{}l@{}}Avg. \\ SSIM\end{tabular} $\uparrow$}    & \multirow{2}{*}{\begin{tabular}[c]{@{}l@{}}Avg. \\ LPIPS\end{tabular} $\downarrow$} \\
        \cmidrule(r){2-9}
                                        & \small{bouncingballs} & \small{hellwarrior}   & \small{hook}      & \small{jumpingjacks}  & \small{lego}      & \small{mutant}    & \small{standup}   & \small{trex}                                    &                               &                               &            \\
        \midrule            
        Ours                            & 33.83                 & 18.93                 & 23.54             & 24.24                 & 23.69             & 26.59             & 26.06             & 26.05                                             & 25.37                        & 0.942                         & 0.063      \\
        Type 1                          & 33.99                 & 18.01                 & 24.01             & 24.26                 & 23.91             & 26.95             & 24.55             & 26.56                                             & 25.28                        & 0.941                         & 0.064      \\
        Type 2                          & 33.35                 & 18.08                 & 23.82             & 24.58                 & 24.08             & 26.85             & 24.46             & 26.84                                             & 25.26                        & 0.941                         & 0.063        \\
        Type 3                          & 32.74                 & 18.64                 & 24.24             & 24.83                 & 23.99             & 27.08             & 25.17             & 26.81                                             & 25.44                        & 0.942                         & 0.062      \\
        \bottomrule
    \end{tabular}
    }%
\end{table}

\section{Visualization of Disentangling Coordinate Network and Tensorial Feature}
\label{app_disentangle}

\begin{figure}[t]{
    \centering
        \begin{subfigure}{0.75\columnwidth}
            \includegraphics[width=\columnwidth]{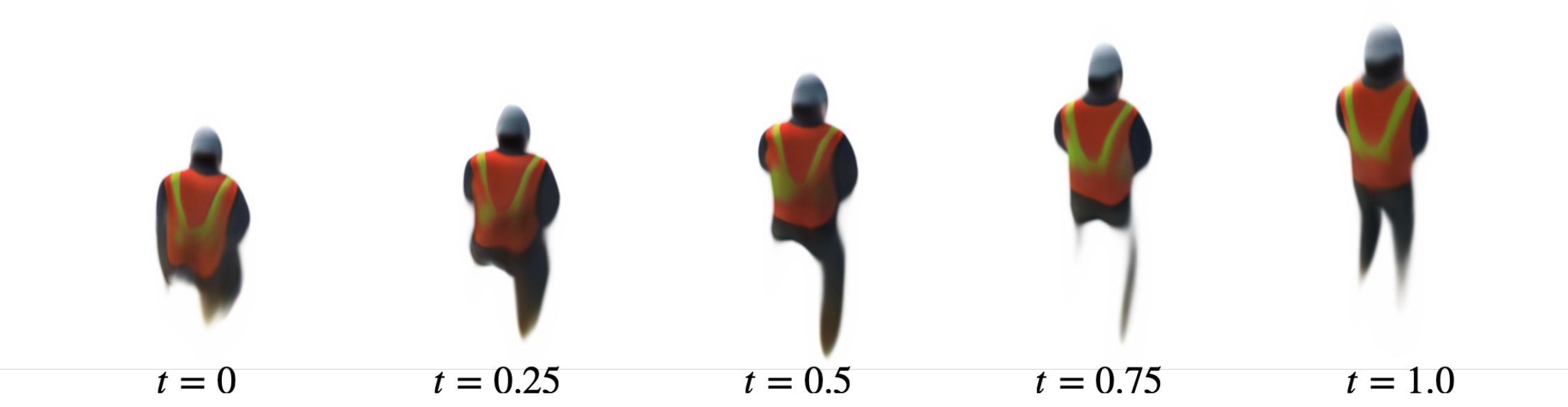}
            \caption{Rendered image by ours with only coordinate networks}
            \label{app_fig_standup_by_mlp}
        \end{subfigure}    
        
        \begin{subfigure}{0.75\columnwidth}
            \includegraphics[width=\columnwidth]{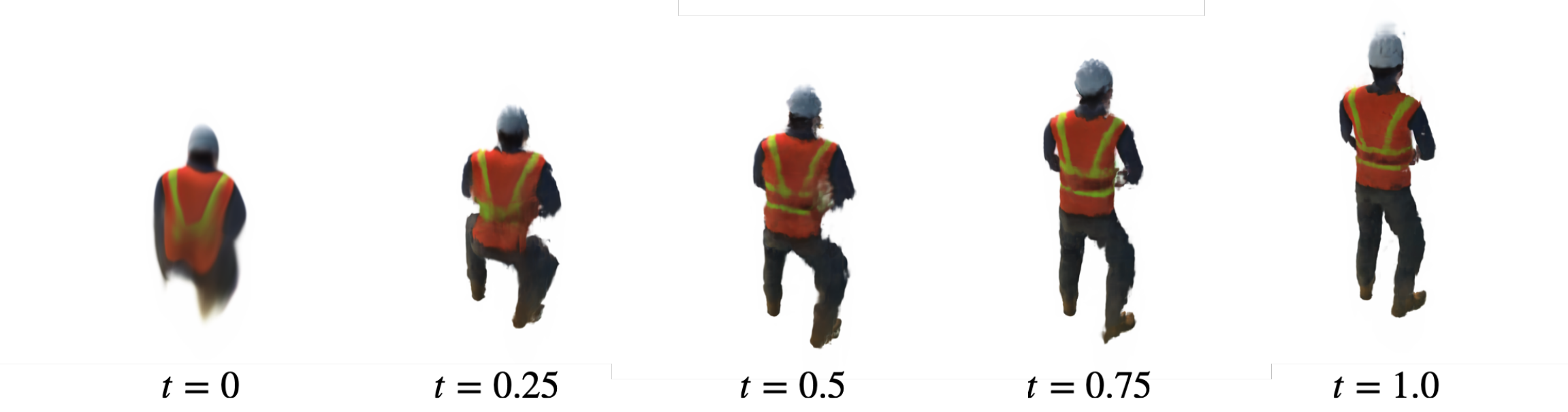}    
            \caption{Rendered image by ours with curriculum weighting}
            \label{app_fig_standup_by_curriculum_weight}
        \end{subfigure}    

        \begin{subfigure}{0.75\columnwidth}
            \includegraphics[width=\columnwidth]{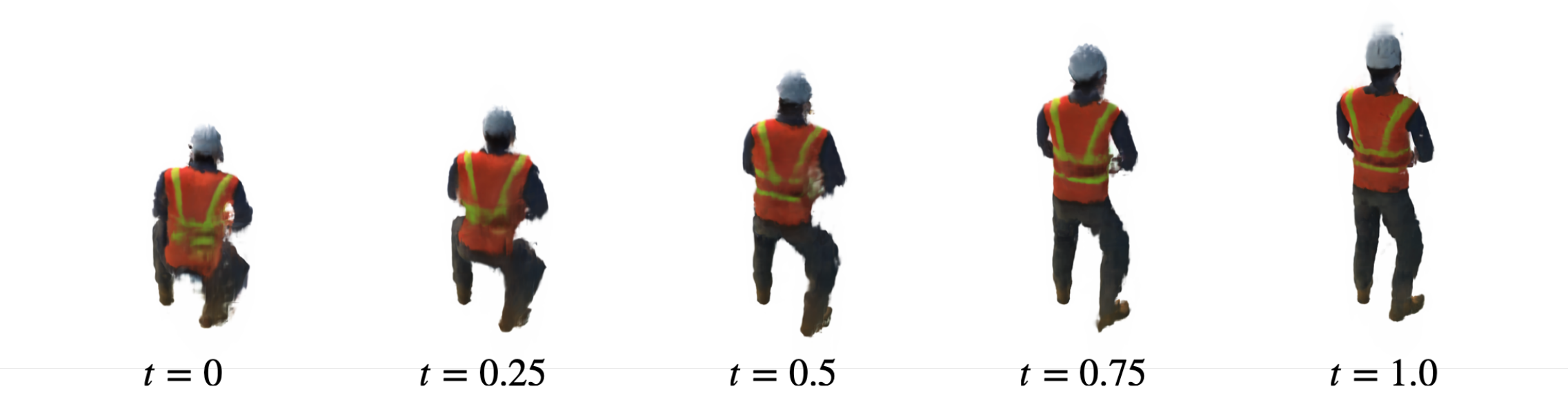}    
            \caption{Rendered image with full engangement of multi-plane}
            \label{app_fig_standup_by_multiple_plane}
        \end{subfigure}    
        
    \caption{Rendering results using different feature combinations. We show rendering results from three distinct combination of encoding features, (a) using only coordinates, (b) coordinates with progressively activating multi-plane encoding, and (c) full features. $t$ indicates the timesteps normalized to 1, and we use \texttt{standup} scene.}
    \label{app_fig_analysis_disentanglement}
}    
\end{figure}

Our model employs a disentanglement strategy that separates global shape and detail into coordinate networks and multi-plane features respectively. Additionally, we implement a progressive learning approach on the channel axis within plane features, enhancing the model's ability to cover details from global to local scales. We demonstrate disentangling features into: (1) heterogeneous two features and (2) channel-wise distinct features. First, we explore disentanglement between heterogeneous features. An ablation study on dynamic NeRFs with 25 training views helped us understand the role of coordinate-based networks in our method. Testing the model solely with coordinate networks, as seen in \autoref{app_fig_standup_by_mlp}, revealed that they capture the scene’s global context, such as object shapes and significant motions. Although \citet{lindell2022bacon} indicated a potential dominance of high-frequency features, our model maintains a balance, demonstrating the synergistic function of coordinate network and multi-plane feature.

\begin{figure}[!ht]{
    \centering
        \begin{subfigure}{0.45\columnwidth}
            \includegraphics[width=\columnwidth]{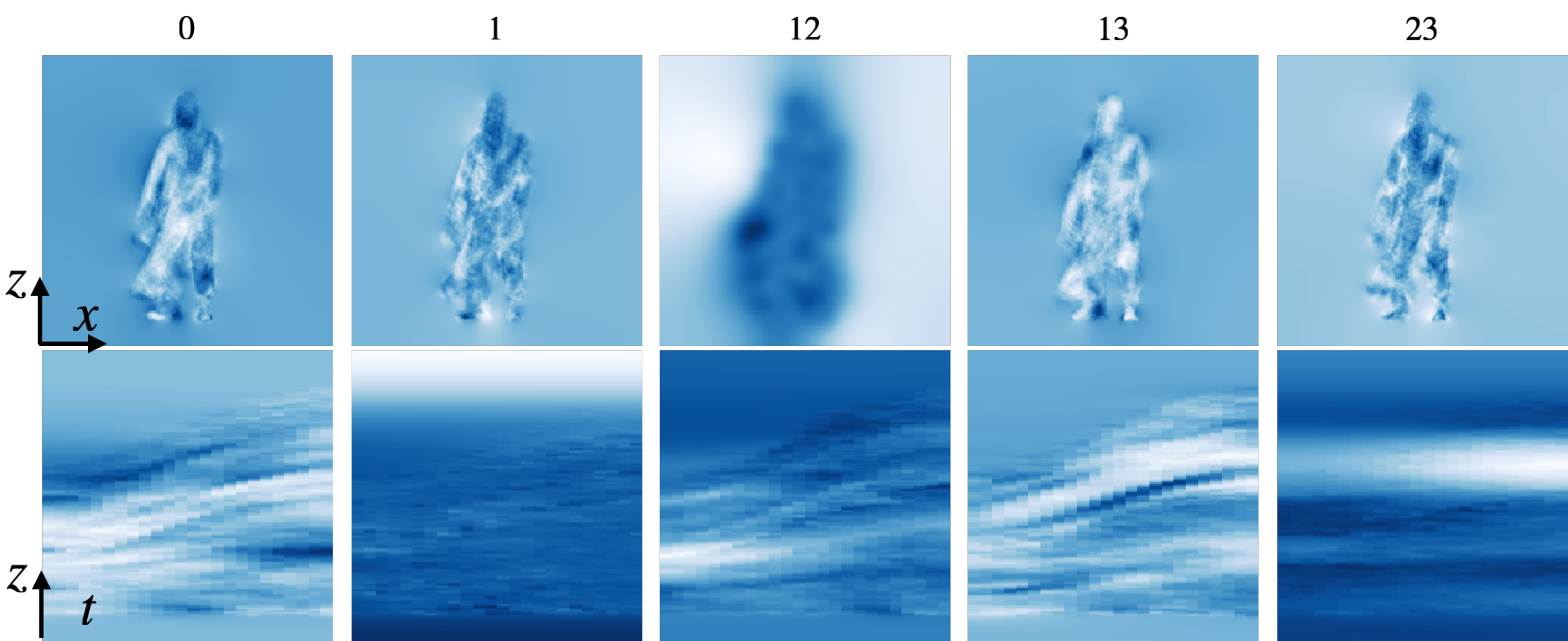}
            \caption{HexPlane($\lambda_1=0.001$) trained on full views}
            \label{app_fig_experiment_hexplane_full_view_analysis}
        \end{subfigure}    
        \begin{subfigure}{0.45\columnwidth}
            \includegraphics[width=\columnwidth]{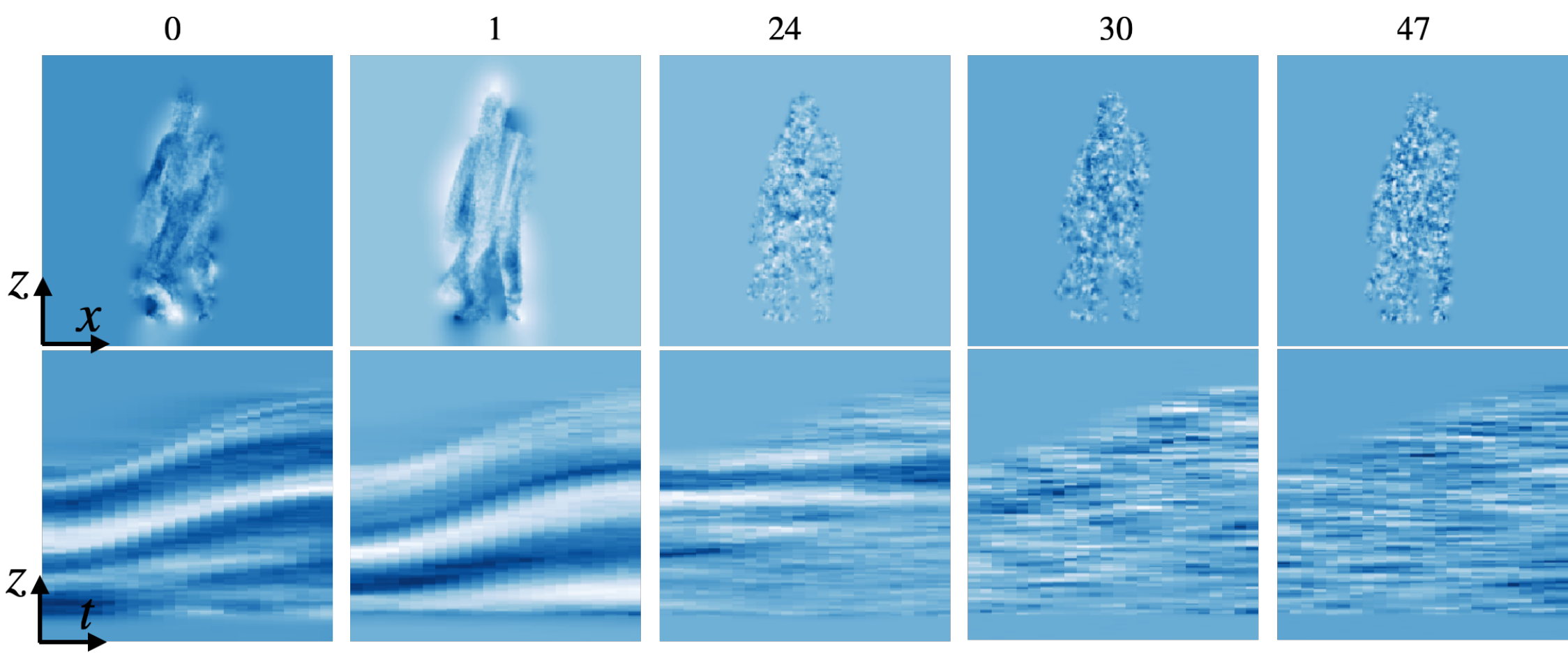}    
            \caption{Ours trained on full views}
            \label{app_fig_experiment_TensorRefine_full_view_analysis}
        \end{subfigure}    

        \begin{subfigure}{0.45\columnwidth}
            \includegraphics[width=\columnwidth]{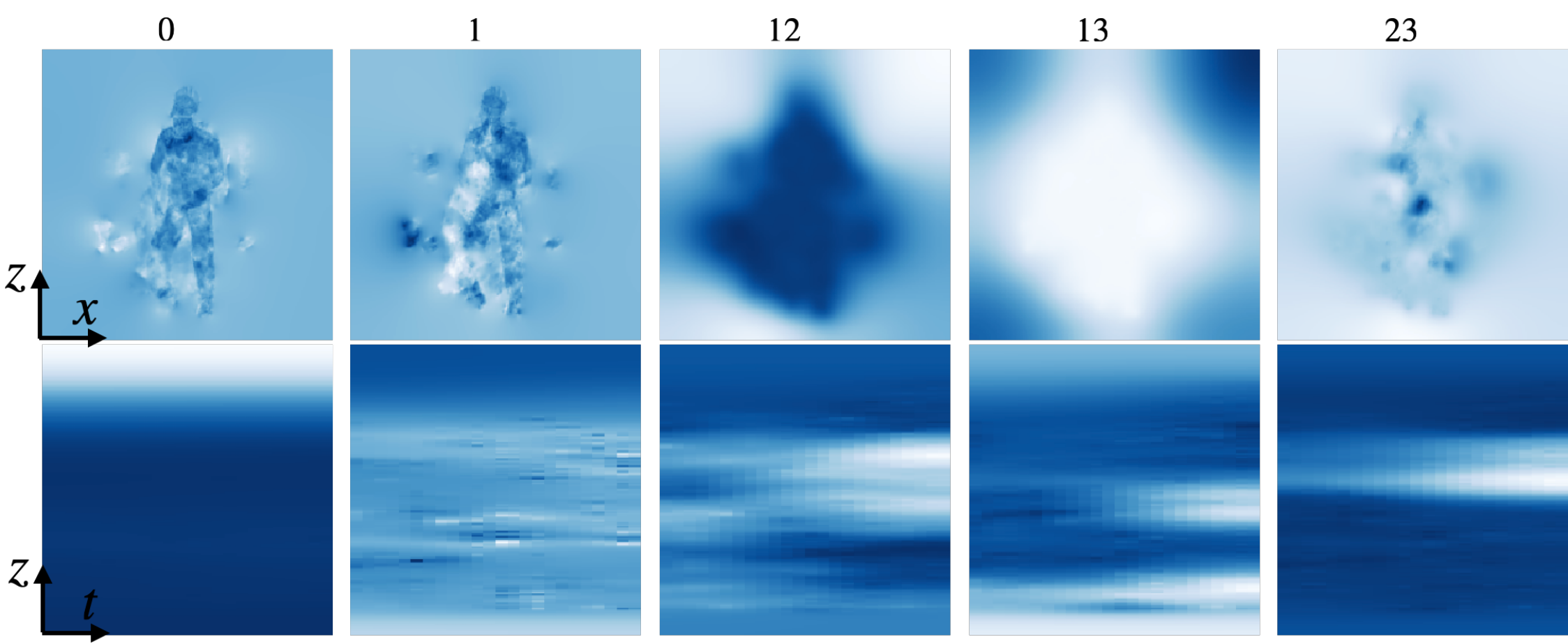}    
            \caption{HexPlane($\lambda_1=0.001$) trained on 25 views}
            \label{app_fig_experiment_hexplane_25_view_analysis}
        \end{subfigure}
        \begin{subfigure}{0.45\columnwidth}
            \includegraphics[width=\columnwidth]{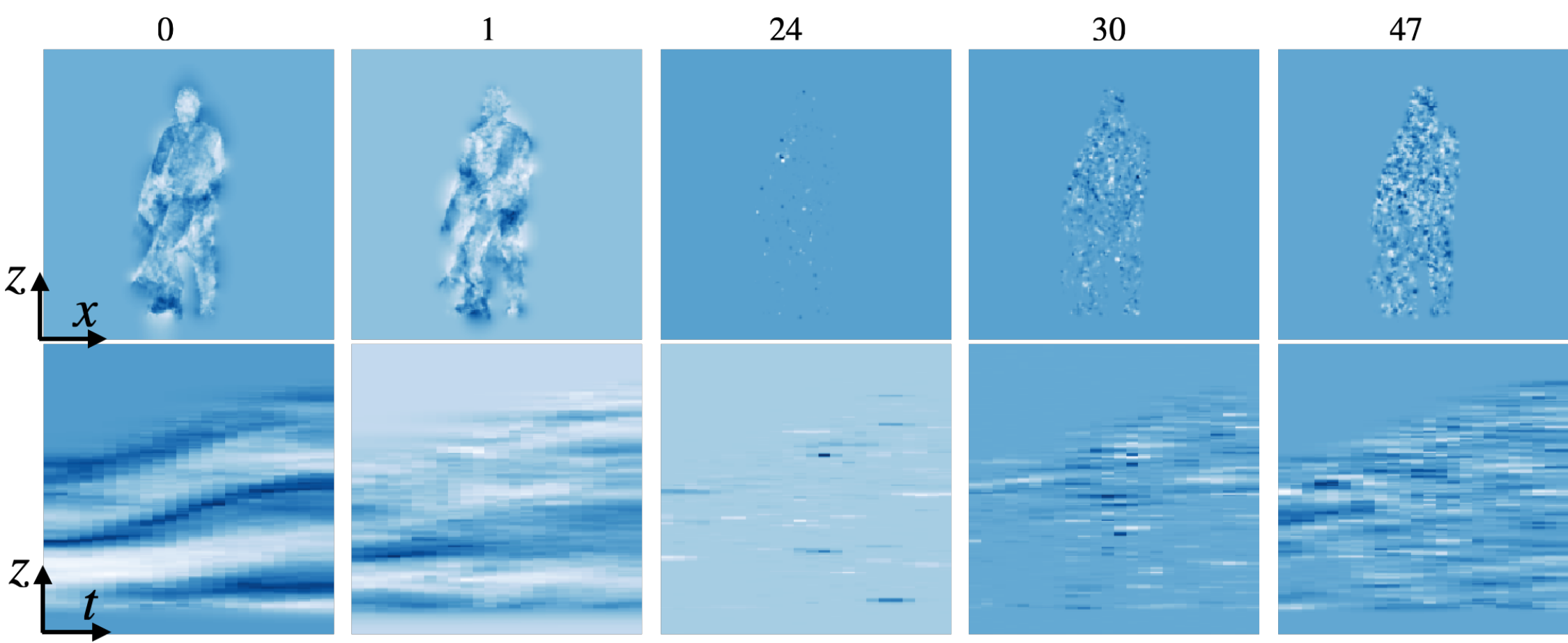}    
            \caption{Ours trained on 25 views}
            \label{app_fig_experiment_TensorRefine_25_view_analysis}
        \end{subfigure}    
        
    \caption{Visualization of plane encoding features. We visualize 5 representative features from the plane encodings of Hexplane and Ours trained on \texttt{standup} scene.}
    \label{app_fig_multi_plane_analysis}
}    
\end{figure}

Second, we assess channel-wise disentanglement among plane features. We compared multi-plane features of HexPlane and our method, trained on both full views and 25 views of the \texttt{standup} scenes, as depicted in \autoref{app_fig_multi_plane_analysis}. In the \texttt{standup} scenario, the $z-x$ plane should represent the front shape of the person, and the $z-t$ plane should depict upward movement. HexPlane features for full views, shown in \autoref{app_fig_experiment_hexplane_full_view_analysis}, do not differentiate well between global shape information and intricate local details, with some channels overlapping in learned information. In contrast, our method distinguishes between multi-plane features along the channel axis, enhancing expressiveness as shown in \autoref{app_fig_experiment_TensorRefine_full_view_analysis}. Our method maintains this differentiation even with fewer views, as observed in \autoref{app_fig_experiment_hexplane_25_view_analysis}. Moreover, HexPlane exhibits floating artifacts and lacks visibility of upward movement on the time axis, whereas our model shows consistency in both full and reduced view scenarios, as seen in \autoref{app_fig_experiment_TensorRefine_25_view_analysis}. Particularly, our model can selectively learn channel-wise, indicating minimal impact through flat representations. These findings confirm that our dual disentanglement strategies for distinct features and channel-wise distinctions effectively enable learning from global to detailed features, enhancing expressiveness and robustness in handling sparse inputs.

\clearpage
\section{Comparison of the Number of Parameters and Analysis of Training/Rendering Times}
\label{app_training_time}

We evaluated our model against iNGP, TensoRF, K-Planes, and other methods using static and dynamic NeRF datasets, limiting the training steps to 15,000 and using only 8 views for training. Rendering time was measured over 200 frames for static NeRFs and 20 frames for dynamic scenarios. We demonstrate the model parameters in \autoref{app_table_training_params} and \autoref{app_table_training_params_dynamic}, where bracketed numbers indicate channel counts in multi-plane features. K-Planes model, featuring multi-resolution multi-plane characteristics, calculates total channels as the product of resolutions and channel dimension per resolution.

\autoref{app_table_training_params} shows that our method achieves comparable performance to TensoRF with the optimized $\lambda_1$, yet uses significantly fewer parameters. TensoRF, despite optimal performance at 64 channels, faces instability during training and rendering. Reducing channels to 20 causes convergence issues in scenes like \{chair, ficus, mic\}, highlighting its limitations with sparse inputs. Although K-Planes appears stable, it underperforms and demands more parameters. Our method, while slower in rendering compared to TensoRF and K-Planes, excels in stability during both training and rendering, ensuring consistent performance even with fewer channels, which is ideal for sparse inputs. 
In the case of dynamic NeRFs, our method outperforms all baselines even with reduced parameters (approximately 1 million) as shown in \autoref{app_table_training_params_dynamic}. While slightly slower in rendering, the performance difference is minimal. This is attributed to the low frequency detail handled by coordinate network, eliminating the need to apply multi-plane features for low frequency detail. Consequently, we accomplish minimal use of parameters because multi-plane features are tasked solely with high-frequency details. 

Despite its slower rendering speed, especially in complex scenes like \texttt{ficus} and \texttt{drums}, we argue that our model focuses on refining NeRF architecture, ensuring compatibility with fast training NeRFs frameworks without compromising on stability during sparse input training. Future efforts could explore using sequential MLPs like \texttt{tinycudann} to enhance rendering speed, although this might introduce instability given our current focus on maintaining robustness in sparse scenarios. Overall, our experiment highlights the robustness of our method in maintaining stable training and consistent rendering quality, proving crucial in conditions with sparse inputs where reliability across various settings is essential.

\begin{table}[!ht]
    \caption{
        Comparison of the number of parameters and analysis of training and rendering time in static NeRFs. $\dagger$ indicates the optimized hyper-parameter $\lambda_1=0.001$ used.
    }  
    \label{app_table_training_params}
    \centering
        \begin{tabular}{lcccc}
        \toprule
        \multicolumn{1}{c}{\begin{tabular}[c]{@{}c@{}}Model \\ Name\end{tabular}} & \begin{tabular}[c]{@{}c@{}}\# Params\\ {[}M{]}\end{tabular} & \begin{tabular}[c]{@{}c@{}}Avg. \\ PSNR\end{tabular} & \begin{tabular}[c]{@{}c@{}}Avg. Training \\ Time [min]\end{tabular} & \begin{tabular}[c]{@{}c@{}}Avg. Rendering \\ Time [min]\end{tabular} \\
        \midrule
        iNGP \footnotesize{(T=19)}                                         & 11.7M                                                         & 19.26                                                & 7.60                                                         & 0.82                                                         \\
        iNGP \footnotesize{(T=18)}                                         &6.4M                                                         & 19.99                                                & 6.40                                                         & 0.91                                                         \\
        K-Planes \footnotesize{(3*16)}                                         & 17M                                                         & 23.95                                                & 17.61                                                         & 6.83                                                         \\
        K-Planes \footnotesize{(2*16)}                                         & 4.4M                                                        & 23.16                                                & 13.72                                                         & 6.51                                                         \\
        TensoRF$^\dagger$ \footnotesize{(64)}                                            & 17.3M                                                       & 25.23                                                & 7.72                                                          & 7.82                                                         \\
        TensoRF$^\dagger$ \footnotesize{(20)}                                            & 6.1M                                                        & -                                                    & -                                                             & -                                                              \\
        Ours \footnotesize{(48)}                                               & 6.0M                                                        & 24.36                                                & 31.16                                                         & 46.02                                                        \\
        Ours \footnotesize{(24)}                                               & 3.0M                                                        & 23.74                                                & 24.06                                                         & 40.76                                                       \\
        \bottomrule
        \end{tabular}
\end{table}

\begin{table}[!ht]
    \caption{
        Comparison of the number of parameters and analysis of training and rendering time in dynamic NeRFs. 
    }  
    \label{app_table_training_params_dynamic}
    \centering
        \begin{tabular}{lcccc}
        \toprule
        \multicolumn{1}{c}{\begin{tabular}[c]{@{}c@{}}Model \\ Name\end{tabular}} & \begin{tabular}[c]{@{}c@{}}\# Params\\ {[}M{]}\end{tabular} & \begin{tabular}[c]{@{}c@{}}Avg. \\ PSNR\end{tabular} & \begin{tabular}[c]{@{}c@{}}Avg. Training \\ Time [min]\end{tabular} & \begin{tabular}[c]{@{}c@{}}Avg. Rendering \\ Time [min]\end{tabular} \\
        \midrule
        K-Planes \footnotesize{(3*32)}                                         & 18.6M                                                       & 23.85                                                & 18.93                                                         & 0.83                                                         \\
        K-Planes \footnotesize{(3*4)}                                         & 1.9M                                                        & 23.41                                                & 13.29                                                         & 0.78                                                         \\
        HexPlane \footnotesize{(72)}                                           & 9.7M                                                        & 24.00                                                & 6.78                                                          & 0.60                                                         \\
        HexPlane \footnotesize{(6)}                                            & 0.8M                                                        & 22.08                                                & 6.38                                                          & 0.68                                                              \\
        Ours \footnotesize{(48)}                                               & 3.4M                                                        & 25.17                                                & 12.22                                                         & 2.14                                                        \\
        Ours \footnotesize{(12)}                                               & 1.0M                                                        & 25.10                                                & 8.77                                                          & 1.73                                                       \\
        \bottomrule
        \end{tabular}
\end{table}

\clearpage
\section{Experimental Result of Real-world Dataset : Tanks and Temples}
\label{app_tanksandtemples}

\begin{figure}[!ht]{
    \centering
        \begin{subfigure}{0.70\columnwidth}
            \includegraphics[width=\columnwidth]{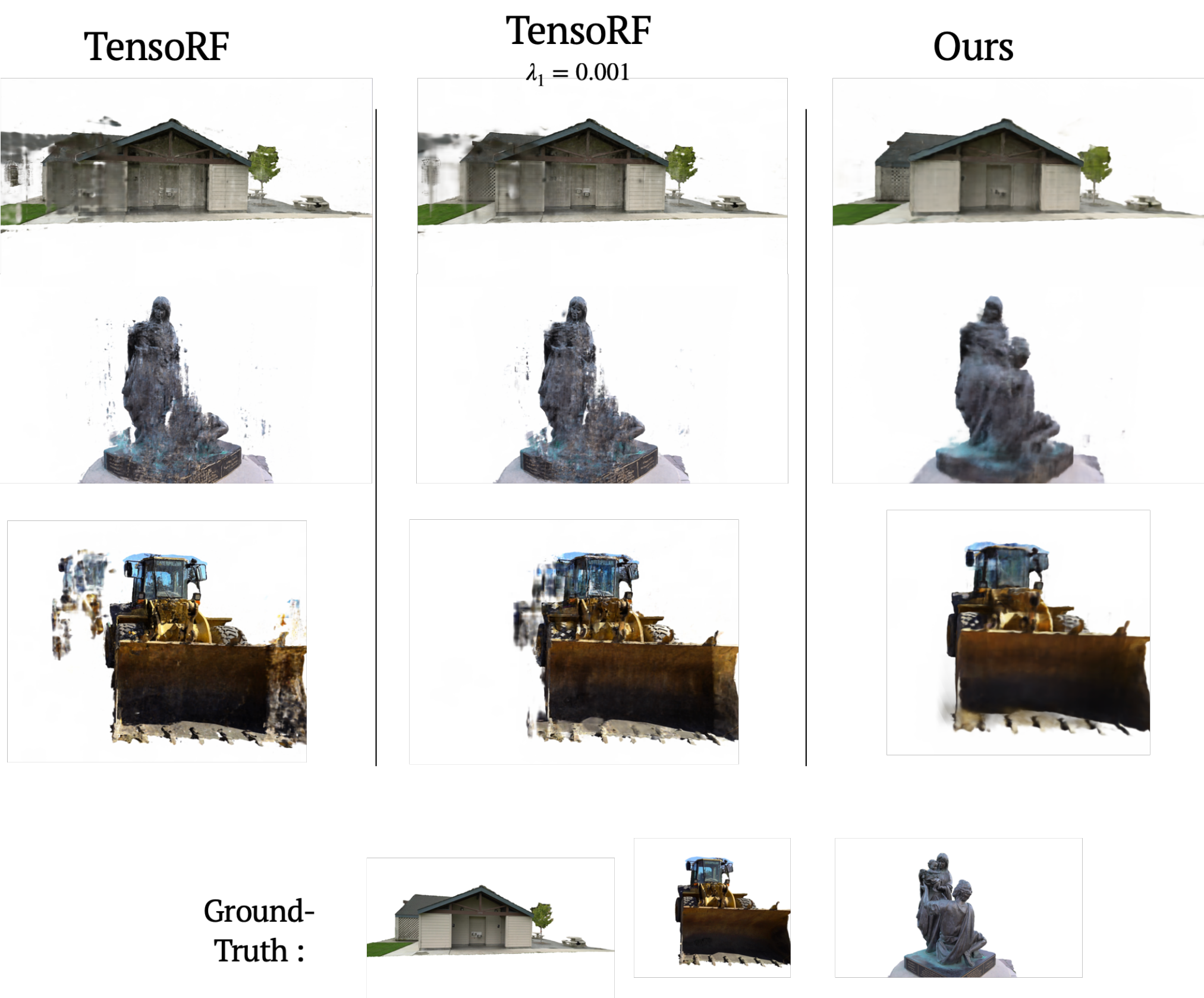}
            \label{fig_app_tanksandtemple}
        \end{subfigure}    
    \caption{The qualitative results of baselines and the proposed method on the Tanks and Temples dataset. We specifically show $\{47,11,12\}$-th images of \texttt{Barn}, \texttt{Family} and \texttt{Caterpillar} from the test dataset. 
    We use $\{7,10,15\}$ percentiles of the training views for the \texttt{Caterpillar} \texttt{Barn} and \texttt{Family} scenes, respectively.}
    \label{fig_result_tanksandtemple}
}    
\end{figure}

The proposed method is also evaluated on the real-world Tanks and Temples dataset \cite{Knapitsch2017}, where it was compared with the baseline TensoRF models, including the optimized setting ($\lambda_1 = 0.001$). We focus on how each method handles the preservation of global context in scenes. As shown in the \autoref{fig_result_tanksandtemple}, the proposed method consistently produces better rendered images than the baselines by preserving the global context. This is crucial when dealing with sparse input situations where maintaining the overall structure and shape of objects is essential even in real-world situations. Despite TensoRF's focus on local details leading to partial but incomplete reconstructions seen in the case of \texttt{family}, our method excels in capturing the overall scene composition. This ability ensures that the larger structure and form of objects in the scene are accurately reconstructed, even at the cost of some finer details. Therefore, we demonstrate that the proficiency of our method becomes more apparent under conditions of sparse input data, making it particularly suitable for real-world applications where input data might be limited or incomplete.

\begin{figure}[!ht]{
    \centering
        \begin{subfigure}{0.40\columnwidth}
            \includegraphics[width=\columnwidth]{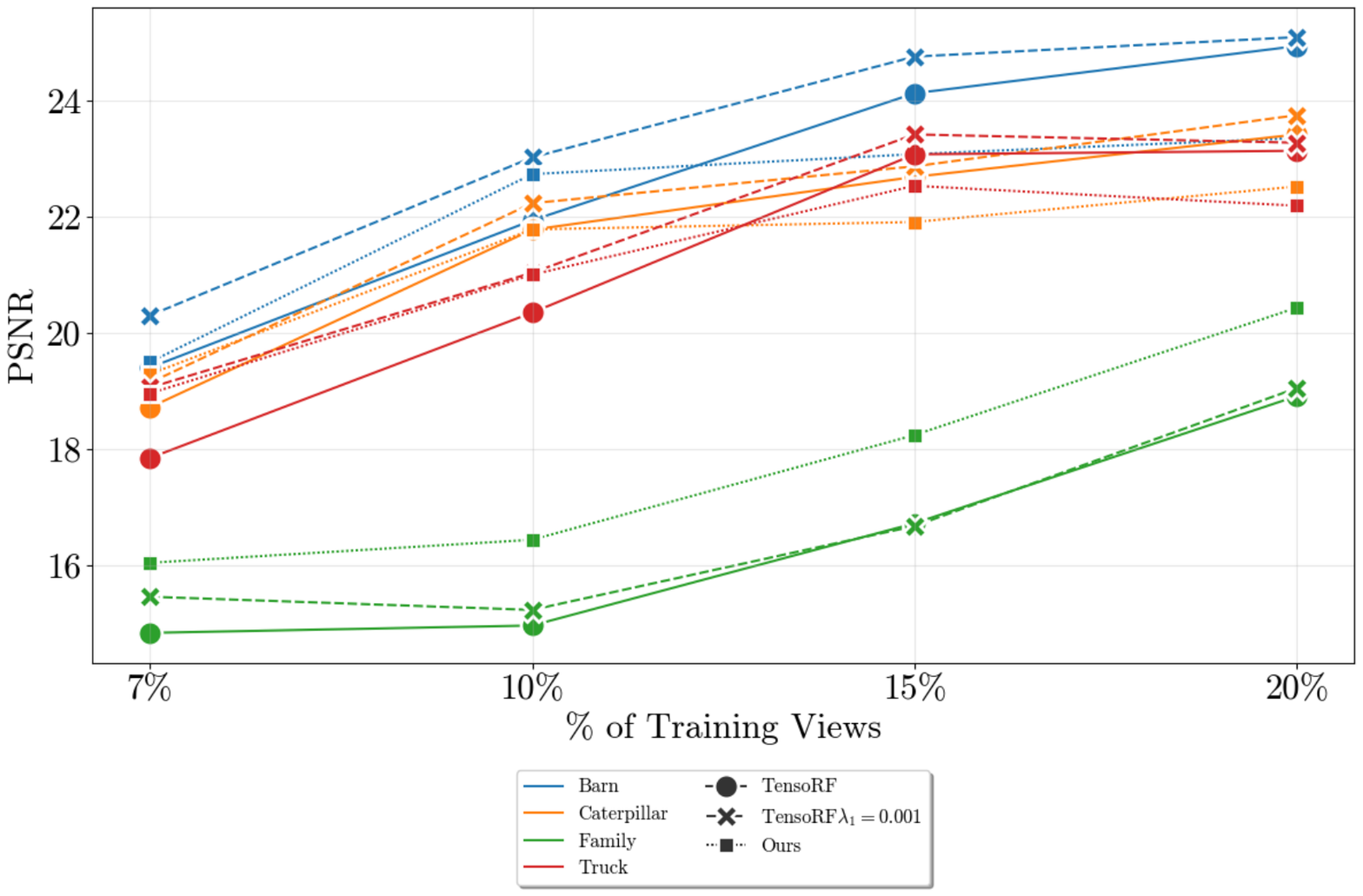}
            \caption{PSNR}
            \label{fig_app_tanksandtemple_psnr}
        \end{subfigure}    
        \begin{subfigure}{0.40\columnwidth}
            \includegraphics[width=\columnwidth]{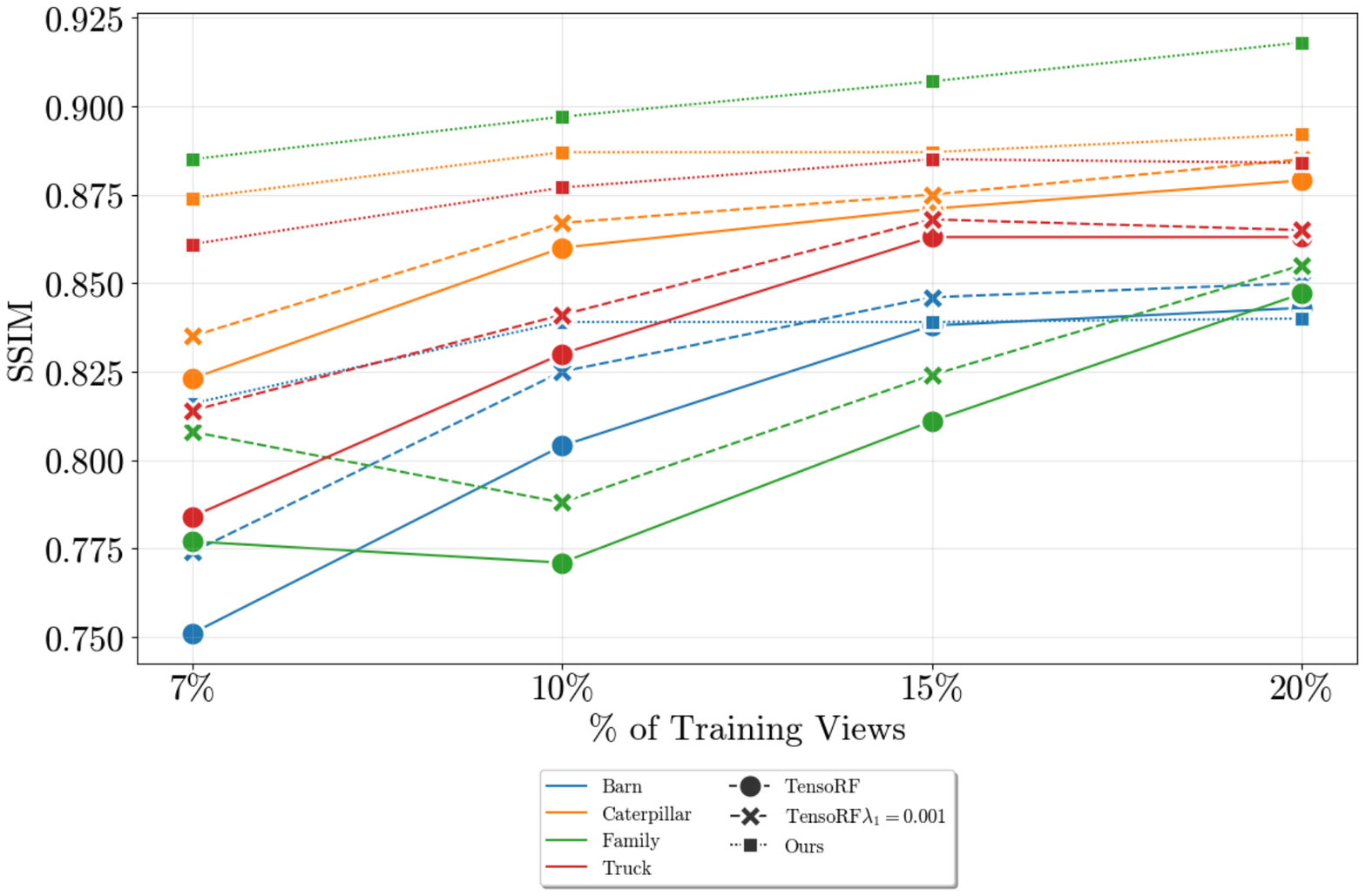}
            \caption{SSIM}
            \label{fig_app_tanksandtemple_ssim}
        \end{subfigure}    
        
    \caption{The line plots of PSNR and SSIM on the Tanks and Temples dataset varying the number of training views.}
    \label{fig_quantitative_result_tanksandtemple}
}    
\end{figure}

Quantitatively, the proposed method shows its strength, especially in SSIM scores. While PSNR is a valuable metric for image quality, it can be biased in this context due to the lack of mask information and the inclusion of full-resolution white backgrounds. On the other hand, SSIM focuses on the perceived quality of structural information in the images. As shown in \autoref{fig_app_tanksandtemple_ssim}, the proposed method consistently achieves higher SSIM scores across all scenes, indicating its superior capability in preserving the structural integrity and overall composition of scenes.

To sum up, the proposed method distinguishes itself from the baselines through its robust ability to preserve the global context of scenes, handle sparse input data effectively, and render images that are both structurally sound and visually realistic. These inherent properties highlight its potential for broader application in real-world scenarios, where input data is often sparse and incomplete.

\end{document}